\documentclass[preprint,3p]{elsarticle}

\PassOptionsToPackage{hyphens}{url}\usepackage{hyperref}
\usepackage[T1]{fontenc}
\usepackage[usenames,dvipsnames]{xcolor}
\usepackage[breakable, theorems, skins]{tcolorbox}
\usepackage{color,soul}
\tcbset{enhanced}
\usepackage{graphicx}
\usepackage{amssymb}
\usepackage{amsmath}
\usepackage{paralist}
\usepackage{todonotes}
\usepackage[algoruled,vlined,boxed,linesnumbered]{algorithm2e}
\SetKwRepeat{Do}{do}{while}
\usepackage{algpseudocode}
\usepackage{multirow, makecell}
\usepackage{enumitem}
\usepackage{wrapfig}
\usetikzlibrary{automata,arrows}
\usetikzlibrary{datavisualization}
\usetikzlibrary{datavisualization.formats.functions}
\usetikzlibrary{positioning, calc, backgrounds, arrows}
\usetikzlibrary{shapes.geometric}
\usetikzlibrary{decorations.pathreplacing}
\usetikzlibrary{patterns}
\usepackage{caption}
\usepackage{subcaption}
\usepackage{makecell}
\usepackage{lscape}
\usepackage{pifont}

\journal{Information Systems}

\usepackage[figuresright]{rotating}

\newtheorem{definition}{Definition}[section]

\makeatletter
\newcommand{\shorteq}{%
    \mkern3mu
  \settowidth{\@tempdima}{--}
  \resizebox{\@tempdima}{\height}{=}%
  \mkern3mu
}
\makeatother
\newcommand{\cmark}{\ding{51}}%
\newcommand{\xmark}{\ding{55}}%

\newcommand{\labels}{\Sigma}

\newcommand{\lbl}{\lambda}

\newcommand{\logL}{\mathit{L}}
\newcommand{\dafsa}{\mathit{D}}

\newcommand{\lnet}{\mathit{PN}}
\newcommand{\places}{\mathit{P}}
\newcommand{\netTransitions}{\mathit{T}}
\newcommand{\netArcs}{\mathit{F}}
\newcommand{\netLabel}{\lambda}

\newcommand{\inTr}[1]{\bullet{#1}}
\newcommand{\outTr}[1]{#1\bullet}
\newcommand{\wNet}{\mathit{WN}}

\newcommand{\sysNet}{\mathit{SN}}
\newcommand{\initMarking}{m_0}
\newcommand{\finalMarkings}{M_R}

\newcommand{\reachGraph}{\mathit{RG}}

\newcommand{\match}{\mathit{MT}}
\newcommand{\lhide}{\mathit{LH}}
\newcommand{\rhide}{\mathit{RH}}

\newcommand{\sync}{\beta}
\newcommand{\op}{\mathit{op}}

\newcommand{\filterSyncDFA}{\alignment|_{\dafsa}}
\newcommand{\filterSyncRG}{\alignment|_{\reachGraph}}
\newcommand{\alignment}{\epsilon}

\newcommand{\costF}{\mathit{g}}

\newcommand{\start}{\mathit{s}}
\newcommand{\TRtype}{\alpha}
\newcommand{\reps}{\mathit{k}}

\newcommand{\redLog}{\mathit{RL}}

\newcommand{\position}{\mathit{i}}

\newcommand{\alignments}{\mathcal{A}}

\newcommand{\reduce}{\mathit{reduce}}

\newcommand{\open}{\mathit{o}}

\begin{document}

\begin{frontmatter}



\title{Generalization in Automated Process Discovery:\\
        A Framework based on Event Log Patterns}


\author{Daniel Rei{\ss}ner\corref{cor1}\fnref{Melb}}
\ead{dreissner@student.unimelb.edu.au}
\cortext[cor1]{Corresponding author}
\author{Abel Armas-Cervantes\fnref{Melb}}
\ead{abel.armas@unimelb.edu.au}
\author{Marcello La Rosa\fnref{Melb}}
\ead{marcello.larosa@unimelb.edu.au}

\affiliation[Melb]{organization={University of Melbourne},
            country={Australia}}

\begin{abstract}
Automated process discovery is a tactical process mining capability that aims to describe the behavior of a business process recorded in an event log by means of a process model, in order achieve business process transparency. With a plethora of automated process discovery algorithms available, the importance of measuring the quality of a process model for a given event log has increased. 
One of the key quality aspects, generalization, is concerned with measuring the degree of overfitting of a process model w.r.t. an event log, since the recorded behavior is just an example of the true behavior of the underlying business process.
Existing generalization measures exhibit several shortcomings that severely hinder their applicability in practice. For example, they assume the event log fully fits the discovered process model, and cannot deal with large real-life event logs and complex process models.
More significantly, current measures neglect generalizations for clear patterns that demand a certain construct in the model. For example, a repeating sequence in an event log should be generalized with a loop structure in the model.
We address these shortcomings by proposing a framework of measures that generalize a set of patterns discovered from an event log with representative traces and check the corresponding control-flow structures in the process model via their trace alignment. We instantiate the framework with a generalization measure that uses tandem repeats to identify repetitive patterns that are compared to the loop structures and a concurrency oracle to identify concurrent patterns that are compared to the parallel structures of the process model.
In an extensive qualitative and quantitative evaluation using 74 log-model pairs using against two baseline generalization measures, we show that the proposed generalization measure consistently ranks process models that fulfil the observed patterns with generalizing control-flow structures higher than those which do not, while the baseline measures disregard those patterns. Further, we show that our measure can be efficiently computed for datasets two orders of magnitude larger than the largest dataset the baseline generalization measures can handle.
\end{abstract}


\begin{highlights}
\item A framework of generalization measures is defined in the field of automated process discovery that relates patterns of process behavior identified from an event log to the control-flow structures of a process model.
\item The framework is instantiated with a generalization measure that tests the repetitive and concurrent patterns of a process.
\item The proposed generalization measure can handle unfitting event logs, which is a hindrance for existing generalization measures.
\item An extensive evaluation with both qualitative and quantitative experiments shows that the proposed measure scales well to large datasets and consistently ranks process models that fulfil the observed patterns higher than models which do not, while state-of-the-art measures disregard those patterns.

\end{highlights}

\begin{keyword}
Process mining \sep Automated process discovery \sep Generalization \sep Event log patterns \sep Tandem repeats \sep Concurrency oracle


\end{keyword}

\end{frontmatter}


\section{Introduction}\label{sec:introduction}
Business processes are the backbone of organizations~\cite{fBPM2}. Processes such as loan origination in banks or claims handling in insurance companies and are executed thousands of times contributing to the core success factors of financial institutions. Business processes are usually supported by enterprise systems like loan management or payment systems in banks and claims management systems in insurance companies.
These systems record detailed execution traces of the processes in the form of \emph{event logs}. A trace in a sequence of events, i.e. activity occurrences such as ``Check loan application'' or ``Assess credit risk'' that are timestamped based on the activity completion time.

Process mining aims to gain insights from event logs in order to assist organizations in
their operational excellence or digital transformation programs~\cite{ProcessMiningBook,fBPM2}. 
Automated process discovery is a tactical process mining capability to achieve business process transparency by automatically discovering a process model from an event log. This task is challenging since event logs are often incomplete and exhibit only a sample of the possible behavior of the underlying business process. Quality measures have been devised to evaluate the result of discovery algorithms. These are fitness, precision, generalization and simplicity~\cite{ConformanceDimensions}.

In this article we focus on generalization, which measures the ability of a process model to generalize the sample behavior observed in an event log, hence avoiding overfitting that behavior. Current state-of-the-art measures, e.g.~\cite{NegativeEvents,AntiAlignments}, suffer from several drawbacks. 
First, they assume that event logs are fully fitting to the process model. 
This assumption is oftentimes violated in real-life event logs making the measures only applicable with workarounds such as applying the measure to an aligned event log which might distort the analysis results.
Second, current measures focus on whether the process model can cover a lot of variations in its behavior, but disregard clear patterns in the event logs that demand specific types of generalization. For instance, it is sensible to assume that a repeating sequence of activities in an event log should be generalized with a loop structure in the process model, or that activities observed in different orders in the log should be generalized with a concurrency structure in the process model. 
Last, current measures do not scale up to large and complex real-life event logs and process models. In our experiments, we found that current generalization measures were not applicable to event logs with over 280 unique traces. This is problematic since logging mechanisms of modern enterprise systems can easily lead to thousands of unique traces in a given event log, to reflect increasingly more complex business processes.

We aim to address these shortcomings by exploiting the notion of pattern in an event log. A pattern is a well-known ordering of activity instances in the event log, such as repeated activities or activity pairs observed in different orders across different log traces. 
These patterns are generalized naturally by corresponding control-flow structures in the process model, such as loops for sequences of repeated activities, or parallel blocks for activities that appear in different orders in the event log.
Based on the assumption that event log patterns capture generalization requirements, this article proposes a five-step framework to compute a generalization measure. First, the selected patterns are found in the input event log. Second, a set of representative traces is derived from the identified patterns. Third, trace alignments are computed between this set of representative traces and the input process model. Next, partial fulfilments of each pattern are extracted by comparing the alignments with the definitions of the patterns. Last, the generalization measure is computed by aggregating the partial pattern fulfilments into a single generalization value. In this article, we specifically focus on two types of patterns: repetitive and concurrency patterns.


We identify repetitive patterns with tandem repeats in the event log and then extend the tandem repeats such that the process model must include a loop structure to match the repeating sequence. We then measure the fulfilment of a repetitive pattern as the fraction of activities of the repeating sequence that can be matched by the process model in every iteration. 
Similarly, we identify concurrent patterns with a concurrency oracle by collapsing all traces in the log into partial orders and deriving sets of concurrent activities. The partial fulfilment of a concurrent pattern is then measured by the number of different orders of the concurrent activities that can be matched by the process model. Finally, the pattern fulfilments are aggregated into a single generalization value using their weighted average. 
We implemented the pattern generalization measure with repetitive and concurrent patterns as an open-source tool. Using the tool, we evaluated the measure extensively, using both qualitative and quantitative techniques, against two state-of-the-art baseline measures using 74 real-life event logs and the corresponding  process models discovered by two well-established process discovery algorithms.

The remainder of this paper is organized as follows. Section~\ref{sec:background} discusses current generalization measures and patterns that can be identified in event logs. Section~\ref{sec:preliminaries} introduces preliminary concepts that are used to present our technical contributions. Section~\ref{sec:approach} presents the generalization framework and its instantiation via a measure that exploits repetitive and concurrent patterns. Section~\ref{sec:evaluation} shows the results of both the qualitative and quantitative evaluation. Finally, Section~\ref{sec:conclusion} summarizes the contributions and discusses directions for future work.
\newpage
\section{Related work}\label{sec:background}
This section reviews existing generalization measures as well as the use of patterns in process mining.

\vspace{-.5\baselineskip}
\subsection{Generalization measures}\label{sec:generalization_measures}
\begin{figure}[h!]
\centering
\includegraphics[width=0.6\textwidth]{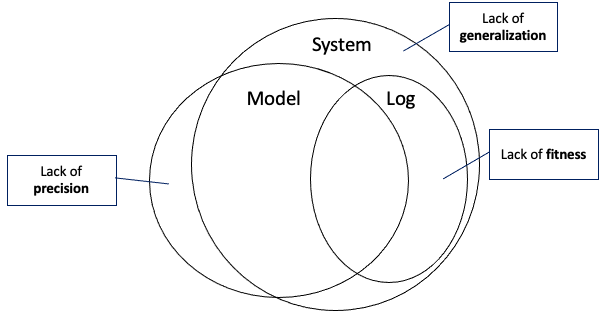}
\caption{Quality dimensions in process mining.}\label{fig:QualityDimensions}
\vspace{-.5\baselineskip}
\end{figure}

In process mining, the automated discovery of process models has received much attention. With a plethora of techniques developed to discover a process model from a given event log, the importance to assess the quality of process models has become essential. 
In particular, the quality will be measured by comparing three constructs of a process: the event log, the process model and the (process) system representing the recorded, the normative and the true behavior of a process, respectively.
Fig.~\ref{fig:QualityDimensions} illustrates how three orthogonal quality dimensions have been proposed based on a set comparison of the three constructs.
\emph{Fitness} (a.k.a. recall) measures 
the fraction of the behavior of a log that is covered by the model. 
\emph{Precision} measures the fraction of the behavior of a process model that was not observed in the event log. The \emph{generalization} dimension considers that an event log is just a sample of the true behavior of the process system. It is concerned with measuring the degree to which a process model describes the underlying system and hence avoids overfitting the event log, i.e. it measures the fraction of the system that is covered in the model.
Besides automated discovery, fitness, precision and generalization also can be used for the purpose of conformance checking in process mining~\cite{ProcessMiningBook,fBPM2,GeneralizationAxioms}.
Finally, a fourth orthogonal quality dimension \emph{simplicity} measures how complex a process model is, and it is based on the Occam's Razor principle: the best model is the simplest model explaining the behaviour of the log.
Please note that this work only considers logs that do not contain behavior outside the process system, which can be removed in a preprocessing step as described in~\cite{EventLogCleaning}.
This article proposes a framework for measures in the generalization dimension. For that purpose, we discuss the generalization definition next.

\emph{Generalization definition.}
Generalization measures how well the process model describes the underlying system~\cite{ConformanceDimensions}, however, 
there lies a significant problem since the complete behavior of the system, i.e. the true behavior of a process, is unknown.
In~\cite{SystemFitness}, the behavior of the process system was specified to act like an event log with unknown traces. This allows to relate the system to the process model with both existing fitness and precision measures adapted to the system instead of the log, i.e. the system fitness and system precision measures the fraction of the system behavior that is covered by the model and the fraction of model behavior that is covered by the system, respectively.
It is common to assume that the system precision is 1, i.e. that a process model does not contain behavior outside the system. Hence, the generalization of a process model should measure the fitness of the model towards the system.
Additionally, generalization measures should also be orthogonal to other measures such as log fitness and log precision. Hence, the generalization measure should only measure the fraction of new unseen behavior of the system that is covered by the process model and has not been recorded in the event log yet. 
Given a process model and an event log, a common definition for generalization measures then was proposed to measure the probability that new unseen traces fit the process model~\cite{ConformanceDimensions,GeneralizationAxioms}.

Hence, every generalization measure faces the challenge of defining the behavior of unseen traces of the system based on the example behavior that was observed in the event log. Next, we discuss current state-of-the-art measures for the generalization dimension.

\emph{Alignments generalization.} 
The first generalization measure proposed in~\cite{AlignmentsGeneralization} relates each activity occurrence, a.k.a. event, in an event log to an execution state in the process model just before executing the activity. For every execution state, it collects the frequency of events being executed in the state and the number of different activities that can be executed at the execution state. An estimator then measures the probability that a new event 
visiting the state will be a new unseen event as a fraction of different activities towards the frequency of the state. A generalization measure is then one minus the average of all estimators for all execution states. 
The alignment generalization measure is however only applicable to fully fitting event logs. The proposed workaround of using alignments for the traces of an event log with the process model as a new event log distorts the measurement of generalization.
We did not include the alignment generalization in our evaluation since it was later shown in~\cite{NegativeEvents} that this measure assigns high generalization values to simple models that do not contain any generalizing constructs. 

\emph{Negative events generalization.}
The measure proposed in~\cite{NegativeEvents} explicitly defines new and unseen behavior by inserting negative events at each trace position by cross referencing events from other traces. These negative events are weighted with a moving window by the fraction of the prefix that has been observed in another trace to include a measure for the confidence of the negative events.
The measure then counts the number of allowed generalizations by replaying the traces of the event log on the process model and at each state testing whether negative events could be replayed. If that is the case the allowed generalizations is increased by one minus the weight of the negative event and otherwise the disallowed generalizations are increased. The generalization is then the fraction of allowed generalizations to the overall number of generalizations.

\emph{Anti-alignments generalization.}
In~\cite{AntiAlignments}, a generalization measure was proposed based on the idea that if newly observed behavior introduces only new paths through the process model but does not introduce new states then the process model generalizes new behavior well.
This idea is instantiated by means of a recovery distance, i.e. the maximal distance between the states visited by the log and the states visited by an anti-alignment. A process model generalizes the behavior of an event log well, if it has a minimal recovery distance and a large distance to the anti-alignment. A generalization measures then is defined for each anti-alignment when leaving exactly one trace out of the event log. 
Hence, the anti-alignment generalization does not specify new unseen behavior outside the event log.
Further, the anti-alignments measure is also only applicable to fully fitting event logs.

\emph{Adversarial networks generalization.}
A recent generalization measure proposed in~\cite{GANS_Generalization} uses sequence generative adversarial networks and the Metropolis-Hastings algorithm to sample unobserved trace variants based on the trace distribution of the input event log. The generalization is then computed as the harmonic mean between the fitness and the precision of the process model and the newly computed trace variants. 
As the authors state, it still has to be seen how well the generated trace variants describe the underlying system as the system's behavior is unknown, in the absence of a ground truth in the experiments reported in this paper.
While the application of adversarial networks is promising to find new trace variants, the general applicability of this approach in practice is hindered by its specific hardware requirements (high GPU).

This article presents a more structured approach to identify behavior not observed in an event log. The approach is based on the idea that patterns in an event log are excerpts of the process behavior.
For each pattern we propose to find a set of traces representing unseen behavior of the system. We then compute alignments between the unobserved traces and the process model, and then test the fulfilment of each pattern by analysing the such alignments.
For example, a sequence of activities in the event log that is repeated a certain number of times implies that this sequence could be repeated with a different number of repetitions in the underlying system. 
This observation should be generalized with a loop structure for the repeating sequence in the process model. We derive a new unobserved trace with a high number of repetitions and align it with the process model. If the alignment could match all repetitions of the sequence then the pattern was successfully generalized with a loop structure in the process model.
\newpage
The proposed pattern-generalization measures have the following advantages:
\begin{inparaenum}[(1)]
    \item the generalization is more intuitive for process analysts to understand because the unseen behavior is related to specific patterns;
    \item the proposed measure can handle also unfitting event logs because it uses alignments to relate the model and the log, and 
    \item improvements for generalization issues can be easily identified by testing certain control structures related to the patterns in the process models.
    \end{inparaenum}

\subsection{Patterns in process mining}
Patterns are abstractions for recurring structures in non-arbitrary contexts~\cite{DefPatterns}. They have first been used in the field of computer science to define 23 design patterns for interactions in object-oriented systems~\cite{PatternsInCompScience}. Patterns have the advantage of specifying behavior of a system while being both independent from technical implementations as well as specific domain requirements~\cite{AdvantagePatterns}. 
In this article, we consider patterns in the field of process mining that describe the behavior of a process system.
We aim to identify patterns based on sample structures in the recorded behavior of a process, i.e. in its event log. The patterns can inform about more general and possibly unobserved behavior of the underlying process and can hence be used to define a new generalization measure.
Next, we survey existing process patterns and discuss their suitability for defining generalization measures.

\emph{Workflow patterns.} In~\cite{WorkflowPatterns} a set of 20 workflow patterns have been proposed that define possible constructs that describe different possible behavior in a workflow process. The workflow patterns have been grouped into seven categories based on different aspects of a process.
Basic control flow structure patterns capture the elementary control structures in a process such as sequences, parallel splits, synchronizations, exclusive choices or simple merges. 
Advanced branching and synchronization patterns consider control structures that do not have straight forward support in workflow systems but have common occurrences in real-life processes such as multi-choice, synchronizing merge, multi merge and discriminator patterns. 
Structural patterns cover typical restrictions posed on workflow systems such as arbitrary loops and implicit termination. Multi-instance patterns describe common structures used when handling multiple instances of a process simultaneously, i.e. handling multiple instances with design time knowledge and with or without runtime knowledge.
State-based patterns capture situations where the execution of the following activities depends on the execution state of a process instead of explicit choices such as deferred choice, interleaved parallel routings or milestone patterns.
Finally, cancellation patterns handle situations where an activity cancels another activity or the entire case.
While the workflow patterns thoroughly capture different general behavior of a process, their suitability to define a generalization measure is limited since it is not specified how sample behavior of the patterns is recorded in an event log. 

\emph{Change patterns.} In~\cite{ChangePatterns} 18 patterns were proposed for abstracting changes to a process. The article considers adaptation patterns documenting high level changes to a process or region-specific change patterns. The adaptation patterns cover cases for structural process changes such as inserting/deleting fragments, moving/replacing fragments, adding/removing levels, adapting control dependencies or changing transition conditions. The region-specific change patterns define runtime changes for specific regions of a process to address uncertainties by late selecting, late modelling or by late composing fragments of a process or by instantiating an activity multiple times.
Change patterns can imply unobserved behavior of a process by considering which changes to the recorded behavior could potentially correspond to behavior of the underlying process. However, similar to the workflow patterns it was not specified how the patterns would manifest in an event log and hence change patterns are also not suitable to define a generalization measure.

\emph{Event log patterns.} In~\cite{TaxonomyOfPatterns} patterns were defined based on observations in the event log that correspond to control structures in a process. These patterns were then used to discover a process model with the control structures defined by the patterns.
In particular, repeating sequences of activities in an event log, aka. tandem repeats, were linked to loops in a process model. Regions of similar activities, i.e. conserved regions, were linked to sub-processes. Finally, relaxing the orders of activities with repeat alphabets is linked to parallel blocks in a process model. 
Other discovery techniques~\cite{CausalRunsDiscovery,Unfolding} proposed the use of concurrency oracles instead of repeat alphabets to derive partial orders linked to parallel blocks in the process model. 
\newpage
The article~\cite{TaxonomyOfPatterns} also discusses strategies for defining more complex patterns that combine the three basic pattern types. First, more abstract sub-processes can be identified by first collapsing the orders of activities and then identifying conserved regions. More complex loop structures such as loops with skip-able activities can be found with tandem repeats that allow approximations. Finally, nested control structures in a process model such as a parallel block inside a loop or vice versa or nested loops can be identified by iteratively applying and collapsing the base patterns.
The event log patterns are suitable for a generalization measure since they both specify example behavior in an event log as well as generalize the behavior of underlying process in a structured way linking them to control structures. 

In this article, we propose a framework for pattern-based generalization measures that assesses how well patterns identified in an event log are abstracted with corresponding control structures in the process model. 
Based on the framework, an analyst can select all relevant process patterns to measure the generalization of a process model.
The focus of the article lies on how to measure the generalization 
for the selected patterns rather than which patterns should be selected to form an ideal generalization measure.
Next, we discuss some criteria to select patterns for the generalization measure.
First, patterns should be linked to the behavior recorded in an event log.
This is a necessary requirement 
since a generalization measure has to assess how well a process model abstracts 
the recorded behavior of an event log. 
So far, only the event log patterns from~\cite{TaxonomyOfPatterns} fulfill this requirement. 
Another criteria for selecting patterns 
is to avoid over-generalizations.
The event log patterns differ in their level of abstraction from the recorded process behavior, i.e. approximate patterns abstract more than the base patterns. Since the true behavior of the underlying process is unknown, it is uncertain whether high level abstractions are still a part of the process system are go beyond its behavior. 
Finally, patterns should induce behavior unobserved in the event log to properly adhere to the generalization definition from subsection~\ref{sec:generalization_measures}. For example, the conserved regions as a pattern to identify sub-processes will not find unobserved behavior. In comparison, generalizing a tandem repeat in an event log with a loop in a process model will include unobserved behavior since the loop will contain the repeating sequence with a different numbers of repetitions.
We initiate the framework by proposing a pattern generalization measure with the two remaining base patterns from~\cite{TaxonomyOfPatterns} that fulfil all of the selection criteria, i.e. with concurrent patterns based on a concurrency oracle and with repetitive patterns based on tandem repeats. 
\newcommand{\rightshift}{\blacktriangleright}
\newcommand{\uniqueTraces}{\mathit{unique}}
\newcommand{\countTr}{\mathit{count}}
\newcommand{\concOracle}{\gamma}
\newcommand{\df}{\mathit{df}}
\newcommand{\conc}{\parallel}
\newcommand{\dfCount}{\mathit{dfC}}
\newcommand{\noiseThr}{\epsilon}
\newcommand{\PO}{\mathit{PO}}
\newcommand{\Events}{\mathit{E}}
\newcommand{\Causality}{\leq}
\newcommand{\initialEvent}{\mathit{s_0}}
\newcommand{\finalEvent}{\mathit{f_0}}
\newcommand{\concat}{\text{\raisebox{.8ex}{$\smallfrown$}}}
\newcommand{\multisetplus}{\uplus}
\newcommand{\logPOs}{\mathit{PO}_{\logL}}
\newcommand{\po}{\pi}
\newcommand{\head}{\mathit{head}}
\newcommand{\tail}{\mathit{tail}}
\newcommand{\universeTraces}{\mathcal{T}}
\newcommand{\universePO}{\mathcal{PO}}
\newcommand{\poLabelling}{\ell}
\newcommand{\universeLogs}{\mathcal{L}}
\newcommand{\universeSystemNets}{\mathcal{S}}
\newcommand{\syncCostStandard}{\mathit{cost}}

\section{Preliminaries}\label{sec:preliminaries}
This section presents the foundations of the paper. 
The first subsection presents the concepts related to event logs. The second subsection covers the concepts related to the first type of generalization pattern we take into account: tandem repeats. 
Then, in the third subsection, concurrency oracles and partial orders are defined, concepts relevant to the second type of generalization. 
Finally, the last two subsections introduce Petri nets and alignments, respectively. 

For the sake of consistency, throughout this paper, $\labels$ represents a set of activity labels. 

\subsection{Event logs}
\emph{Event logs}, or simply \emph{logs}, record the executions of a business process. These executions are stored as sequences of \emph{events}, which are activity occurrences. A sequence of events corresponding to an instance of a process is called a \emph{trace}. Each event in the trace is represented by the label of a corresponding activity, and it is possible that several events within a trace are occurrences of the same activity, in which case the trace will contain repeated labels. 
Event logs are a multiset of traces because several executions of a process may represent the same activities occurring in the same order. 

\begin{definition}[Trace and Event Log]\label{def:log}
Let $\labels$ be a set of activity labels, $E$ be a set of events and $\lambda: E \rightarrow \labels$ be a labelling function. A \emph{trace} $t$ is a finite sequence $t=\langle \lambda(e_1), \lambda(e_2),\dots,\lambda(e_n)\rangle \in \labels^*$, where $\lambda(e_i)$ is the activity label of the event $e_i$ for $1 \leq i \leq n$. 
Then, an \emph{event log} $\logL$ is a multiset of traces. 
The universe of traces and event logs are represented by $\universeTraces$ and $\universeLogs$, respectively.
\end{definition}

A notion of equivalence can be defined between pairs of traces, so that two traces are equivalent ($\equiv$) if they have the same activity occurrences (events) that occurred in the same order. In a log, a trace $t$ is a \emph{distinct trace} if there is not another trace equivalent to $t$.
The set representation of a log $\logL$, where every trace is a \emph{distinct trace}, is shorthanded as $\uniqueTraces(\logL)$. $\countTr(t,\logL)$ counts the number of occurrences of a trace $t\in\logL$ in the log $\logL$, i.e. $\countTr(t,\logL) = \left|\{t'\in\logL\mid t' \equiv t\}\right|$. 
Given a trace $t$ and a count $c_1$, we use the notation $\logL\multisetplus(t,c_1)$ to add the trace to the log $c_1$ times. If $\logL$ already contains $t$ with another trace count $c_2$, then the two trace counts will be added, i.e. $\countTr(t,\logL\multisetplus(t,c_1))=c_1+c_2$.

Let us define some assorted operations over traces. The size of a trace $t$, denoted as $\left|t\right|$, is the number of elements in $t$, and the $i$-th element can be accessed as $t[i]$, $1 \leq i \leq \left|t\right|$.
A subtrace of $t$ from position $i$ to position $j$ is shorthanded as $t[i,j]$. Two traces can be concatenated using the operator $\concat$, e.g., $t = t[1]\concat t[2,\left|t\right|]$.
Finally, $\head(t)$ returns the first element of a trace, while the rest of the elements are returned by $\tail(t)$, i.e. $\head(t)=t[1]$ and $\tail(t)=t[2,\left|t\right|]$.

\begin{figure*}[htbp]
\centering
\resizebox{0.66\textwidth}{!}{ 
 \tikzstyle{ID} = [draw, rectangle, fill=white, align=center, minimum height=5mm, text width={width("$t  (1)  t$")}, font=\footnotesize]
 \tikzstyle{TRACE} = [draw, rectangle, fill=white, align=left, minimum height=5mm, text width={width("$t X,A,X,A,X,A,X,A,X,A,C,B t$")}, font=\footnotesize]
 \tikzstyle{COUNT} = [draw, rectangle, fill=white, align=center, minimum height=5mm, text width={width("$t count(dt,L) t$")}, font=\footnotesize]
 \begin{tikzpicture}[>=stealth', node distance=-0.3pt]
  \node[TRACE] (log) {\bf{$dt\in\uniqueTraces(\logL)$}};
  \node[ID, left=of log] (id) {\bf{ID}};
  \node[COUNT, right=of log] (count){\bf{$\countTr(dt,\logL)$}};
  \node[TRACE, below=of log] (trace1) {$\langle X,A,B,C \rangle$};
  \node[ID, left=of trace1] (id1) {(1)};
  \node[COUNT,right=of trace1] (count1) {1000};
  \node[TRACE, below=of trace1] (trace2) {$\langle X,A,C,B \rangle$};
  \node[ID, left=of trace2] (id2) {(2)};
  \node[COUNT,right=of trace2] (count2) {1000};
  \node[TRACE, below=of trace2] (trace3) {$\langle A,B,C \rangle$};
  \node[ID, left=of trace3] (id3) {(3)};
  \node[COUNT,right=of trace3] (count3) {200};
  \node[TRACE, below=of trace3] (trace4) {$\langle B,A,C \rangle$};
  \node[ID, left=of trace4] (id4) {(4)};
  \node[COUNT, right= of trace4] (count4) {200};
  \node[TRACE, below=of trace4] (trace5) {$\langle C,A,B \rangle$};
  \node[ID, left=of trace5] (id5) {(5)};
  \node[COUNT, right= of trace5] (count5) {200};
  \node[TRACE, below=of trace5] (trace6) {$\langle X,X,X,X,A,A,A,A,B,C \rangle$};
  \node[ID, left=of trace6] (id6) {(6)};
  \node[COUNT, right= of trace6] (count6) {1000};
  \node[TRACE, below=of trace6] (trace7) {$\langle X,X,A,X,X,A,X,B,C \rangle$};
  \node[ID, left=of trace7] (id7) {(7)};
  \node[COUNT, right= of trace7] (count7) {500};
  \node[TRACE, below=of trace7] (trace8) {$\langle X,A,X,A,X,A,C,B \rangle$};
  \node[ID, left=of trace8] (id8) {(8)};
  \node[COUNT, right= of trace8] (count8) {200};
  \node[TRACE, below=of trace8] (trace9) {$\langle X,A,X,A,X,A,X,A,X,A,C,B \rangle$};
  \node[ID, left=of trace9] (id9) {(9)};
  \node[COUNT, right= of trace9] (count9) {200};
  \end{tikzpicture}
  }
 \caption{Event log of the running example.}\label{fig:runningExampleLog}
\end{figure*}

Figure~\ref{fig:runningExampleLog} shows an event log, where every trace is distinct and has an identifier and a trace count. 
The identifier is used to keep track of the unique traces in our running example.

\subsection{Tandem repeats}\label{sec:tandemRepeats}

The first type of generalization considered in this paper is based on repetitive patterns extracted from event logs. 
These patterns, which are repeating sequences of events, are called \emph{tandem repeats}.
A tandem repeat in a trace $t$ is a triplet $(\start,\TRtype,\reps)$, where $\start$ is the starting position of the tandem repeat, $\TRtype$ is a sequence of events representing the repetitive pattern -- \emph{a.k.a. repeat type} --, and $\reps$ is the number of repetitions of $\TRtype$ in $t$. 
In order to identify the tandem repeats in a trace, we use the oracle $\Delta$. 
For a trace $t$, $\Delta(t)$ retrieves the set of tandem repeats, such that the repeat types occurs at least twice (in other words, any tandem repeat $(\start,\TRtype,\reps)$ has $\reps \geq 2$). 
For the evaluation (Section~\ref{sec:evaluation}), the approach proposed by Gusfield and Stoye~\cite{FindingTRs} was used as the oracle $\Delta$. That approach uses suffix trees to find tandem repeats in linear time, linear with respect to the length of the input string, and defines an order between the tandem repeats by reporting the leftmost occurrences first, 
i.e. tandem repeats shifted right by any amount of characters are omitted. Further improvements to the technique in~\cite{FindingTRs} have been proposed by using suffix arrays as the underlying data structure~\cite{SuffixArrays}. 

The tandem repeats considered in this work are \emph{maximal} and \emph{primitive}~\cite{BoseTRinProcessMining}. A tandem repeat is called maximal if no repetitions of the repeat type occur at the left or right side of the tandem repeat. The tandem repeat is primitive, if the repeat type is not itself a tandem repeat. 

\newpage
\begin{definition}[Maximal and primitive tandem repeat with no right shifts]
Given a trace $t$, a tandem repeat $(\start,\TRtype,\reps) \in \Delta{(t)}$ is \emph{maximal}, if neither $(\start-\left|\TRtype\right|,\TRtype,\reps+1)$ nor $(\start,\TRtype,\reps+1)$ is a tandem repeat,
and \emph{primitive} if $\alpha$ is not itself a tandem repeat. 
An operation to shift a tandem repeat to the right by $x$ characters is defined as $(\start,\TRtype,\reps)\rightshift x=(\start+x,\TRtype[x+1,\left|\TRtype\right|]\concat \TRtype[1,x],\reps)$ for $1\leq x < \left|\TRtype\right|$. All right-shifts of any tandem repeats are omitted, i.e. for a $(\start,\TRtype,\reps) \in \Delta(t)$ then $\forall_{1\leq x < \left|\TRtype\right|} (\start,\TRtype,\reps)\rightshift x \notin \Delta(t)$.
\end{definition}

Figure~\ref{fig:runningExampleTandemRepeats} shows the primitive and maximal tandem repeats with no right shifts for the event log of Fig.~\ref{fig:runningExampleLog}. 
First, traces with IDs (1)-(5) are not included in Fig.~\ref{fig:runningExampleTandemRepeats} because they do not contain any tandem repeats.
Trace (6) contains one tandem repeat $(1,X, 4)$ starting at position 1, where the repeat type $X$ is repeated 4 times. 
Another possible tandem repeat for trace (6) is $(1,XX, 2)$, but this is not a primitive repetitive type because $XX$ is itself another tandem repeat $(1,X,2)$. 
The tandem repeat $(1,XXA,2)$ in trace (7) is primitive because its repeat type is not a tandem repeat despite containing the tandem repeat $(1,X,2)$ -- 
this tandem repeat $(1,X,2)$ is also known as \emph{nested tandem repeat}.
Another tandem repeat in trace (7) is  $(2, XAX,2)$, but this is omitted since it is the same as $(1,XXA,2)$ shifted right by one character.
Gusfield and Stoye~\cite{FindingTRs} show how to avoid detecting tandem repeats shifted right by any number of characters.
Last, trace (8) contains a tandem repeat $(1,XA,2)$, but it is omitted because it is not maximal, i.e. it can be extended to the right side by one more repetition to $(1,XA, 3)$.

\begin{figure*}[htbp]
\centering
\resizebox{0.6 \textwidth}{!}{
 \tikzstyle{ID} = [draw, rectangle, fill=white, align=center, minimum height=5mm, text width={width("$t  (1)  t$")}, font=\footnotesize]
 \tikzstyle{block} = [draw, rectangle, fill=white, align=left, minimum height=5mm, text width={width("$t$  test Maximal and primitive Tandem Repeat $t$")}, font=\footnotesize]
 \begin{tikzpicture}[>=stealth', node distance=-0.3pt]
  \node[block] (log) {\bf{Maximal and primitive Tandem Repeats}};
  \node[ID, left=of log] (id) {\bf{ID}};
  \node[block, below=of log] (trace6) {$(1, X, 4), (5, A, 4)$};
  \node[ID, left=of trace6] (id6) {(6)};
  \node[block, below=of trace6] (trace7) {$(1, XXA, 2), (1,X,2), (4,X,2)$};
  \node[ID, left=of trace7] (id7) {(7)};
  \node[block, below=of trace7] (trace8) {$(1, XA, 3)$};
  \node[ID, left=of trace8] (id8) {(8)};
  \node[block, below=of trace8] (trace9) {$(1, XA, 5)$};
  \node[ID, left=of trace9] (id9) {(9)};
  \end{tikzpicture}
  }
 \caption{Primitive and maximal tandem repeats with no right shifts
  for the event log in Fig.~\ref{fig:runningExampleLog}.}\label{fig:runningExampleTandemRepeats}
\end{figure*}

\subsection{Concurrent events}\label{sec:Concurrency}

Another type of generalization considered in this paper is concurrency. Traces of a log are transformed into partially ordered sets of events, \emph{partial orders} for short, where it is possible to represent concurrency relations between events. The computation of the concurrency relations is done via a concurrency oracle, denoted as $\concOracle(\logL)$. The concurrency oracle $\concOracle(\logL)$ is a black box that receives an event log $\logL$ and returns the concurrency relations between events, as well as the partial order representation for each trace in the log. 
Let us define a partial order first, before formally introducing the concurrency oracle. 

\begin{definition}[Partial order]
Given a set of activity labels $\labels$, a \emph{partial order} $\po$ is a tuple $\po=(\Events, \Causality, \poLabelling)$, where $\Events$ is a set of events, $\Causality$ is the partial order over $\Events$ and $\poLabelling : \Events\rightarrow\labels$ is a labelling function. There are two special events $\initialEvent,\finalEvent\in\Events$ representing the initial and final events, respectively, such that $\forall e \in E : \initialEvent \leq e  \leq \finalEvent$. The universe of partial orders is denoted as $\universePO$.
\end{definition}

Intuitively, $\Causality$ defines the order of execution between the events. For instance, $a \Causality b$ represents that event $a$ has to occur before $b$ can occur. Formally, given a set of events $X$, $\Causality$ is reflexive ($x \Causality x$ for all $x \in X$), antisymmetric (if $x \Causality y$ and $y \Causality x$ then $x=y$) and transitive (if $x \Causality y$ and $y \Causality z$ then $x \Causality z$).
In the following definition, the concurrency oracle is defined as a pair of functions, one that retrieves the concurrency relations for a pair of events, and another that transforms a trace into a partial order. 

\newpage
\begin{definition}[Concurrency oracle]\label{def:conc:oracle}
Let $\logL$ be an event log, $t=\langle l_1,l_2,\dots,l_n\rangle \in \uniqueTraces(\logL)$ be a trace, and $\labels$ be the set of activity labels. 
A \emph{concurrency oracle} is the tuple $\concOracle = (\conc, \xi)$ where: 

\begin{itemize}
\item $\conc$
represents the concurrency between pairs of activity labels. $\conc(l_x,l_y) = 1$ iff $l_x$ and $l_y$ are concurrent, where $1 \leq x,y \leq n$; whereas $\conc(l_x,l_y) = 0$ if $l_x$ and $l_y$ are not concurrent. 
If one label is empty, i.e. $l_x=\perp$ or $l_y=\perp$, $\conc$ returns 0. 
Concurrency is symmetric, hence $\conc(y,x) = \conc(x,y)$.

\item $\xi : \universeTraces \rightarrow \universePO$ maps traces to partial orders. Let $t'=\langle\initialEvent\rangle\concat\langle e_1,e_2,\dots,e_n\rangle\concat\langle\finalEvent\rangle$ be a sequence of events, such that $l_i = \lambda(e_i)$, which is the event at the i$th$ position in $t$. 
The partial order of $t$ is $\xi(t) = (\Events, \Causality,\poLabelling)$, where $\Events = \{e ~|~ e \in t'\}$; $\poLabelling[e_i \rightarrow \lambda(e_i)]$ for $1\leq i \leq n$, $\poLabelling[\initialEvent \rightarrow \perp]$ and $\poLabelling[\finalEvent \rightarrow \perp]$; and 
$\Causality= (\{(e,e') ~\mid~ e = s_0 \lor e' = f_0 \lor (e = e_x \land e' = e_y \land x\leq y \land \conc(\poLabelling(e_x),\poLabelling(e_y))=0)\})^*$.

The transitive reduction of the partial order is denoted as $\Causality^-$, where $(e_x,e_y) \in \Causality^-$ if $\nexists(e_x,e_z),(e_z,e_y)\in\Causality^-$ for $1\leq x\leq z\leq y\leq\left|t'\right|$.

\end{itemize}

\end{definition}

Using a concurrency oracle, a multiset of partial orders can be obtained from an event log $\logL$, where $\xi(t)$ is a partial order for a trace $t \in \logL$. 
Note that two traces that are not equivalent can yield to the same partial order in the presence of interleavings.
We overload the notation of functions $\uniqueTraces$ and $\countTr$ for partial orders, such that $\uniqueTraces(\xi(\logL))$ retrieves the unique partial orders for an event log and $\countTr(\po,\xi(\logL))$ returns the count of a given partial order $\po$ in the log.

The definition of a concurrency oracle is generic and can be instantiated with different sets of concurrency relations. We show some existing instantiations next. 

\subsubsection{Global oracles}
Global oracles define concurrency relations at the level of activities (event labels), so that if a pair of events are observed concurrently, then the corresponding activities are assumed to be concurrent everywhere in the log (globally). Existing techniques that have been adopted as global concurrency oracles use, so-called, directly follow relations to define concurrency. 
An activity $a$ is directly followed by an activity $b$ iff there is a trace $t=\langle l_1,l_2,\dots,l_n\rangle \in \logL$, where $l_i = a$ and $l_{i+1} = b$, for $1\leq i < \left|t\right| -1$. By the abuse of notation, let $\df(x,y) = 1$ denote when $x$ is directly followed by $y$, and $\df(x,y) = 0$ otherwise. Some of the techniques commonly used as global concurrency oracles are presented next.

\begin{compactenum}
   \item The \emph{alpha concurrency oracle} \cite{Alpha-Concurrency} deems two activities $x$ and $y$ as concurrent, $\conc(x,y)=1$, iff $\df(x,y) = 1$ and $\df(y,x) = 1$. This concurrency oracle has the pitfall of mixing up concurrency with short loops, e.g., trace (8) in Fig.~\ref{fig:runningExampleLog} represents a short loop of activities $X$ and $A$. 
   \item The \emph{alpha+ concurrency oracle} \cite{Alpha+-Concurrency} addresses the shortcoming of the alpha concurrency oracle by adding an additional constraint: two concurrent activities cannot occur in a short loop in any trace of the event log, i.e. $\conc(x,y)=1$ iff $\nexists t\in\logL : t[i]=x\land t[i+1]=y\land t[i+2]=x\lor t[1]=y\land t[i+1]=x\land t[i+2]=y$ where $1\leq i\leq\left|t\right|-2$.
   \item The \emph{alpha++ concurrency oracle} \cite{Alpha++-Concurrency} uses information about the execution time (start and end) of the events for discovering concurrency as overlapping of timestamps. In this oracle, two activities $x$ and $y$ are concurrent, if the start of $y$ was recorded after the start but before the end of $x$. Unfortunately, if the log does not contain information about the start and end timestamps for the events, then it is not possible to use this oracle. This oracle is not used in this paper because the considered datasets did not have timestamp information.
\end{compactenum}

\begin{figure}[!h]
\centering
\begin{subfigure}{.25\textwidth}
  \centering
  \includegraphics[width=0.85\textwidth]{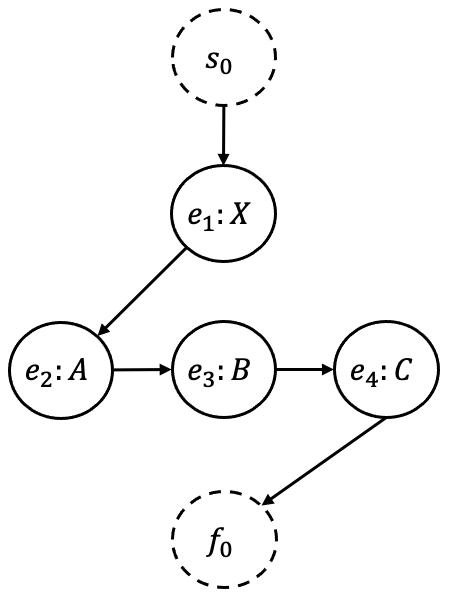}
    \caption{Graph of trace (1)} \label{fig:PO_trace}
    \vspace{\baselineskip}
\end{subfigure}%
\hspace{1cm}
\begin{subfigure}{.25\textwidth}
  \centering
  \includegraphics[width=0.85\textwidth]{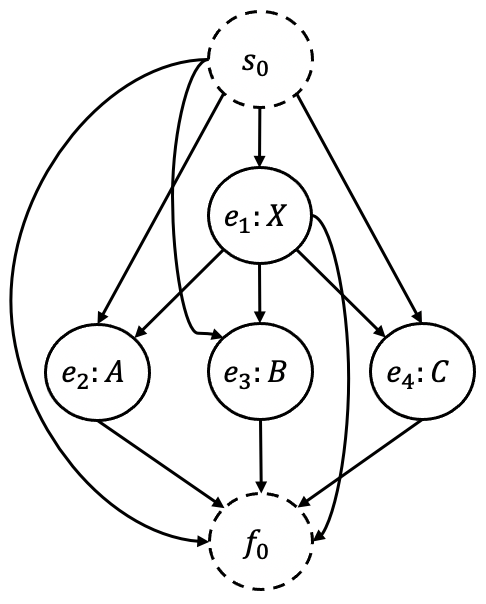}
    \caption{Removing arcs of concurrent activities} \label{fig:PO_RemoveArcs}
    \vspace{\baselineskip}
\end{subfigure}%
\hspace{1cm}
\begin{subfigure}{.25\textwidth}
  \centering
  \includegraphics[width=0.85\textwidth]{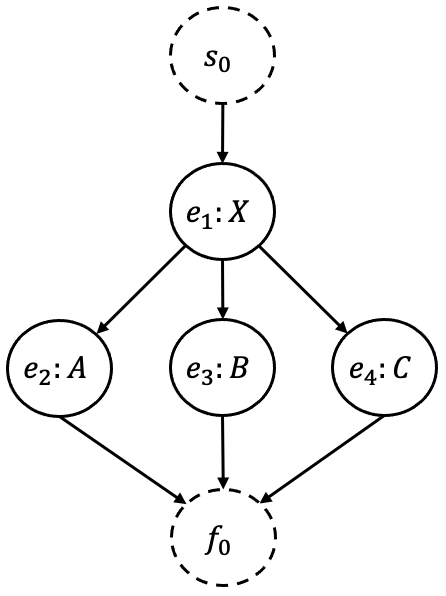}
    \caption{Transitive Reduction} \label{fig:PO_TransitiveReduction}
    \vspace{\baselineskip}
\end{subfigure}
\caption{Building a partial order for trace (1) of the running example from Fig.~\ref{fig:runningExampleLog}}
\label{fig:BuildingPO}
\end{figure}

Let's consider the following example, in the log displayed in Fig.~\ref{fig:runningExampleLog}, the alpha+ concurrency oracle would extract three pairs of concurrent events: $\concOracle(\logL) = \{(A,B),(A,C),(B,C)\}$. 
Figure~\ref{fig:BuildingPO} shows the steps to transform trace (1) into a partial order. 
Figure~\ref{fig:PO_trace} shows a graphical representation of trace (1), where the nodes are the events and the arrows represent the order defined in the trace. Observe that this graph shows the two special nodes $s_0$ and $f_0$ in the partial order.
Figure~\ref{fig:PO_RemoveArcs} shows the resulting partial order, where (A,B),(B,C) and (A,C) are concurrent, and Fig.~\ref{fig:PO_TransitiveReduction} shows its transitive reduction. 
The partial order represents that, after executing activity X, the three activities A,B and C can occur concurrently. Observe that several traces may lead to the same partial order. For example, trace (2) will also lead to the same partial order in Fig.~\ref{fig:PO_RemoveArcs}.


\subsubsection{Global oracles with infrequency filtering}
Infrequent behaviour can contribute to observe interleaving pairs of activities, which can lead to numerous concurrency relations when using global oracles. To handle this cases, frequency filters can be applied to try to remove such infrequent behaviour and limit the number of concurrency relations detected. For example, in~\cite{GlobalConcFilter} a filter was applied to the frequency of directly follows relation. This filter uses the number of the directly follow relations, denoted by the function $\dfCount$, i.e. given two activities $x$ and $y$, $\dfCount(x,y)$ will return the frequency in which $x$ was directly succeeded by $y$ in any trace $t\in\logL$, i.e. $\dfCount(x,y)=\left|\{(x,y)_{t,i} \mid t\in\logL \land 1\leq i<\left|t\right| \land t[i]=x\land t[i+1]=y\}\right|$. Given a threshold $0\leq\noiseThr\leq1$, \cite{GlobalConcFilter} decides if two activities $x$ and $y$ are directly followed, i.e. $\df(x,y)=1$, iff the directly follows count $\dfCount(x,y)$ is larger than a threshold $\delta=\noiseThr *\left|\logL\right|/\left|\labels\right|$.

In this paper, we use a similar filter that defines the threshold $\delta$ for each directly follows relation based on a noise-level $\noiseThr$ and the frequency of the two activities involved, see the formula for $\delta$ below:

\begin{equation}
	\delta(x,y)=\noiseThr*(\sum_{z\in\labels\setminus x} \dfCount(x,z)+\sum_{z\in\labels\setminus y} \dfCount(z,y))/2
\end{equation}

$\delta$ takes into account the co-occurrence of the two activities before filtering the directly follow relation and hence will remove less activities than the filter proposed in~\cite{GlobalConcFilter}. Once the filtering of the directly follows is applied, a global concurrency oracle is used to compute the concurrency relations.

\subsubsection{Local oracles}
Global oracles can result too lax in the presence of duplicate activities. For example, a pair of activities can occur concurrently in some parts of the process, while in other parts they always occur in a given order. This can be observed in trace (1) and (2) of the running example in Fig.~\ref{fig:runningExampleLog}. The global oracle 
asserts that all three activities A,B and C are concurrent, because they were observed in any order in traces (3),(4) and (5). However, for traces (1) and (2) after executing X only the activities B and C are executed in any order, while A is always their predecessor. 
Hence, global oracles over-generalize the concurrency relations after executing activity X.

A \emph{local concurrency oracle}~\cite{Local-Concurrency} addresses this shortcoming by finding concurrency relations for pairs of activities at a given execution context. Then, $\conc$ requires three elements, two activity labels $x,y$ and an execution context $C$, i.e., $\conc(x,y,C) = 1$ if $x$ and $y$ are concurrent at $C$, and $\conc(x,y,C) = 0$ otherwise. For the sake of brevity, the details of the local oracle presented in~\cite{Local-Concurrency} are not shown in this paper, and the oracle is used as a blackbox.


\newpage
Consider the log of the running example in Fig.~\ref{fig:runningExampleLog}.
The local concurrency oracle identified two execution contexts: After the initial event, activities $A$,$B$ and $C$ are all concurrent. However, after executing $X$ only $B$ and $C$ are concurrent.
Hence, a local concurrency oracle can identify less over-generalizing concurrency relations. However, the technique presented in~\cite{Local-Concurrency} has been shown to be less scalable than global oracles and cannot handle large datasets.


In Section~\ref{sec:evaluation}, we use the alpha+ oracle~\cite{Alpha+-Concurrency} with a frequency filter as a representative global concurrency oracle and~\cite{Local-Concurrency} as a local concurrency oracle. 

\subsection{Petri nets}
Process models can be represented in various modelling languages, in this work we use Petri nets due to its well-defined execution semantics. This modelling language has two types of nodes, transitions, which in our case represent activities, and places, which represent execution states. The formal definition for Petri nets is presented below.

\begin{definition}[(Labelled) Petri net]
A (labelled) \emph{Petri net}, or simply a \emph{net}, is the tuple $\lnet = (\places, \netTransitions, \netArcs, \netLabel)$, where $\places$ and $\netTransitions$ are disjoint sets of \emph{nodes}, \emph{places} and \emph{transitions}, respectively; $\netArcs \subseteq (\places \times \netTransitions) \cup (\netTransitions \times \places)$ is the flow relation, and $\netLabel : \netTransitions \to \labels\cup\tau$ is a labelling function mapping transitions to labels $\labels \cup \{\tau\}$, where $\tau$ is a special label representing an unobservable action.
\end{definition}

Transitions with label $\tau$ represent silent steps whose execution leaves no footprint but that are necessary for capturing certain behavior in the net (e.g., optional execution of activities or loops). In a net, we will often refer to the preset or postset of a node, the preset of a node $y$ is the set $\inTr{y} = \{x \in P \cup T \mid (x,y) \in F\}$ and the postset of $y$ is the set $\outTr{y} = \{z \in P \cup T \mid (y,z) \in F\}$.

The work presented in this paper considers a sub-family of Petri nets: uniquely-labeled free-choice workflow nets~\cite{WFNets,FreeChoice}. In a uniquely labelled net, every label is assigned to at most one transition. Given that these nets are workflow and free choice nets, they have two special places: an initial and a final place and, whenever two transitions $t_1$ and $t_2$ share a common place $s \in \inTr{t_1} \cap \inTr{t_2}$, then all places in the preset are common for both transitions $\inTr{t_1} = \inTr{t_2}$. The formal definitions are given below.

\begin{definition}[Uniquely-Labelled, free-choice, workflow net]\label{def:wNet}
A (labelled) \emph{workflow net} is a triplet $\wNet=(\lnet,i,o)$, where $\lnet = (\places, \netTransitions, \netArcs, \netLabel)$ is a labelled Petri net, $i\in\places$ is the initial and $o\in\places$ is the final place, and the following properties hold:
\begin{compactitem}
    \item $i$ has an empty preset and $o$ has an empty postset, i.e., $\inTr{i}=\outTr{o}=\varnothing$, and
    \item if a transition $t^*$ were added from $o$ to $i$, such that $\inTr{i} = \outTr{o} = \{t^*\}$, then the resulting net is strongly connected.
\end{compactitem}
A workflow net $\wNet=(\lnet,i,o)$, where $\lnet = (\places, \netTransitions, \netArcs, \netLabel)$, is \emph{uniquely-labelled} and \emph{free-choice} if the following holds:
\begin{compactitem}
    \item (Uniquely-labelled) for any $t_1,t_2\in \netTransitions,\netLabel(t_1)=\netLabel(t_2)\neq\tau \Rightarrow t_1=t_2$, and 
    \item (Free-choice) for any $t_1,t_2 \in \netTransitions$: $s \in \inTr{t_1} \cap \inTr{t_2} \implies \inTr{t_1} = \inTr{t_2}$.
\end{compactitem}
\end{definition}

The execution semantics of a net can be defined by means of \emph{markings} representing its execution states and the \emph{firing rule} describing when an action can occur. 
A marking is a multiset of places, i.e. a function $m : P \rightarrow \mathbb{N}_0$ that relates each place $p \in P$ to a natural number of \emph{tokens}. A transition $t$ is enabled at marking $m$, represented as $m[t\rangle$, if each place of the preset $\inTr{t}$ contains a token, i.e. $\forall p \in \inTr{t} : m(p) \geq 1$. An enabled transition $t$ can fire to reach a new marking $m'$, the firing of $t$ removes a token from each place in the preset $\inTr{t}$ and adds a token to each place in the postset $\outTr{t}$, i.e. $m' = m \setminus \inTr{t} \uplus \outTr{t}$. A fired transition $t$ at a marking $m$ reaching a marking $m'$ is represented as $m[t\rangle m'$. 
A marking $m'$ is \emph{reachable} from $m$, if there exists a sequence of firing transitions $\sigma = \langle t_1,\dots t_n\rangle$, such that $m_{i-1}[t_i\rangle m_i$ holds for all $1\leq i\leq n$, $m_0=m$ and $m_n=m'$. In addition, every marking is reachable by itself.

A net with an initial and a final marking is called a \emph{(Petri) system net}.  
In the case of a workflow net, the initial marking has only one token in the special place $i$ and there is only one final marking with a token in the special place $o$.

\begin{definition}[System net]
A \emph{system net} $\sysNet$ is a triplet $\sysNet = (\wNet, \initMarking, \finalMarkings)$, where $\wNet=(\lnet,i,o)$ is a labelled workflow net, $\initMarking=\{i\}$ is the initial marking with the special place $i$ and $\finalMarkings=\{\{o\}\}$ is the set of final markings containing only one final marking with the special place $o$.
The universe of all system nets is denoted with $\universeSystemNets$.
\end{definition}

A marking is $k$-bounded if every place at a marking $m$ has up to $k$ tokens, i.e., $m(p) \leq k$ for any $p \in P$. 
A system net is $k$-bounded if every reachable marking in the net is $k$-bounded. 
Additionally, a system net is \emph{sound}~\cite{WFNets}, if:
\begin{inparaenum}[(1)]
	\item from any marking $m$ (reachable from $\initMarking$), it is possible to reach a final marking $m_f \in \finalMarkings$;
	\item when a final marking $m_f \in \finalMarkings$ is reached there are no tokens in any other places; and 
	\item each transition is enabled in at least one reachable marking.
\end{inparaenum}
Sound system nets that are 1-bounded are also called safe nets.
This article considers workflow system nets that are safe~\cite{FreeChoice}.

\begin{figure}[h!]
\centering
\includegraphics[width=0.66\textwidth]{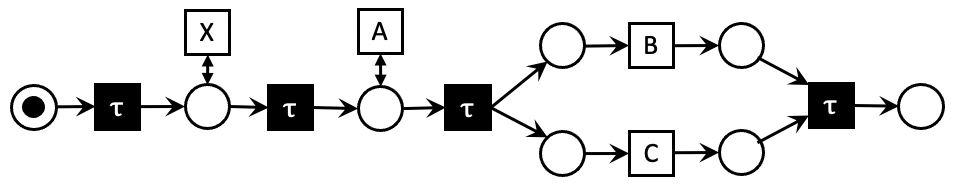}
\caption{System net of the running example.}\label{fig:runningExampleModel}
\end{figure}

The system net shown in Fig.~\ref{fig:runningExampleModel} is going to be used as the running example throughout the paper. The net contains several generalizations for the log of the running example shown in Fig.~\ref{fig:runningExampleLog}: activities $X$ and $A$ are repeatable and skippable, but $A$ can not be executed before $X$; Activities $B$ and $C$ are concurrent and can not occur before $X$ or $A$. 

\subsection{Alignments}

Alignments are sequences of steps, where each step relate an element in a log trace and/or an element in the model. 
Intuitively, alignments are used to find the closest execution of a log trace in a model.
There exist different definitions of alignment steps in the literature:
\begin{inparaenum}[(1)]
    \item The trace alignment technique~\cite{ILP-Alignment} uses Petri net representations of the model and the traces in the log. Each alignment step represents transitions from a synchronous net between a trace and the model.
    \item The automata technique~\cite{S-Components} represents the model and the log as automata. Then, each alignment step represents an operation over an arc in the log automaton and an arc in the model automaton.
\end{inparaenum}

In order to abstract our definition of an alignment step away from any specific definition and technique, we define an \emph{alignment step} as a tuple that has an activity label and one of three operations:
\begin{inparaenum}[(1)]
	\item $\lhide$ executes only the activity in the log trace,
	\item $\rhide$ executes only the activity in the model and
	\item $\match$ executes the activity in both model and log trace.
\end{inparaenum}
We define an alignment and alignment step below.


\begin{definition}[Alignment step, Alignment]
Let $\labels$ be a set of activity labels, an \emph{alignment step} is a pair $\sync=(\op,\ell)$, where $\op \in \{\match, \lhide, \rhide\}$ is an operation and $\ell\in\labels$ is an activity label.
An \emph{alignment} is a sequence of steps $\alignment = \langle \sync_1, \sync_2, \dots, \sync_n\rangle$.  
\end{definition}

Given an alignment step $\sync=(\op,\ell)$, the operation and the label are retrieved by $\op(\sync) = \op$ and $\lbl(\sync)=\ell$, respectively.
As a shorthand, functions $\op$ and $\lbl$ can also be used for an alignment by applying the functions to each step within.
For an alignment, we are sometimes interested in the subsequence of steps that represent the execution of the trace or the model.
For the execution of the trace, we define function $\filterSyncDFA(\alignment)$ that retrieves all steps with $\op(\sync)\neq\rhide$, while for the execution of the model, function $\filterSyncRG(\alignment)$ retrieves all steps with $\op(\sync)\neq\lhide$. 
An alignment $\alignment$ that represents complete executions of both a trace $t$ and a model is called \emph{proper}, i.e. it both fulfils $\lbl(\filterSyncDFA(\alignment))=t$ and $\lbl(\filterSyncRG(\alignment))$ represents an execution path through the Petri net (a firing sequence that reaches the final marking from the initial marking).

\newpage
Intuitively, an alignment represents the number of operations to transform a trace into an execution in the Petri net. 
Alignment steps can be associated with a cost, which can be defined individually for each activity with domain-specific knowledge. For simplicity purposes, we use the standard cost function~\cite{ILP-Alignment,ReissnerCDRA17}, where a weight of 1 is assigned to steps with $\lhide$ and $\rhide$ operations, and 0 to steps with $\match$ operations. The formal definition is given next. 


\begin{definition}[Cost function]\label{def:CostFunction}
Given an alignment $\alignment$ with $1\leq \position\leq\left|\alignment\right|$, we define the cost of an alignment step at position $i$ with function $\syncCostStandard$:
\[
\syncCostStandard(\alignment,\position)=
\begin{cases}
    1,& \text{if } \op(\alignment[\position])=\rhide \lor \op(\alignment[\position])=\lhide\\
    0,& \text{if } \op(\alignment[\position])=\match
\end{cases}
\]
The total cost $\costF$ for an alignment is the sum of function $\syncCostStandard$ for each of its elements:
\[
\costF(\alignment) = \sum_{\position\in1\dots\left|\alignment\right|}\syncCostStandard(\alignment,\position)
\]
\end{definition}
For computing alignments with domain-specific knowledge the cost function could be extended by using multipliers for each activity of the event log or process model.
\newcommand{\tandemOracle}{\Delta}
\newcommand{\posTandems}{\tandemOracle_{\pos}}
\newcommand{\pos}{\mathit{pos}}
\newcommand{\rt}{\mathit{rt}}
\newcommand{\extend}{\mathit{extend}}
\newcommand{\et}{\mathit{et}}
\newcommand{\extLog}{\mathit{EL}}
\newcommand{\compRepPartialFulfilment}{\mathit{computeRepetitivePartialFulfilment}}
\newcommand{\alignmentTraceOps}{\alignment_{tr,\op}}
\newcommand{\setPositions}{\mathit{Pos}}
\newcommand{\partialFulfilment}{\mathit{pf}}
\newcommand{\defineRepPattern}{\mathit{defineRepetitivePattern}}
\newcommand{\identifyRepPatterns}{\mathit{identifyRepetitivePatterns}}
\newcommand{\Patterns}{\mathcal{P}}
\newcommand{\pattern}{\mathit{p}}
\newcommand{\repPatterns}{\Patterns_{\mathit{rep}}}
\newcommand{\repPattern}{\pattern_{\mathit{rep}}}
\newcommand{\repPatternsSet}{\repPatterns}
\newcommand{\universeRepPatterns}{\mathcal{U}_{\Patterns,\mathit{rep}}}
\newcommand{\tracePos}{\mathit{Pos}_{
\mathit{t}}}
\newcommand{\traceCount}{\#\mathit{t}}
\newcommand{\concPattern}{\pattern_{\mathit{conc}}}
\newcommand{\concPatterns}{\Patterns_{\mathit{conc}}}
\newcommand{\repTraces}{\mathcal{T}_{\mathit{rep}}}
\newcommand{\compRepTraces}{\mathit{computeRepresentativeTraces}}
\newcommand{\concurrent}{\mathit{conc}}
\newcommand{\memory}{\mathit{mem}}
\newcommand{\patternGen}{\mathit{G}_{\mathit{pattern}}}
\newcommand{\powerSet}{\mathit{P}}
\newcommand{\outVertices}{\mathit{U}}
\newcommand{\compPartialFulfilment}{\mathit{computeFulfilmentWPartialMatching}}
\newcommand{\compInterleavingsFulfilment}{\mathit{computeFulfilmentWInterleavingsMatching}}
\newcommand{\usePartialMatching}{\mathit{usePM}}
\newcommand{\universeConcPatterns}{\mathcal{U}_{\Patterns,\mathit{conc}}}
\newcommand{\MT}{\mathit{MT}}
\newcommand{\LH}{\mathit{LH}}
\newcommand{\repPatternGen}{\mathit{G}_{\mathit{rep}}}
\newcommand{\concPatternGen}{\mathit{G}_{\mathit{conc}}}

\section{Pattern-based Generalization}\label{sec:approach}

This section presents a framework on how to compute generalization between an input event log and a process model. The measure is based on the idea that patterns in an event log induce behavior of the underlying process that is still unrecorded but likely to occur in future executions of the process. It is desirable for a process model to include control structures to generalize these patterns to avoid overfitting the event log. Since there exists a plethora of patterns in the event log that can be linked to control structures of a process model, we can define a framework of generalization measures based on the patterns selected to measure the generalization. In general, the more patterns are used to compute a generalization measure the more fine-grained it can distinguish process models, but the more unlikely it becomes the patterns will correspond to behavior of the underlying process. Also, with an increasing number of patterns the measure will scale less towards larger event logs and process models.

\begin{figure}[h]
\centering
\includegraphics[width=\textwidth]{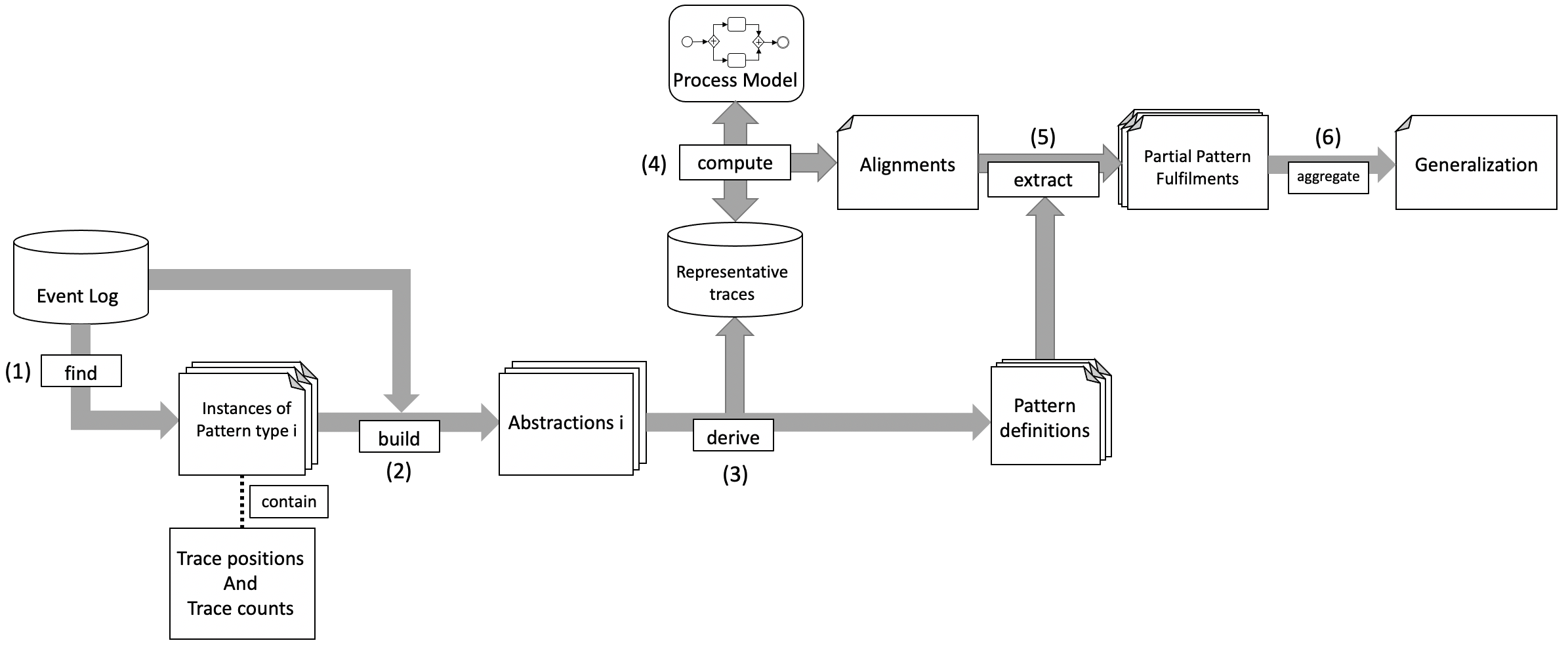}
\caption{Framework for computing pattern-based generalization measures} \label{fig:framework}
\end{figure}

Figure~\ref{fig:framework} shows the steps of the proposed framework to compute a pattern-based generalization measure. First, all instances of the selected pattern types are found in the input event log (1). Second, the pattern instances are used with the traces of the event log to build abstractions representing all possible behavior of the patterns (2). Third, a set of representative traces is derived from the abstractions and each pattern is defined in terms of traces positions within the representative traces (3). Fourth, the alignments are computed between the input process model and the set of representative traces (4). Next, partial fulfilments of each pattern are extracted by comparing the alignments with the definitions of the patterns (5).
Last, the generalization is computed by aggregating the partial pattern fulfilments to a single generalization value (6).

Every pattern type in a generalization measure can be instantiated with three steps:
\begin{itemize}
    \item All instances of a pattern type need to be identified acc. to some properties of the pattern, e.g. for repetitive patterns all tandem repeats can be identified.
    \item The instances then need to be grouped together to build an abstraction for the example behavior, e.g. all tandem repeats with the same repeating sequence but a different number of repetitions can be mapped to the same abstraction, a reduced trace.
    \item Finally, for each abstraction a set of representative traces needs to be derived that can be aligned to the process model. In addition, the trace positions need to be collected at which the pattern occurs in the representative traces. For example, a reduced trace can be extended to a large number of repetitions such that the model needs to align the repeat to a loop structure. 
\end{itemize}

While the framework can be extended with new patterns observed in the log, this section introduces a generalization based on two patterns: repetitive and concurrent behaviour. Intuitively, the repetitive and concurrent patterns observed in the log are traced to possible executions of the model, and then it is measured how much repetition and concurrency the model allows.
Three considerations are made in the presented generalization measure: 
\begin{itemize}
	\item Alignments~\cite{ILP-Alignment,TandemRepeats} are used to relate patterns in the log with executions of the model. 
	
	\item Patterns can be completely and partially fulfilled. To compute the measure, representative traces are derived from the patterns detected in the log, then such traces are aligned over the model and, if the pattern was fully aligned (only $\match$ operations were used), then the model is generalizing the pattern ``completely''. 
Complete fulfilment of patterns observed in the log can result too strict in some cases. For example, if in the log a repeating sequence of activities is observed $A,B,A,B$, and the model contains a cycle only with activity $A$, then the pattern should still be considered with a partial score during the computation of the generalization, even though the repetitive pattern is only partially fulfilled. 
Thus, for a more fine grained generalization measure, we propose to use the pattern generalization framework in composition with partial fulfilments of patterns instead of complete fulfilment. That is, the fraction of trace positions of the pattern that could be matched in alignments for its representative traces. 

	\item Finally, the frequency of occurrence of a pattern is considered when computing the measure, as patterns occurring in a larger number of traces are also more likely to represent behavior that also belongs to the underlying process. 
\end{itemize}

\newpage
\subsection{Pattern-based generalization with repetitive and concurrent patterns}\label{sec:measure}


This subsection presents both types of generalization considered: repetitive and concurrent patterns. The repetitive patterns computed from the log are based on tandem repeats; whereas the concurrent patterns are extracted from the transformation of a trace into a partial order using concurrency oracles. 


\begin{figure}[h]
\centering
\includegraphics[width=1\textwidth]{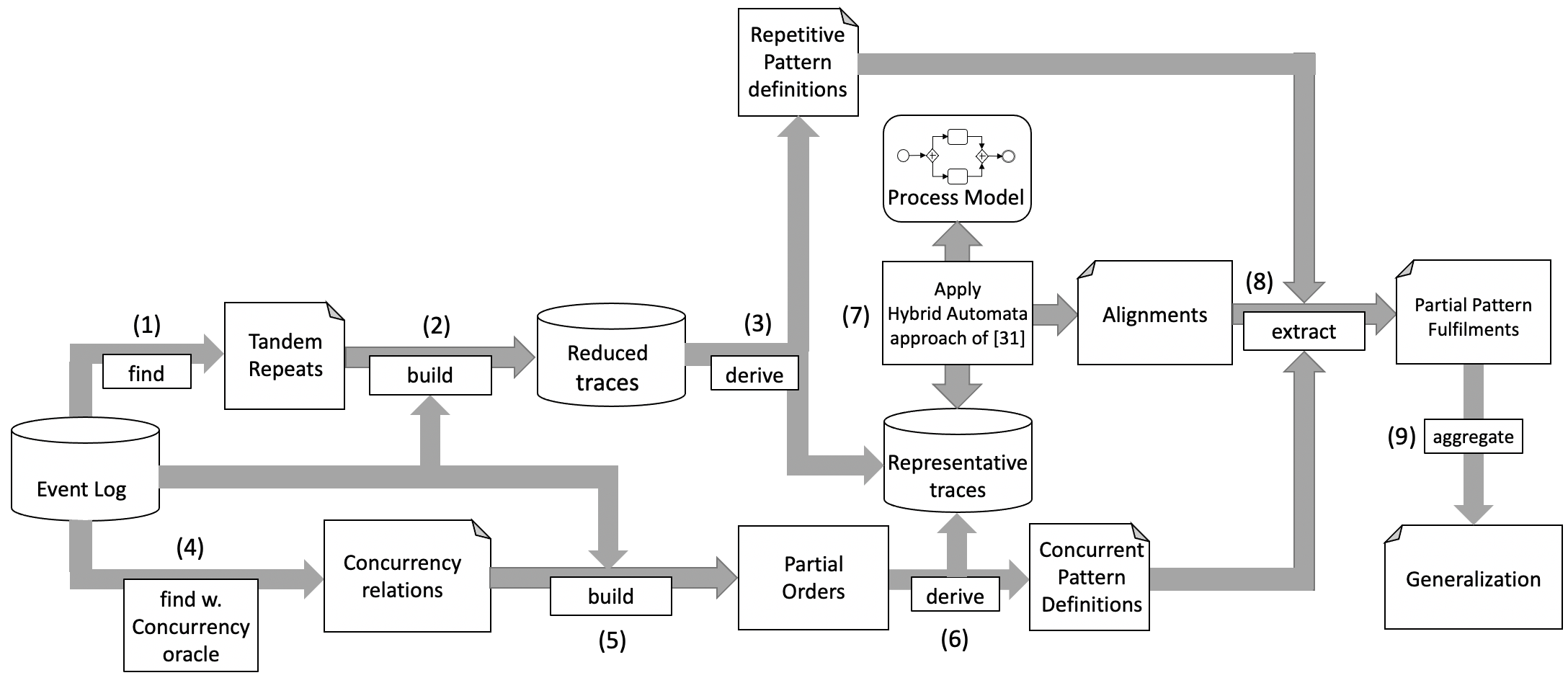}
\caption{Approach for instantiating our generalization framework into a measure based on repetitive and concurrent patterns} \label{fig:framework_implementation}
\end{figure}

Figure~\ref{fig:framework_implementation} provides an overview of the generalization measure based on repetitive and concurrent patterns.  
The identification and definition of repetitive and concurrent patterns can be carried out independently, i.e. steps (1),(2) and (3) can be executed at the same time as steps (4),(5) and (6).
For the repetitive patterns, first, all tandem repeats in the event log are found (1), and then each trace containing tandem repeats is transformed into a trace with only two copies of the repeating sequence (2). Traces with the same tandem repeat but with a different number of repetitions are reduced to the same unique ``reduced'' trace. These reduced traces are then expanded to derive representative traces by adding a large number of repetitions (3), in this way, alignments between a model and the traces will be forced to align the tandem repeats to cycles in the models. During the expansion, we also define the repetitive patterns with their repeating sequence and their trace positions. 
For the concurrent patterns, a concurrency oracle is used to find concurrency relations between pairs of activities (4). These concurrency relations are then used to build partial orders for the traces in the event log (5). We then derive a set of representative traces by generating all interleavings of each partial order and, at the same time, define the concurrent patterns denoting the concurrent activities and their trace positions (6). 

In step (7), we compute the alignments between the input process model and the representative traces for both the repetitive and concurrent patterns by applying the hybrid automata approach in~\cite{TandemRepeats}.
Then, we compare the alignments with the definitions of the patterns to extract partial pattern fulfilment (8).
Finally, the overall generalization is the weighted average of all partial fulfilments weighted with the trace counts of the corresponding patterns (9).

In the following two subsections, we will present the identification of each of the repetitive and concurrent patterns, as well as the measurements of their partial fulfilments. Subsection~\ref{sec:aggregateGeneralization} shows how to aggregate all pattern fulfilments into an overall generalization measure, and the last subsection compares the proposed pattern generalization against the axioms proposed in~\cite{GeneralizationAxioms}.

\newpage
\subsection{Identifying and measuring fulfilment of repetitive patterns}

The generalization involving repetitive patterns is based on the following assumption: sequences of activities observed repeatedly in the log represent repetitive behaviour that the underlying system can produce. Thus, in an ideal scenario, a process model completely generalizing the repetitive behaviour includes a loop structure that can replay the observed repetitive patterns in the log.
The proposed approach will consider also the cases when a process model partially fulfils a repetitive pattern. Partial fulfilment of repetitive patterns means that the process model contains a loop for a subset activities repeated in the log and it contributes a partial score to the generalization.

For the definition of the repetitive patterns, we rely on the identification of tandem repeats with an oracle $\Delta(t)$ as described in Section~\ref{sec:tandemRepeats}. 
Tandem repeats capture all repetitive behavior in an event log.
However, several tandem repeats can correspond to the same repeating sequence; for example, traces (8) and (9) in Fig.~\ref{fig:runningExampleLog} contain the same repeating sequence $XA$ with three and five repetitions, respectively. 
Hence, to consolidate the repetitive behaviour, in a first step we will reduce each tandem repeat to retain only two copies.
We will store these ``reduced traces'' with tandem repeats in a new \emph{reduced event log} $\redLog$.
Collapsing tandem repeats possibly ``folds'' several unique traces of the event log into a unique reduced trace; in the case of traces (8) and (9), they are reduced to $\langle X,A,X,A,C,B\rangle$. Note that, the trace count for the reduced trace will be the sum of the counts of the folded traces.

\begin{algorithm}[h]
{
    \caption{Reduce traces with tandem repeats}\label{alg:ReduceTandems}    
    \SetKwInOut{Input}{input}
    \SetKwProg{Fn}{Function}{}{}
    \Input{Event log $\logL$; Tandem repeat oracle $\tandemOracle(t)$;}
    $\redLog\leftarrow\{\}$\;\label{ln:begin_compRedLog}
    \For{$t\in\uniqueTraces(\logL)$}
    {
        \lIf{$\Delta(t)\neq\varnothing$}
        {
            $\redLog \leftarrow \redLog \multisetplus (\reduce(t),\countTr(t,\logL))$
        }
    }
    \Return $\redLog$\;\label{ln:end_compRedLog}
    \Fn{$\reduce(t)$}{
    $\rt\leftarrow\langle\rangle$;
    $\pos\leftarrow1$\;
    \While{$\pos\leq\left|t\right|$}
    {
        $\posTandems\leftarrow \{(\pos,\alpha,k)\in\tandemOracle(t)\}$\;
        \uIf{$\posTandems=\varnothing$}{
            $\rt\leftarrow\rt\concat t[\pos]$\;
            $\pos\leftarrow\pos+1$\;
        }
        \Else{
            $(i,\alpha,k)\leftarrow \max_{\left|\alpha\right|*k}\posTandems$\;
            $\rt\leftarrow\rt\concat t[i,i+\left|\alpha\right|*2]$\;
            $\pos\leftarrow \pos+\left|\alpha\right|*k$\;
        }
    }
    \Return{$\rt$}\;
    }
}
\end{algorithm}

The reduction of an event log is shown in Algorithm~\ref{alg:ReduceTandems}. Given an event log $\logL$ and a tandem repeat oracle $\tandemOracle$, Lines \ref{ln:begin_compRedLog}-\ref{ln:end_compRedLog} build a new event log $\redLog$, where unique traces with at least one tandem repeat are reduced with function $\reduce(t)$. 
The reduced traces are added to the reduced event log $\redLog$ with the trace count of their original corresponding trace. When two or more unique traces of the event log are reduced to the same trace, their trace counts will be added up in $\redLog$.
The function $\reduce(t)$ takes a trace $t$ as input and constructs a new reduced trace $\rt$. In $\reduce(t)$, variable $\pos$ is used as a pointer to the current position and it moves from left to right. 
If trace $t$ does not contain a tandem repeat at the current position $\pos$, $\rt$ adds the activity of $t$ at $\pos$ and moves the position by one.
Otherwise, the tandem repeat starting at the current position $\pos$ and with the highest length is selected, where the length is the size of the repeating sequence $\alpha$ times the repetitions $k$. 
At the end, the reduced trace $\rt$ contains two copies of the tandem repeat, and the trace position $\pos$ is moved to the position after the last repetition of the repeating sequence.

\begin{wrapfigure}{r}{0.45\textwidth}
\centering
\includegraphics[scale=.4]{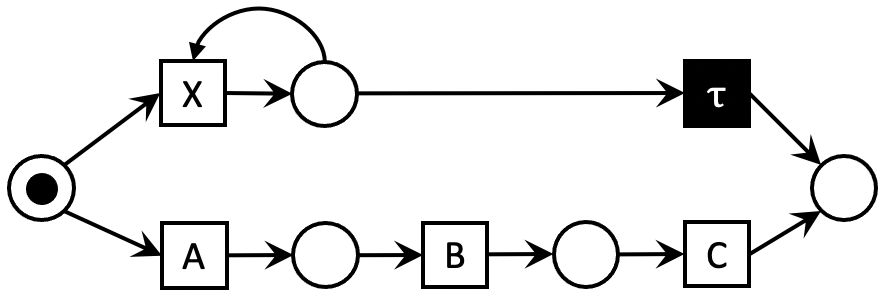}
\caption{System net with looping $x$.}\label{fig:ModelExtension}
\end{wrapfigure}

Alignments are used to find out if a process model contains a loop that can (partially or fully) emulate the detected tandem repeats. An alignment, however, will choose the path through the model with the least amount of mismatches ($\lhide$ and $\rhide$). In the context of the repetitive pattern, this may lead to the situation where the alignment skips the loop in the model, which is undesirable. Take for example the trace $\langle a,b,c,x,x\rangle$ and the system net in Fig.~\ref{fig:ModelExtension}, which allows to either execute activities $a$, $b$ and $c$ or traverse loop $x$. An alignment would choose to traverse activities $a,b,c$ in the model and skip activities $x,x$, because it can match more activities than going through the branch in the model with the cycle. However, we want to test whether the tandem repeat $x,x$ can be matched to a loop, in this case, $x$ in the model. 
Hence, in order to avoid that situation, a reduced trace is expanded by adding repetitions of each tandem repeat, such that the number of repetitions is as long as the reduced trace. Thus, any alignment will try to find an execution in the process model with a loop for the repeating sequence if possible, as it will outweigh all possible matches of the other activities in the trace. Then, in the example in Fig.~\ref{fig:ModelExtension}, the repetitions of tandem repeat $x,x$ is expanded to the length of the trace: $\langle a,b,c,x,x,x,x,x\rangle$. For the extended trace, an alignment would now choose to skip activities $a,b,c$ in the trace and match activities $x,x,x,x,x$ to the loop in the model. 

Given a reduced event log $\redLog$, Alg.~\ref{alg:ExtendTandems} shows how to compute an extended event log $\extLog$ with the extended versions of the unique reduced trace $\rt$. The extended traces will then be added to $\extLog$ with the corresponding trace counts of the reduced traces.
Function $\extend(\rt)$ creates an extended trace $\et$ for a reduced trace $\rt$. The extension uses $\pos$ to iterate over $\rt$. 
At each position $\pos$, if $\rt$ does not contain a tandem repeat, $\et$ will add the current activity of the reduced trace and move $\pos$ by one trace position.
If $\rt$ contains a tandem repeat at $\pos$, we choose the tandem repeat with the highest length.
We extend the tandem repeat in $\et$ by adding as many copies of the repeating sequence $\alpha$ as necessary (size of the reduced trace).
Then, the index $\pos$ is moved to the trace position after the tandem repeat of $\rt$.

\begin{algorithm}[h]
{
    \caption{Compute representative traces with extended tandem repeats}\label{alg:ExtendTandems}    
    \SetKwInOut{Input}{input}
    \SetKwProg{Fn}{Function}{}{}
    \Input{Reduced Event log $\redLog$; Tandem repeat oracle $\tandemOracle(t)$;}
    $\extLog\leftarrow\{\}$\;
    \For{$\rt\in\uniqueTraces(\redLog)$}
    {
        $\extLog\leftarrow\extLog\multisetplus(\extend(\rt),\countTr(\rt))$\;    
    }
    \Return{$\extLog$}\;
    \Fn{$\extend(\rt)$}
    {
        $\et\leftarrow\langle\rangle$;
        $\pos\leftarrow 1$\;
        \While{$\pos\leq\left|\rt\right|$}{        
            $\posTandems\leftarrow \{(pos,\alpha,k)\in\tandemOracle(\rt)\}$\;
            \uIf{$\posTandems=\varnothing$}{
            $\et\leftarrow\et\concat\rt[\pos]$\;
            $\pos\leftarrow\pos+1$\;
            }
            \Else{
                $(i,\alpha,k)\leftarrow \max_{\left|\alpha\right|*k}\posTandems$\;        
                \lFor{$1\leq k\leq\left|\rt\right|$}{
                $\et\leftarrow\et\concat\rt[\pos,\pos+\left|\alpha\right|-1]$
                }
                $\pos\leftarrow\pos+\left|\alpha\right|*2$\;
            }
        }    
        \Return{$\et$}\;    
    }
}
\end{algorithm}

\newpage
Based on the identified extended traces, we can now define the repetitive patterns of an event log, i.e., these are the extended tandem repeats.
In particular, given a tandem repeat, a repetitive pattern has the trace positions of the first tandem repeat repetition and the number of repetitions.
Its extended trace is the representative trace of the pattern. 
What is left to do is to determine how well the repetitive pattern is fulfilled by the system net. For that, we relate its representative trace to the system net by computing their alignment. 
An activity inside the repeating sequence is deemed as ``fulfilled'' by a loop in the system net if it was matched by the alignment in every repetition of the tandem repeat.

For the complete fulfilment of the pattern, a repetitive pattern is then completely fulfilled, if all labels of the repeating sequence of its tandem repeat can be matched in every repetition; whereas for the partial fulfilment of the repetitive pattern, it ranges from zero to one by finding the fraction of labels of the repeating sequence that can be matched in the alignment in every repetition of the tandem repeat. We only consider the partial fulfilment for the definition of a repetitive pattern since it allows for a more fine-grained analysis and includes the case of a completely fulfilled pattern. Finally, the trace count is included as the relative importance of the pattern. 
An extended trace can have more than one tandem repeat and hence it can also have several repetitive patterns. 
Next, we present the definition of a repetitive pattern and a function relating a trace to a set of its repetitive patterns.

\begin{definition}[Repetitive patterns]
Given a trace $\et$, a repetitive pattern $\repPattern$ is a touple $\repPattern=(\tracePos,k,\traceCount,\partialFulfilment)$, where $
\tracePos$ is a set of trace positions in $\et$ of the first repeating sequence, $k$ is the number of repetitions, $\traceCount$ is the trace count of the pattern and $\partialFulfilment$ is the partial fulfilment of the pattern. $\universeRepPatterns$ is the universe of all repetitive patterns.
Additionally, we define function $\repPatterns : \universeTraces\rightarrow\powerSet(\universeRepPatterns)$ that relates a trace $\et$ to a set of repetitive patterns, where $\powerSet$ denotes the power set of $\universeRepPatterns$.
\end{definition}

Given a repetitive pattern $\repPattern=(\tracePos,k,\traceCount,\partialFulfilment)$, we use auxiliary functions to access the properties of a repetitive pattern, $\tracePos(\repPattern)=\tracePos$, $k(\repPattern)=k$, $\traceCount(\repPattern)=\traceCount$ and $\partialFulfilment(\repPattern)=\partialFulfilment$.


\begin{algorithm}[!h]
{
    \caption{Identifying and defining repetitive patterns}\label{alg:DefineRepPatterns}    
    \SetKwInOut{Input}{input}
    \SetKwProg{Fn}{Function}{}{}
    \Input{Extended Event log $\extLog$; System net $\sysNet$; Tandem repeat oracle $\tandemOracle(\et)$;}
    $\alignments\leftarrow \text{align}(\extLog,\sysNet)$\;    
    $\repPatterns\leftarrow\{\}$\;    
    \For{$\et\in\uniqueTraces(\extLog)$}{
        $\pos\leftarrow 1$\;
        \While{$\pos\leq\left|\et\right|$}
        {
            $\posTandems\leftarrow(i,\alpha,k)\in\tandemOracle(\et)\mid i=\pos$\;
            \lIf{$\posTandems=\varnothing$}{
            $\pos\leftarrow\pos+1$
            }
            \Else{
                $(i,\alpha,k)\leftarrow \max_{\left|\alpha\right|*k}\posTandems$\;        
                $\repPatterns(\et)\leftarrow\repPatterns(\et)\cup \{\defineRepPattern(\et,\alignment,i,\alpha,k)\}$\;
                $\pos\leftarrow\pos+\left|\alpha\right|*k$\;
            }                
        }
    }
    \Return{$\repPatterns$}\;
    \Fn{$\defineRepPattern(\et,\alignment,i,\alpha,k)$}
    {
       $\alignmentTraceOps\leftarrow\op(\filterSyncDFA(\alignment))$\;
       $\partialFulfilment\leftarrow 0$;
       $\pos\leftarrow i$\;
       \While{$\pos\leq i+\left|\alpha\right|$}{
            $\setPositions\leftarrow\{\pos+j*\left|\alpha\right\|\mid 0\leq j< k\}$\;
           \lIf{$\forall_{j\in\setPositions} :\alignmentTraceOps[j]=\match$}{
               $\partialFulfilment\leftarrow\partialFulfilment+1$
            }
       }
       $\partialFulfilment\leftarrow\partialFulfilment/\left|\alpha\right|$\;
       $\repPattern\leftarrow(\{j\mid i\leq j<\left|\alpha\right|\},k,\countTr(\et,\extLog),\partialFulfilment)$\;
       \Return{$\repPattern$}\;    
    }
}
\end{algorithm}

Algorithm~\ref{alg:DefineRepPatterns} shows how to identify and define the repetitive patterns for an extended event log $\extLog$, a system net $\sysNet$ and a tandem repeat oracle $\tandemOracle(\et)$. 
First, the alignments between $\extLog$ and $\sysNet$ are computed with the algorithm described in~\cite{TandemRepeats}. Function $\alignments$ relates each trace $\et\in\extLog$ to its alignment.
Then, the algorithm builds the function $\repPatterns$ by identifying and relating a set of repetitive patterns to each each unique trace $\et$ and its alignment $\alignment$.
The repetitive patterns are gathered while parsing the extended unique trace $\et$ from left to right using $\pos$.
At every position, the tandem repeat ($i,\alpha,k$) that starts at the current position is selected if possible, such that $i=\pos$ with the highest length (length of the repeating sequence times the number of repetitions).
We then use function $\defineRepPattern$ to define the repetitive pattern. Afterwards, position $\pos$ moves to after the tandem repeat and continue until the trace is parsed and, finally, the set of collected patterns is returned.
Given the trace $\et$ and the tandem repeat $(i,\alpha,k)$, three parameters of a pattern are defined: the trace positions of the first repeating sequence, its number of repetitions and the trace count of $\et$. Thus, function $\defineRepPattern$ only needs to compute the partial fulfilment of the pattern.
As a first step, we filter the given alignment $\alignment$ to only retain alignment steps related to the trace, i.e. with $\lhide$ or $\match$ operations and then we only keep the operations of each alignment step in $\alignmentTraceOps$.
Each position of the repeating sequence is tested by checking if the operation in the alignment is a match for every repetition of the tandem repeat.
The partial fulfilment is then the fraction of positions of the first repeating sequence that fulfil the condition divided by the length of the repeating sequence.

Hence, given an event log, a system net and a tandem repeat oracle, we can identify its repetitive patterns and their partial fulfilment by
\begin{inparaenum}[(1)]
    \item reducing the traces with tandem repeats of the event log with Alg.~\ref{alg:ReduceTandems} to retrieve a reduced event log;
    \item extending the traces of the reduced log with Alg.~\ref{alg:ExtendTandems} to get an extended event log;
    \item and then applying Alg.~\ref{alg:DefineRepPatterns} to identify and define the repetitive patterns for the traces of the extended log.
\end{inparaenum}
We overload the notation of $\repPatterns$ to return a set of repetitive patterns for a given event log $\logL$ and system net $\sysNet$ after applying all algorithms, shorthanded as $\repPatterns(\logL,\sysNet)=\{\repPatterns(\et)\mid\et\in\extLog\}$.

Figure~\ref{fig:runningExampleRepPatterns} demonstrates the identified repetitive patterns and intermediate results when applying Algs.~\ref{alg:ReduceTandems},\ref{alg:ExtendTandems} and \ref{alg:DefineRepPatterns} to the running example. 
Additionally, Fig.~\ref{fig:runningExampleExtAlignments} shows the alignments for the extended traces of the running example linked through the trace identifiers.
The traces with identifiers ranging from one to five do not contain any tandem repeats and hence they are discarded when computing reduced traces. The reduced traces abbreviate repetitive sequences with tandem repeats. For example, $\langle X,A,X,A\rangle$ is abbreviated as $\langle (1,XA,2) \rangle$. Both traces (8) and (9) map to the same reduced trace. The extended traces show how each trace is extended significantly by expanding the tandem repeats; for instance, trace (7) is increased from trace length 9 to trace length 30. The incidental increase in complexity for computing alignments can be alleviated by applying the technique described in~\cite{TandemRepeats} 
that finds alignments for a trace with reduced tandem repeats and then extends the tandem repeats within the alignment.
If two unique traces map to the same reduced trace. their trace counts are added up (e.g., the trace count for traces (8) and (9) will be added up to 400 in the running example).
In total, the running example contains four repetitive patterns corresponding to the four tandem repeats from the extended traces. The two repetitive patterns in trace (6) have the highest weight with a total of 2000 and both achieve a perfect fulfilment since the process model contains both the corresponding cycles for activities $X$ and $A$. However, the process model can not fulfil the repetitive patterns fully from traces (8) and (9), where $X$ and $A$ are repeated alternately while in the process model $X$ can only be repeated before executing $A$. For example, in traces (8) and (9) only one of the two activities can be matched in all repetitions and hence the pattern can only achieve a partial fulfilment of 0.5.

\begin{figure*}[h!]
\centering
\resizebox{1\textwidth}{!}{ 
 \tikzstyle{ID} = [draw, rectangle, fill=white, align=center, minimum height=5mm, text width={width("$t  (8),(9)  t$")}, font=\footnotesize]
 \tikzstyle{REDTRACE} = [draw, rectangle, fill=white, align=left, minimum height=5mm, text width={width("$t  1,X,2 , 3,A,2 ,B,C t$")}, font=\footnotesize]
 \tikzstyle{EXTTRACE} = [draw, rectangle, fill=white, align=left, minimum height=5mm, text width={width("$t  1,X,6 , 7,A,6 ,B,C t$")}, font=\footnotesize]
 \tikzstyle{COUNT} = [draw, rectangle, fill=white, align=center, minimum height=5mm, text width={width("$t count(dt,L) t$")}, font=\footnotesize]
 \tikzstyle{REPPATTERNS} = [draw, rectangle, fill=white, align=left, minimum height=5mm, text width={width("$t   1 ,6,1000,1 ,  7 ,6,1000,1ttttt  t$")}, font=\footnotesize]
 \begin{tikzpicture}[>=stealth', node distance=-0.3pt]
  \node[ID] (id) {\bf{ID}};  
  \node[REDTRACE, right=of id] (redLog) {\bf{$\rt\in\uniqueTraces(\redLog)$}};
  \node[EXTTRACE, right=of redLog] (extlog){\bf{$\et\in\uniqueTraces(\extLog)$}};
  \node[COUNT, right=of extlog] (count){\bf{$\countTr(\et,\extLog)$}};
  \node[REPPATTERNS, right=of count] (repPatterns) {\bf{$\repPatterns(\et)$}};
  \node[ID, below=of id] (id6) {(6)};
  \node[REDTRACE, right=of id6] (redTr6) {$\langle (1,X,2),(3,A,2),B,C \rangle$};
  \node[EXTTRACE, right=of redTr6] (extTr6){$\langle (1,X,6),(7,A,6),B,C \rangle$};
  \node[COUNT, right= of extTr6] (count6) {1000};
  \node[REPPATTERNS, right=of count6] (repPatterns6){$\{(\{1\},6,1000,1),(\{7\},6,1000,1)\}$};
  \node[ID, below=of id6] (id7) {(7)};      
  \node[REDTRACE, right=of id7] (redTr7) {$\langle (1,XXA,2),X,B,C \rangle$};
  \node[EXTTRACE, right=of redTr7] (extTr7) {$\langle (1,XXA,9),X,B,C \rangle$};
  \node[COUNT, right= of extTr7] (count7) {500};
  \node[REPPATTERNS, right=of count7] (repPatterns7) {$\{(\{1,2,3\},9,500,0.\overline{66})\}$};
  \node[ID, below=of id7] (id89) {(8),(9)};  
  \node[REDTRACE, right=of id89] (redTr89) {$\langle (1,XA,2),C,B \rangle$};
  \node[EXTTRACE, right=of redTr89] (extTr89) {$\langle (1,XA,6),C,B \rangle$};
  \node[COUNT, right= of extTr89] (count89) {400};
  \node[REPPATTERNS, right=of count89] (repPatterns89) {$\{(\{1,2\},6,400,0.5)\}$};
  \end{tikzpicture}
  }
 \caption{Repetitive patterns of the running example.}\label{fig:runningExampleRepPatterns}
\end{figure*}

\begin{figure*}[h!]
\centering
\resizebox{0.8\textwidth}{!}{ 
 \tikzstyle{ID} = [draw, rectangle, fill=white, align=center, minimum height=5mm, text width={width("$t  (8),(9)  t$")}, font=\footnotesize]
 \tikzstyle{ALIGNMENT} = [draw, rectangle, fill=white, align=left, minimum height=5mm, text width={width("$t count(dt,L)ttttttttttttttttttttttttttttttttttttttttttttttttttttttttttttttttttttttttttttttttt t$")}, font=\footnotesize]
 \begin{tikzpicture}[>=stealth', node distance=-0.3pt]
  \node[ID] (id) {\bf{ID}};  
  \node[ALIGNMENT, right=of id, align=center] (alignment){$\textit{align}(\et,\sysNet)$};
  \node[ID, below=of id] (id6) {(6)};
  \node[ALIGNMENT,right=of id6] (alignment6) {$\langle(1,(\MT,X),6),(7,(\MT,A),(\MT,B),(\MT,C)\rangle$};
  \node[ID, below=of id6] (id7) {(7)};
  \node[ALIGNMENT, right=of id7] (alignment7) {$\langle(1,(\MT,X),(\MT,X),(\LH,A),8),(\MT,X),(\MT,X),(\MT,A),(\LH,X),(\MT,B),(\MT,C)\rangle$};
  \node[ID, below=of id7] (id89) {(8),(9)};
  \node[ALIGNMENT, right=of id89] (alignment89) {$\langle(1,(\MT,X),(\LH,A),5),(\MT,X),(\MT,A),(\MT,C),(\MT,B)\rangle$};
  \end{tikzpicture}
  }
 \caption{Alignments for the extended traces of the running example.}\label{fig:runningExampleExtAlignments}
\end{figure*}

\subsection{Identifying and measuring fulfilment of concurrent patterns}\label{sec:ConcPatternsFulfilment}

The second type of generalization is based on concurrent patterns. Concurrency present in a process is captured as interleavings in the event log, where concurrent activities appear in different orders in the traces. For computing the generalization, concurrency relations between events are extracted from event logs using a concurrency oracle. The concurrency relations over the events can be used to transform traces into partial orders that are aligned over the model. In order to consider the partial fulfilment of concurrent patterns, this section considers the case when only a subset of concurrent events observed in the log can be aligned with a subset of the concurrent activities.

The identification and definition of concurrent patterns uses a partial order representation for each of the traces in the log. In order to construct a partial order from a trace, a concurrency oracle (Sec.~\ref{sec:Concurrency}) is used.
Several unique traces in the log will be transformed into isomorphic partial orders, since a single execution of the process where activities are executed concurrently can be represented by different interleaving representations. 
In order to test the concurrency patterns using alignments, all the representative traces (linearizations) represented by a partial order are computed. 
A concurrent pattern will always occur at the same positions in all representative traces; hence, we define a concurrent pattern as a set of trace positions that is valid for all representative traces of a partial order. 
The concurrent patterns and auxiliary functions are defined next.

\begin{definition}[Concurrent patterns, Representative traces]
Given a partial order $\po$, a concurrent pattern $\concPattern$ is a tuple $\concPattern=(\tracePos,\traceCount,\partialFulfilment)$, where $
\tracePos$ is a set of trace positions of the representative traces of $\po$, $\traceCount$ is the trace count of the pattern and $\partialFulfilment$ is the partial fulfilment of the pattern. $\universeConcPatterns$ denotes the universe of concurrent patterns.
Additionally, we define functions $\concPatterns : \universePO\rightarrow\powerSet(\universeConcPatterns)$ and $\repTraces : \universePO\rightarrow\universeTraces$ that relate a partial order $\po$ to its set of concurrent patterns and representative traces, respectively.
\end{definition}

The representative traces can be computed from a partial order using a breadth-first search traversal strategy. 
The breadth-first search will traverse the partial order and collect events along the causality relations to iteratively construct its representative traces. 

From the implementation point of view, the traversal is applied over the transitive reduction of the partial order. When an event has multiple outgoing events (concurrent events), the traversal will generate a trace for each of the outgoing events to simulate the different orders in which the events can occur.
When encountering concurrent events, the traversal explores one outgoing event and keeps all other concurrent events in a memory. In every iteration, the search will then consider the outgoing events for the event being visited, as well as the events in the memory.
In order to consolidate the concurrent paths, an event with multiple incoming events can only be visited by the traversal, if all its incoming events have been visited, i.e. all concurrent events have been traversed.
That way, the search can enumerate all orders of events in different traces. The traces are complete once the traversal reaches the final event of the partial order.

In order to construct the concurrent patters during the traversal, we start a new concurrent pattern when we encounter an event with multiple outgoing events that are concurrent among them. The pattern will add the trace positions whenever the search adds concurrent events to the constructed trace until the search traverses an event with multiple incoming events and the memory is empty. In that case, the search completes the definition of the concurrent pattern as the set of recorded trace positions and the trace count of the partial order. The partial fulfilment of the pattern can only be determined after all representative traces for the partial order have been computed. 

\begin{algorithm}[h!]
{
    \caption{Compute representative traces and concurrent patterns for partial orders}\label{alg:compRepTracesAndConcPatterns}    
    \SetKwInOut{Input}{input}
    \SetKwProg{Fn}{Function}{}{}
    \Input{Event log $\logL$; Concurrency oracle $\concOracle=(\conc,\xi)$;}
    $\repTraces\leftarrow\{\}$;
    $\concPatterns\leftarrow\{\}$\;
    \For{$\po=(\Events,\Causality,\poLabelling)\in\uniqueTraces(\xi(\logL))$}
    {      
        $\open\leftarrow\langle(\initialEvent,\langle\rangle,\{\},\{\},0)\rangle$\;
        \While{$\open\neq\varnothing$}{        
            $(v,t,\memory_v,\tracePos^v,\concurrent_v)\leftarrow\head(\open)$;
            $\open\leftarrow\tail(\open)$\;
            \lIf{$\left|(v,u)\in\Causality\right|>1$}{$\concurrent_v\leftarrow 1$}
            $\outVertices\leftarrow\{u\mid(v,u)\in\Causality\lor u\in\memory_v\}$\;
            \For{$u\in\outVertices$}{
                \If{$\forall (w,u)\in\Causality : w\in t$}{
                    \lIf{$u=\finalEvent$}{$\repTraces(\po)\leftarrow\repTraces(\po)\cup\{\poLabelling(t)\}$}
                    \Else
                    {
                        $n_u\leftarrow(u,t_u\leftarrow t\concat u,\memory_u\leftarrow\memory_v\setminus\{u\},\tracePos^u\leftarrow\tracePos^v,\concurrent_u\leftarrow\concurrent_v)$\;
                        \If{$\left|(w,u)\in\Causality\right|>1\land\memory_u=\varnothing\land\tracePos^v\neq\varnothing$}{
                            $\concPatterns(\po)\leftarrow\concPatterns(\po)\cup\{\tracePos^v,\countTr(\po,\xi(\logL)),\perp\}$\;
                            $\tracePos^u\leftarrow\{\}$\;
                            $\concurrent_u\leftarrow 0$\;                    
                        }
                        \lIf{$\left|(v,w)\in\Causality\right|>1$}{
                            $\memory_u\leftarrow\memory_u\cup\{w\mid (v,w)\in\Causality\land w\neq u\land (x,w)\in\Causality\Rightarrow x\in t\}$                        
                        }
                        \lIf{$\concurrent_u=1$}{
                               $\tracePos^u\leftarrow\tracePos^u\cup\left|t_u\right|$                 
                        }    
                        $\open\leftarrow\open\concat n_u$\; 
                    }
                } 
            }
        }      
    }
    \Return{$\repTraces,\concPatterns$}\;
}
\end{algorithm}

Given an event log and a concurrency oracle, Alg.~\ref{alg:compRepTracesAndConcPatterns} computes functions relating each unique partial order of the event log to its set of representative traces and its set of concurrent patterns.
The breadth first search traversal is conducted for each partial order via an open list $\open$ that stores nodes of the search as a quintuple $(v,t,\memory_v,\tracePos^v,\concurrent_v)$, which contains the current event $v$, the currently constructed trace of events $t$, the memory $\memory_v$ of a set of concurrent events that still need to be traversed, the trace positions of the current concurrent pattern $\tracePos^v$ and a boolean variable $\concurrent_v$ indicating if the current event is concurrent.
In every iteration of the search, the first node of the search is removed.
If the current event has multiple outgoing events, the search from the current node is set to concurrent, i.e. $\concurrent_v=1$.
Then, the search considers to traverse every event $u$ in the outgoing events of $v$ and from memory $\memory_v$, but only if every incoming event $w$ of $u$ is already traversed, i.e. $w\in t$.
If $u$ is the final event, the search will add the sequence of activity labels of all events of trace $t$ to the set of representative traces of the partial order and continue the search.
Otherwise, a new node for the search is created, where $u$ is the current event, trace $t$ adds event $u$, the memory carries over all events from $\memory_v$ besides $u$ and both $\tracePos^v$ and $\concurrent_v$ are carried over.
If $u$ closes a concurrent block of events, the memory is empty and the set of trace positions is not empty, then we will add the concurrent pattern tracked with $\tracePos^v$ to the set of concurrent patterns of the partial order. Furthermore, the set of trace positions of the new node is emptied and $\concurrent_u$ is set to zero.
If the current event $v$ has multiple outgoing events, then the memory of the new node $\memory_u$ is extended by all outgoing events $w$ of $v$ that are not $u$ and for which all their incoming events are included in the trace $t$.
If the current node is concurrent, i.e. $\concurrent_v=1$, then we extend the set of trace positions $\tracePos^u$ by the position of the currently added event $u$, i.e. $\left|t_u\right|$.
Finally, the new node is added to the open list and the search continues until all possible orders of the partial order have been traversed and all representative traces, as well as all concurrent patterns, have been gathered.

\newpage
The partial fulfilment of the identified patterns can only be determined after all representative traces for a partial order have been determined since the fulfilment relies on the alignments of the representative traces. 
In this article, we propose two ways of measuring the partial fulfilment of a concurrent pattern: one based on \emph{interleavings matching} and one based on \emph{partial matching}. 
The interleavings matching measures how many orders of the concurrent activities, a.k.a. interleavings, can be fully matched by the process model. 
In particular, every ordering of the concurrent activities is captured in one of the representative traces and hence the fulfilment can be determined as the fraction of alignments that can align all trace positions of the concurrent patterns with matches.
The partial matching is a more lenient method and does not require full matches for each order of concurrent activities. Instead, it counts the number of matches at the trace positions of the concurrent pattern for all representative traces and divide it by the overall number of trace positions of all traces.
The choice of matching method for concurrent patterns depends on the user's preferences of how strictly a concurrent pattern should be matched. 
Algorithm~\ref{alg:compPartialFulfilmentConcurrentPatterns} shows the computation of the partial fulfilments for all concurrent patterns. The algorithm allows to select the matching method for computing partial fulfilments with the boolean $\usePartialMatching$. 

\begin{algorithm}[h]
{
    \caption{Compute partial fulfilments of concurrent patterns}\label{alg:compPartialFulfilmentConcurrentPatterns}    
    \SetKwInOut{Input}{input}
    \SetKwProg{Fn}{Function}{}{}
    \Input{Event log $\logL$; System net $\sysNet$; Concurrency oracle $\concOracle=(\conc,\xi)$; Representative Traces $\repTraces$; Concurrent pattern definitions $\concPatterns$; Use partial matching $\usePartialMatching$;}
    \For{$\po\in\uniqueTraces(\xi(\logL)$)}{
        $\alignments\leftarrow \text{align}(\repTraces(\po),\sysNet)$\;
        \For{$\concPattern\in\concPatterns(\po)$}{
            \uIf{$\usePartialMatching=1$}{
                 $\partialFulfilment(\concPattern)\leftarrow\compPartialFulfilment(\concPattern,\alignments)$\;
             }
             \Else{
                 $\partialFulfilment(\concPattern)\leftarrow\compInterleavingsFulfilment(\concPattern,\alignments)$\; 
             }       
        }    
    }
    \Return{$\concPatterns$}\;
    \Fn{$\compPartialFulfilment(\concPattern,\alignments)$}
    {
        $\partialFulfilment\leftarrow 0$\;
        \For{$\alignment\in\alignments$}{        
            $\alignmentTraceOps\leftarrow\op(\filterSyncDFA(\alignment))$\;
            $\partialFulfilment\leftarrow\partialFulfilment+\left|\{\pos\in\tracePos(\concPattern)\mid\alignmentTraceOps[\pos]=\match\}\right|$\;  
        }
        \Return{$\partialFulfilment/(\left|\tracePos(\concPattern)\right|*\left|\alignments\right|)$}\;    
    }
    \Fn{$\compInterleavingsFulfilment(\concPattern,\alignments)$}{
        $\partialFulfilment\leftarrow 0$\;
        \For{$\alignment\in\alignments$}{        
            $\alignmentTraceOps\leftarrow\op(\filterSyncDFA(\alignment))$\;
            \lIf{$\forall \pos\in\tracePos(\concPattern) : \alignmentTraceOps[\pos]=\match)$}{
                $\partialFulfilment\leftarrow\partialFulfilment+1$            
            } 
        }
        \Return{$\partialFulfilment/\left|\alignments\right|$}\; 
    }
}
\end{algorithm}

In summary, given an event log, a system net, a concurrency oracle and the option of a matching method, the concurrent patterns can be defined by:
\begin{inparaenum}[(1)]
    \item Building the partial orders for the unique traces of an event log via the concurrency oracle;
    \item Identifying the representative traces and concurrent patterns of each partial order with a breadth-first search traversal (Alg.~\ref{alg:compRepTracesAndConcPatterns});
    \item Computing the partial fulfilments of all concurrent patterns based on the alignments of the representative traces and the system net via the selected matching method in Alg.~\ref{alg:compPartialFulfilmentConcurrentPatterns}.
\end{inparaenum}
We overload the notation of $\concPatterns$ to return a set of concurrent patterns for a given event log $\logL$ and a system net $\sysNet$ after applying all steps, i.e. $\concPatterns(\logL,\sysNet)=\{\repPatterns(\et)\mid\et\in\extLog\}$.

\begin{figure*}[h!]
\centering
\resizebox{1\textwidth}{!}{ 
 \tikzstyle{IDHL} = [draw, rectangle, fill=white, align=center, minimum height=5.21mm, text width={width("$t  (8),(9)  t$")}, font=\footnotesize]
 \tikzstyle{ID1} = [draw, rectangle, fill=white, align=center, minimum height=10.3mm, text width={width("$t  (8),(9)  t$")}, font=\footnotesize]
 \tikzstyle{ID2} = [draw, rectangle, fill=white, align=center, minimum height=31.1mm, text width={width("$t  (8),(9)  t$")}, font=\footnotesize]
 \tikzstyle{REPTRACE} = [draw, rectangle, fill=white, align=left, minimum height=5mm, text width={width("$t X,A,B,C t$")}, font=\footnotesize]
 \tikzstyle{REPALIGNMENT} = [draw, rectangle, fill=white, align=left, minimum height=5mm, text width={width("$t MT(X),MT(A),MT(B),MT(C)$")}, font=\footnotesize]
 \tikzstyle{CONCPATTERNHL} = [draw, rectangle, fill=white, align=left, minimum height=5mm, text width={width("$pf(pco)w.IMt$")}, font=\footnotesize]
 \tikzstyle{CONCPATTERN1} = [draw, rectangle, fill=white, align=center, minimum height=10.3mm, text width={width("$pf(pco)w.IMt$")}, font=\footnotesize]
 \tikzstyle{CONCPATTERN2} = [draw, rectangle, fill=white, align=center, minimum height=31.1mm, text width={width("$pf(pco)w.IMt$")}, font=\footnotesize]
 \begin{tikzpicture}[>=stealth', node distance=-0.3pt]
  \node[IDHL] (id) {ID};  
  \node[REPTRACE, right=of id] (repTraces) {\bf{$t\in\repTraces(\po)$}};
  \node[REPALIGNMENT, right=of repTraces] (repAlignments){\bf{$\alignment\in\text{align}(\repTraces(\po),\sysNet)$}};
  \node[CONCPATTERNHL, right=of repAlignments] (concPatternsTracePos){\bf{$\tracePos(\concPattern)$}};
  \node[CONCPATTERNHL, right=of concPatternsTracePos] (concPatternsSize){\bf{$\traceCount(\concPattern)$}};
  \node[CONCPATTERNHL, right=of concPatternsSize] (concPatternsPFIM){$\partialFulfilment(\concPattern)\text{ w.IMt}$};
  \node[CONCPATTERNHL, right=of concPatternsPFIM] (concPatternsPFPM){$\partialFulfilment(\concPattern)\text{ w.PMt}$};
  \node[ID1,below=of id] (id12) {(1),(2)};
  \node[ID2, below=of id12] (id345) {(3),(4),(5)};
  \node[REPTRACE, below=of repTraces] (repTrace1) {$\langle X,A,B,C \rangle$};
  \node[REPTRACE, below=of repTrace1] (repTrace2) {$\langle X,A,C,B \rangle$};
  \node[REPTRACE, below=of repTrace2] (repTrace3) {$\langle A,B,C \rangle$};
  \node[REPTRACE, below=of repTrace3] (repTrace4) {$\langle A,C,B \rangle$};
  \node[REPTRACE, below=of repTrace4] (repTrace5) {$\langle B,A,C \rangle$};
  \node[REPTRACE, below=of repTrace5] (repTrace6) {$\langle B,C,A \rangle$};
  \node[REPTRACE, below=of repTrace6] (repTrace7) {$\langle C,A,B \rangle$};
  \node[REPTRACE, below=of repTrace7] (repTrace8) {$\langle C,B,A \rangle$};
  \node[REPALIGNMENT,below=of repAlignments] (repAlignment1) {$\langle \MT(X),\MT(A),\MT(B),\MT(C)\rangle$};
  \node[REPALIGNMENT,below=of repAlignment1] (repAlignment2) {$\langle \MT(X),\MT(A),\MT(C),\MT(B)\rangle$};
  \node[REPALIGNMENT,below=of repAlignment2] (repAlignment3) {$\langle \MT(A),\MT(B),\MT(C)\rangle$};
  \node[REPALIGNMENT,below=of repAlignment3] (repAlignment4) {$\langle \MT(A),\MT(C),\MT(B)\rangle$};
  \node[REPALIGNMENT,below=of repAlignment4] (repAlignment5) {$\langle \MT(B),\LH(A),\MT(C)\rangle$};
  \node[REPALIGNMENT,below=of repAlignment5] (repAlignment6) {$\langle \MT(B),\MT(C),\LH(A)\rangle$};
  \node[REPALIGNMENT,below=of repAlignment6] (repAlignment7) {$\langle \MT(C),\LH(A),\MT(B)\rangle$};
  \node[REPALIGNMENT,below=of repAlignment7] (repAlignment8) {$\langle \MT(C),\MT(B),\LH(A)\rangle$};
  \node[CONCPATTERN1, below=of concPatternsTracePos] (concPatternsTracePos1){$\{3,4\}$};
  \node[CONCPATTERN2, below=of concPatternsTracePos1] (concPatternsTracePos2){$\{1,2,3\}$};
  \node[CONCPATTERN1, right=of concPatternsTracePos1] (concPatternsTraceCount1){$2000$};
  \node[CONCPATTERN2, right=of concPatternsTracePos2] (concPatternsTraceCount2){$600$};
  \node[CONCPATTERN1, right=of concPatternsTraceCount1] (concPatternsPFIM1){$2/2$};
  \node[CONCPATTERN2, right=of concPatternsTraceCount2] (concPatternsPFIM2){$2/6$};
  \node[CONCPATTERN1, right=of concPatternsPFIM1] (concPatternsPFPM1){$4/4$};
  \node[CONCPATTERN2, right=of concPatternsPFIM2] (concPatternsPFPM2){$14/18$};
  \end{tikzpicture}
  }
 \caption{Representative traces and concurrent patterns for the log of running example from Fig.~\ref{fig:runningExampleLog} with the local oracle.}\label{fig:runningExampleConcPatterns}
\end{figure*}

\newpage
Figure~\ref{fig:runningExampleConcPatterns} shows the representative traces and concurrent patterns for the running example (Fig.~\ref{fig:runningExampleLog}) with the local oracle. 
The traces with identifiers (1) and (2) map to the same partial order, while traces (3),(4) and (5) map to another partial order.
When applying Alg.~\ref{alg:compRepTracesAndConcPatterns} to their partial orders, we can derive two and six representative traces $\repTraces(\po)$ that enumerate all interleavings of concurrent activities $\{B, C\}$ and $\{A, B, C\}$, respectively.
The third column shows the alignments for the representative traces with the system net $\sysNet$ in Fig.~\ref{fig:runningExampleModel}. 
The algorithm identifies one concurrent pattern for each partial order, spanning trace positions 3,4 or 1,2 and 3 
of all corresponding representative traces and having trace counts of 2000 and 600, which is the sum of the trace counts of their represented traces.
Next, we take a closer look at the partial fulfilment of the concurrent pattern representing traces (3),(4) and (5).
Applying Alg.~\ref{alg:compPartialFulfilmentConcurrentPatterns} with interleavings matching (short w.IMt), the concurrent pattern can match all three concurrent activities in only two out of 6 alignments.
For partial matching (short w.PMt), the concurrent pattern can match in total 14 out of 18 alignment steps corresponding to the three positions of the pattern over all six of the alignments.
Both matching methods identify that the system net does not model the three concurrent activities with a parallel block. 
Interleavings matching assigns a lower score of fulfilment since only two orders can be fulfilled by the model, while the partial matching rewards the fact that activities $B$ and $C$ are concurrent in the model.

\subsection{Aggregating pattern fulfilments into a generalization measure}\label{sec:aggregateGeneralization}

Once the partial fulfilments of all repetitive and concurrent patterns are identified, the fulfilment scores are aggregated into a single generalization value. 

The pattern generalization is the average of pattern fulfilments weighted by their trace counts. That way, patterns that occur more frequently in the log will have a higher influence on the generalization value. To ease the formalization, we introduce a superset $\Patterns(\logL,\sysNet)$ that combines all repetitive and concurrent patterns of log $\logL$ and system net $\sysNet$, i.e. $\Patterns(\logL,\sysNet)=\repPatterns(\logL,\sysNet)\cup\concPatterns(\logL,\sysNet)$. We can then compute the pattern generalization $\patternGen$ for a log and a system net as follows:

\begin{equation}
    \patternGen(\logL,\sysNet)=\frac{\sum_{\pattern\in\Patterns(\logL,\sysNet)} \partialFulfilment(\pattern)*\countTr(\pattern)}{\sum_{\pattern\in\Patterns(\logL,\sysNet)} \countTr(\pattern)}
\end{equation}

The generalization measure $\patternGen$ has values between zero and one, with zero indicating that the system net does not generalize any patterns observed in the event log and one indicating that all patterns were properly  generalized in the system net. 
A value of one (perfect generalization) is assigned when no patterns can be identified in the event log since no generalization requirements need to be fulfilled by the model, i.e. $\patternGen(\logL,\sysNet)=1$ if $\Patterns(\logL,\sysNet)=\varnothing$.

\newpage
For analysis purposes, the generalization value can be sliced by each pattern type, one generalization value for repetitive patterns $\repPatternGen: \universeLogs\times\universeSystemNets\rightarrow[0,1]$ and one for concurrent patterns $\concPatternGen: \universeLogs\times\universeSystemNets\rightarrow[0,1]$. Each value can be determined as the weighted average of pattern fulfilments and their trace counts for only the set of corresponding patterns, i.e. $\repPatterns(\logL,\sysNet)$ and $\concPatterns(\logL,\sysNet)$. The two measures can give more insight into the strengths or weaknesses of a process model to adapt certain generalizing structures. The following equations show how to compute the generalization of each pattern type:

\begin{equation}
    \repPatternGen(\logL,\sysNet)=\frac{\sum_{\pattern\in\repPatterns(\logL,\sysNet)} \partialFulfilment(\pattern)*\countTr(\pattern)}{\sum_{\pattern\in\repPatterns(\logL,\sysNet)} \countTr(\pattern)}
\end{equation}

\begin{equation}
    \concPatternGen(\logL,\sysNet)=\frac{\sum_{\pattern\in\concPatterns(\logL,\sysNet)} \partialFulfilment(\pattern)*\countTr(\pattern)}{\sum_{\pattern\in\concPatterns(\logL,\sysNet)} \countTr(\pattern)}
\end{equation}

The pattern generalization of the running example for the log from Fig.~\ref{fig:runningExampleLog} and the system net from Fig.~\ref{fig:runningExampleModel} is $0.8\overline{60}$ for a total trace weight of 5.500 for all patterns using a local concurrency oracle and interleavings matching. The repetitive patterns generalization is 0.873 for a trace count of 2900 patterns and the concurrent patterns generalization is 0.846 for a trace count of 2600 patterns. The process model generalizes the patterns well by including both repeating and concurrent structures. The generalization could still be improved by either letting activities $X$ and $A$ be repeatable in an alternating way or by including activity $A$ in a parallel block with activities $B$ and $C$.


\subsection{Assessing the pattern generalization measure against generalization propositions}
The authors of \cite{GeneralizationAxioms} define various propositions that the quality measures of fitness, precision and generalization should satisfy. 
In the case of the generalization measure, it applies to a measure that computes values between zero and one, and determines the probability that a model can fit new (unseen) traces from the underlying process generating an event log. 
Our proposed pattern-based generalization measure fits this description to a certain degree. On the one hand, our pattern-based generalization determines new traces from the patterns observed in the event log and compares these traces against the model to determine its ability to parse these new traces. 
On the other hand, instead of just considering new and unseen traces, the pattern-based generalization checks all representative traces of the patterns against the model's constructs (such as loops or parallel blocks).
As such, the pattern-based generalization measure is focused on identifying improvements to the ability of a process model to generalize the event log by finding unfulfilled structures that still may be included in the process model.
Despite the slightly different definition, we will compare the proposed measure against the ten propositions in~\cite{GeneralizationAxioms} in an informal way. 
Each proposition is denoted with a superscript of either ``+'' showing that the proposition is mandatory and accepted in the community or with ``0'' labelling the proposition as controversial. 
Figure~\ref{fig:propositionFulfilment} shows the fulfilment of propositions by the pattern-based generalization, \cmark ~denotes that a proposition is always fulfilled for any log and model, whereas \xmark ~denotes that a proposition does not always hold. Below, we investigate the fulfilment of each proposition.

\begin{figure*}[htbp]
\centering
\resizebox{1\textwidth}{!}{ 
 \tikzstyle{PROPOSITION} = [draw, rectangle, fill=white, align=center, minimum height=7.5mm, text width={width("$t Proposition t$")}, font=\large]
 \tikzstyle{FULFILMENT} = [draw, rectangle, fill=white, align=center, minimum height=5mm, text width={width("$GPATT$")}, font=\large]
 \begin{tikzpicture}[>=stealth', node distance=-0.3pt]
  \node[PROPOSITION] (PropHL) {\bf{Prop.:}};
  \node[PROPOSITION, below=of PropHL] (FulHL) {$\patternGen$};
  \node[PROPOSITION, right=of PropHL] (Prop1){1:DetPro\textsuperscript{+}};
  \node[PROPOSITION, right=of FulHL] (Ful1){\cmark};
  \node[PROPOSITION,right=of Prop1] (Prop2){2:BehPro\textsuperscript{+}};
  \node[PROPOSITION,right=of Ful1] (Ful2){\cmark};
  \node[PROPOSITION,right=of Prop2] (Prop3){GenPro1\textsuperscript{+}};
  \node[PROPOSITION,right=of Ful2] (Ful3){\cmark};
  \node[PROPOSITION,right=of Prop3] (Prop4){GenPro2\textsuperscript{+}};
  \node[PROPOSITION,right=of Ful3] (Ful4){\xmark};
  \node[PROPOSITION,right=of Prop4] (Prop5){GenPro3\textsuperscript{0}};
  \node[PROPOSITION,right=of Ful4] (Ful5){\cmark};
  \node[PROPOSITION,right=of Prop5] (Prop6){GenPro4\textsuperscript{+}};
  \node[PROPOSITION,right=of Ful5] (Ful6){\cmark};
  \node[PROPOSITION,right=of Prop6] (Prop7){GenPro5\textsuperscript{+}};
  \node[PROPOSITION,right=of Ful6] (Ful7){\cmark};
  \node[PROPOSITION,right=of Prop7] (Prop8){GenPro6\textsuperscript{0}};
  \node[PROPOSITION,right=of Ful7] (Ful8){\cmark};
  \node[PROPOSITION,right=of Prop8] (Prop9){GenPro7\textsuperscript{0}};
  \node[PROPOSITION,right=of Ful8] (Ful9){\cmark};
  \node[PROPOSITION,right=of Prop9] (Prop10){GenPro8\textsuperscript{0}};
  \node[PROPOSITION,right=of Ful9] (Ful10){\cmark};
  \end{tikzpicture}
  }
 \caption{Fulfilment of propositions for the pattern generalization measure.}\label{fig:propositionFulfilment}
\end{figure*}

\textbf{The generalization measure is deterministic (DetPro\textsuperscript{+}).}
The proposed pattern generalization measure is deterministic since all steps used to compute its value are deterministic:
\begin{inparaenum}[(1)]
    \item the selection of repetitive patterns is deterministic since the tandem repeats are selected in the order of the trace with the longest length;
    \item the selection of concurrent patterns is deterministic since the oracles used are deterministic;
    \item while alignments are commonly defined as non-deterministic, \cite{S-Components} defines a deterministic version by defining various orders during the computation of the alignments;
    \item the computations of the partial fulfilments of the patterns, as well as the overall generalization, are deterministic since they compare the definitions of the patterns against certain trace positions of the alignments.
\end{inparaenum}

\textbf{The generalization measure is fully determined by the behavior of the model and not its representation (BehPro\textsuperscript{+}).}
The model is only used to compute alignments and in~\cite{GeneralizationAxioms}, it has been shown that alignments only consider the behavior of the model and not its representation.

\textbf{A model $m_2$ extending the behavior of another model $m_1$ should have a higher or equal generalization value for the same event log (GenPro1\textsuperscript{+}).}
The pattern generalization will compute alignments for the set of representative traces from the identified patterns in the event log and for the two models. Since model $m_2$ contains all behavior from model $m_1$, it either identifies the same alignments as $m_1$ or alignments with more matches, for which the patterns will either achieve the same or higher partial fulfilments.
Hence, the generalization value will be the same or higher for model $m_2$, fulfilling the proposition.

\textbf{An event log $l_2$ extending another event log $l_1$ with fully fitting traces should have a higher or equal generalization value for the same model (GenPro2\textsuperscript{+}).}
Let's assume $l_2$ adds only fitting traces to $l_1$. If the newly added traces do not contain any patterns, then the generalization of $l_2$ will be the same as $l_1$. However, if the newly added traces contain new patterns, their representative traces will include some traces that are not in $l_2$, which might not be fitting the process model. Hence, their alignments might find some mismatches that will lead to partial fulfilments that can lead to a lower generalization value than $l_1$. Hence, this proposition is not fulfilled for all logs and models. 

The proposition is not fulfilled in some cases, because the pattern generalization already derives new unseen traces from a fitting event log that might be unfitting if the model does not generalize the patterns observed in the event log. 

\textbf{An event log $l_2$ extending another event log $l_1$ with non-fitting traces should have a lesser or equal generalization value for the same model (GenPro3\textsuperscript{0}).} 
First, we investigate the direction, where $l_2$ extends $l_1$ with traces that are not included in the behavior of the model. If no patterns are included in the newly added traces, then the generalization of $l_2$ is the same as for $l_1$. Otherwise, the patterns introduced by the new traces will always achieve a partial fulfilment of zero. The last because the alignments of the representative traces of the patterns cannot be matched. Since the partial fulfilments all evaluate to zero, the generalization of $l_2$ can only decrease from the generalization of $l_1$ or stay the same, if $l_1$ already evaluates to a generalization of zero. Hence, the proposition is fulfilled by the pattern generalization. 

\textbf{Duplicating the traces of an event log with mostly fitting (unfitting) traces should lead to a higher (lower) or equal generalization value for the same model (GenPro4\textsuperscript{+},GenPro5\textsuperscript{+}, GenPro6\textsuperscript{0},GenPro7\textsuperscript{0}).}
All propositions are fulfilled by the pattern generalization since it considers relative trace frequencies when aggregating the partial fulfilments of the identified patterns. When duplicating an event log, the relative frequencies stay the same and hence the pattern generalization stays the same for any duplication of an event log. 

\textbf{When a model allows for any behavior, it should achieve a perfect generalization for any log (GenPro8\textsuperscript{0}).}
Aligning any representative trace from any pattern to a model allowing any behavior will always lead to a fully fitting trace and hence a partial fulfilment of one. The model will then achieve a perfect generalization. For the case, where the event log is empty or does not contain any patterns, the pattern generalization also assigns a value of one. Hence, the pattern generalization fulfils the proposition.
\newpage
\section{Evaluation}\label{sec:evaluation}
We implemented our generalization measure based on repetitive and concurrent patterns as a standalone open-source command line tool\footnote{The command line tool is available as PatternGeneralization 1.0 at \url{https://apromore.org/platform/tools};
All public logs and models used in the quantitative evaluation are available at 
\url{https://melbourne.figshare.com/articles/dataset/Public_benchmark_data-set_for_Conformance_Checking_in_Process_Mining/8081426}; Source code is available at \url{https://github.com/reissnda/AutomataConformance}}. Given an event log in XES format and a process model in PNML format, the tool can compute the overall generalization value or provide a breakdown of generalization values for each pattern with their corresponding weights. Optionally, the tool can output the partial fulfilments and alignments of each pattern found. It is possible to choose whether a global or a local concurrency oracle is be used to find concurrent patterns, as well as to set the filtering criteria. Additionally, the concurrent patterns can be evaluated either with a partial or an interleavings matching (shortened to P. Matching and I. Matching, respectively) as introduced in Section~\ref{sec:ConcPatternsFulfilment}.
Using this tool, we evaluated the generalization of artificial datasets to highlight the strengths and weaknesses of the proposed measure. Further, we tested the scalability of the measure by testing it on a set of publicly available event logs with models automatically discovered by two established discovery algorithms. Last, we provide a breakdown of the generalization of the public dataset for the patterns to show the possibility of a more fine-grained analysis of the pattern generalization. 

\subsection{Setup}
When conducting the experiments, we collected the generalization values of each approach and the execution times in milliseconds (ms) to measure their scalability. We set a timeout for each experiment (denoted as t/out) of 10 minutes as an acceptable execution time for a process mining analysis. 
When comparing the generalization values among several process models for the same log, we use a ranking of the values to highlight the strengths and weaknesses of each approach since the absolute difference between generalization results holds no explanatory value.

We conducted the experiments for the pattern-based generalization with both a global and a local concurrency oracle. We applied no filters for the qualitative evaluation, but applied a filter of 5\% for the global oracle and 10\% for balance and 55\% for occurrence for the local concurrency oracle. It was necessary to increase the filters for the quantitative evaluation since the real-life logs contained some infrequent events, a.k.a. noise, that significantly increased the amount of concurrent behavior. 

We chose two generalization measures as baselines: 
\begin{inparaenum}[(1)]
    \item the latest version of the Anti-Alignments generalization proposed in~\cite{AntiAlignments} and available as a plugin in ProM\footnote{\url{http://wwww.promtools.org}} (AA); and
    \item the negative events generalization proposed in~\cite{NegativeEvents}\footnote{Available at \url{http://processmining.be/neconformance/}} (NE).
\end{inparaenum}

For the baseline of Anti-Alignments, we used the ILP-Replay algorithm with a cut-off length of 5 and a backtracking threshold of 2.
For the negative events baseline, we used the ILP-based replay algorithm and a weighted Log bag based negative events inducer.
We conducted these experiments on a single-threaded 22-core Intel Xeon CPU E5-2699 v4 with 2.30GHz and with 128GB of RAM running JVM 8.

We did not consider a further generalization baseline based on adversarial networks from \cite{GANS_Generalization}, because its implementation had specific hardware requirements, warranting a different type of comparison. In other words, we focussed only on measures that can run on standard hardware.  
Further, we did not include the alignments generalization measure~\cite{AlignmentsGeneralization} since it was shown in~\cite{NegativeEvents} to provide less reliable generalization results than negative events generalization. For this reason, this measure was also excluded from the evaluation in~\cite{AntiAlignments}.

\newpage
We used process models discovered by two widely-accepted automated discovery algorithms: Split Miner (SM)~\cite{SplitMiner} and Inductive Miner (IM)~\cite{InductiveMiner}. These two algorithms perform the best in terms of fitness, precision and simplicity according to a recent benchmark~\cite{PD-Discovery-BM}. Moreover, these algorithms have been embedded in commercial tools (e.g.\ Apromore, Celonis, Minit, MyInvenio) and hence the discovered models are representative of those models discovered in practice. We applied these two algorithms with the versions and settings as described in~\cite{TandemRepeats}. 

\subsection{Datasets}\label{sec:datasets}
As for the datasets, we used the artificial datasets from the AA~\cite{AntiAlignments} and NE~\cite{NegativeEvents} papers, in order to showcase the strengths and weaknesses of each generalization measure in a qualitative evaluation. These datasets allow us to highlight specific control structures to explain the differences in the generalization values between the different measures. Furthermore, we used the real-life log-model pairs from~\cite{TandemRepeats} to evaluate the time performance of our measure against that of the two baselines, and provide a drill down of generalization values for the two patterns.

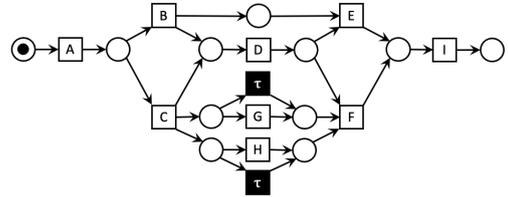
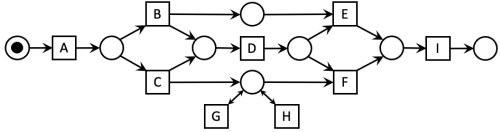
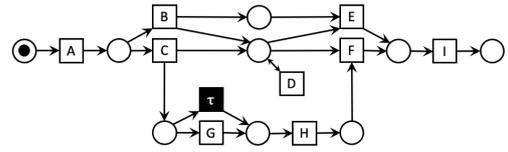
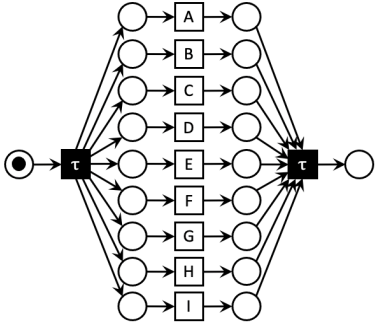
\begin{figure}[!h]
\centering
\begin{subfigure}{.5\textwidth}
  \centering
\resizebox{0.66\textwidth}{!}{ 
 \tikzstyle{ID} = [draw, rectangle, fill=white, align=center, minimum height=5mm, text width={width("$t  (1)  t$")}, font=\footnotesize]
 \tikzstyle{TRACE} = [draw, rectangle, fill=white, align=left, minimum height=5mm, text width={width("$t A,C,D,G,H,F,I t$")}, font=\footnotesize]
 \tikzstyle{COUNT} = [draw, rectangle, fill=white, align=center, minimum height=5mm, text width={width("$t count(dt,L) t$")}, font=\footnotesize]
 \begin{tikzpicture}[>=stealth', node distance=-0.3pt]
  \node[TRACE] (log) {\bf{$dt\in\uniqueTraces(\logL)$}};
  \node[ID, left=of log] (id) {\bf{ID}};
  \node[COUNT, right=of log] (count){\bf{$\countTr(dt,\logL)$}};
  \node[TRACE, below=of log] (trace1) {$\langle A,B,D,E,I \rangle$};
  \node[ID, left=of trace1] (id1) {(1)};
  \node[COUNT,right=of trace1] (count1) {1207};
  \node[TRACE, below=of trace1] (trace2) {$\langle A,C,D,G,H,F,I \rangle$};
  \node[ID, left=of trace2] (id2) {(2)};
  \node[COUNT,right=of trace2] (count2) {145};
  \node[TRACE, below=of trace2] (trace3) {$\langle A,C,G,D,H,F,I \rangle$};
  \node[ID, left=of trace3] (id3) {(3)};
  \node[COUNT,right=of trace3] (count3) {56};
  \node[TRACE, below=of trace3] (trace4) {$\langle A,C,H,D,F,I \rangle$};
  \node[ID, left=of trace4] (id4) {(4)};
  \node[COUNT, right= of trace4] (count4) {23};
  \node[TRACE, below=of trace4] (trace5) {$\langle A,C,D,H,F,I \rangle$};
  \node[ID, left=of trace5] (id5) {(5)};
  \node[COUNT, right= of trace5] (count5) {28};
  \end{tikzpicture}
  }
 \caption{Original Event log of AA dataset.}\label{fig:AA_Log_original}
\end{subfigure}%
\begin{subfigure}{.5\textwidth}
  \centering
  \includegraphics[width=0.85\textwidth]{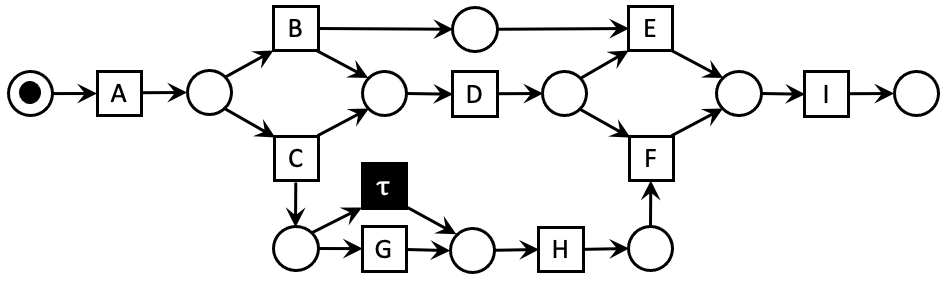}
    \caption{Generating Model} \label{fig:AA_GeneratingModel}
    \vspace{\baselineskip}
\end{subfigure}
\begin{subfigure}{.5\textwidth}
  \centering
  \includegraphics[width=0.85\textwidth]{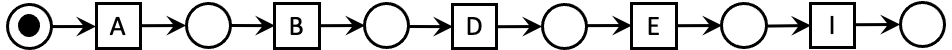}
  \caption{Single Trace Model} \label{fig:AA_SingleTraceModel}
\end{subfigure}%
\begin{subfigure}{.5\textwidth}
  \centering
  \includegraphics[width=0.5\textwidth]{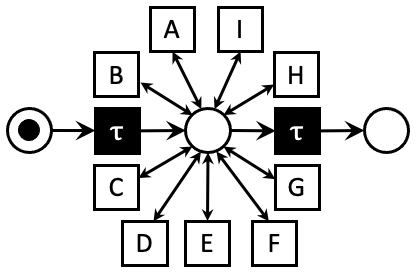}
  \caption{Flower Model} \label{fig:AA_FlowerModel}
  \vspace{\baselineskip}
\end{subfigure}
\begin{subfigure}{.5\textwidth}
  \centering
  \includegraphics[width=1\textwidth]{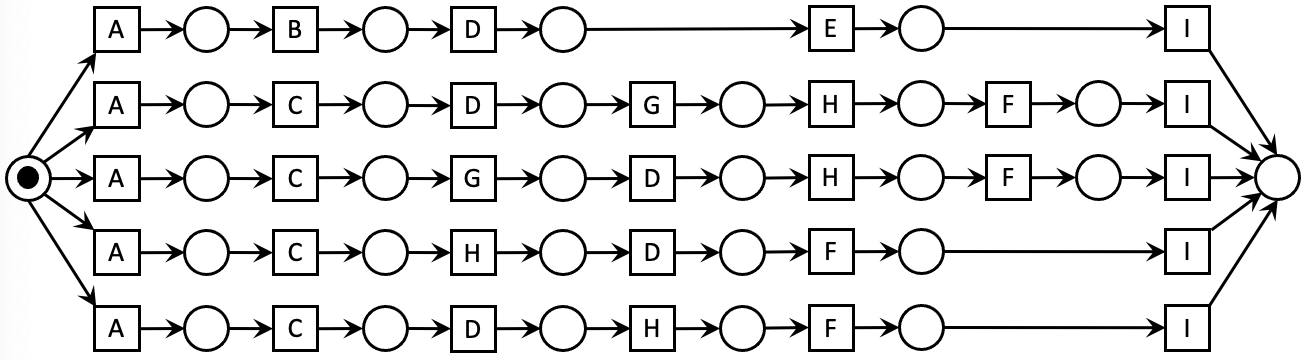}
    \caption{Four distinct traces Model} \label{fig:AA_DistinctTracesModel}
\end{subfigure}%
\begin{subfigure}{.5\textwidth}
  \centering
  \includegraphics[width=0.8\textwidth]{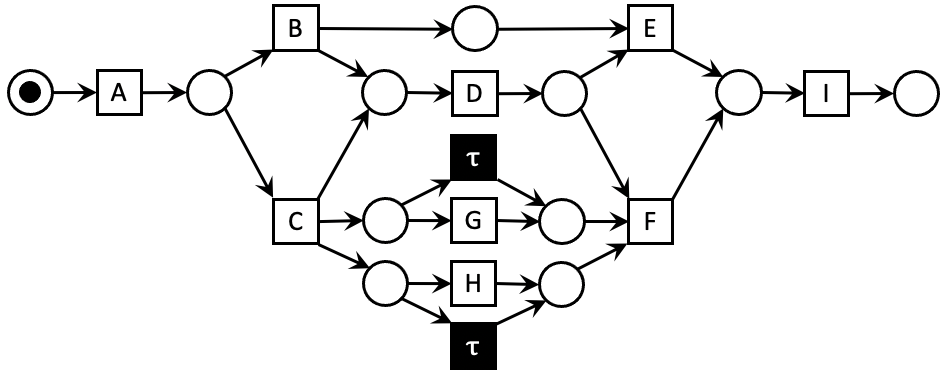}
  \caption{G and H parallel Model} \label{fig:AA_GH_Parallel_Model}
\end{subfigure}
\begin{subfigure}{.5\textwidth}
  \centering
  \includegraphics[width=0.8\textwidth]{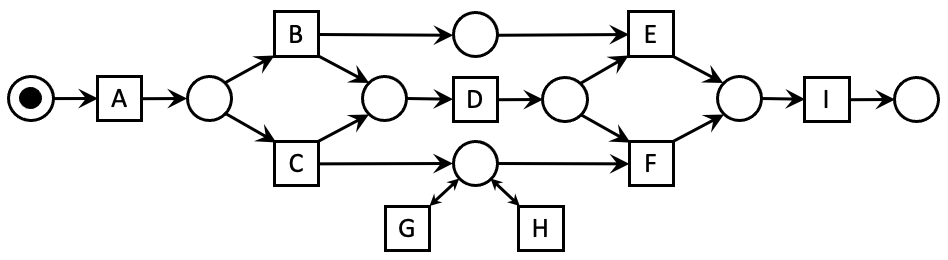}
  \caption{G and H in self loops Model} \label{fig:AA_GH_SelfLoop_Model}
\end{subfigure}%
\begin{subfigure}{.5\textwidth}
  \centering
  \includegraphics[width=0.8\textwidth]{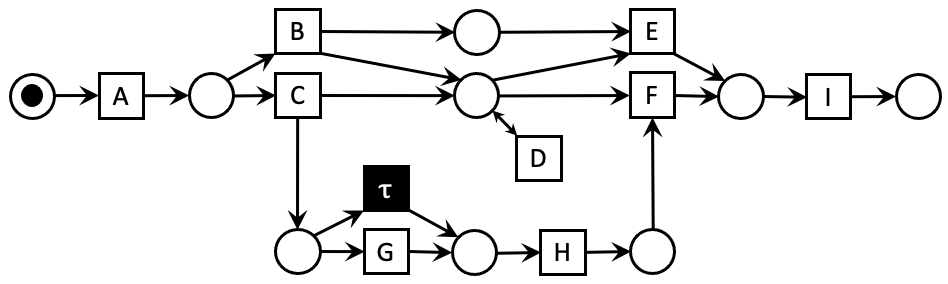}
  \caption{D in a self loop Model} \label{fig:D_SelfLoop_Model}
\end{subfigure}
\begin{subfigure}{.5\textwidth}
  \centering
  \includegraphics[width=0.6\textwidth]{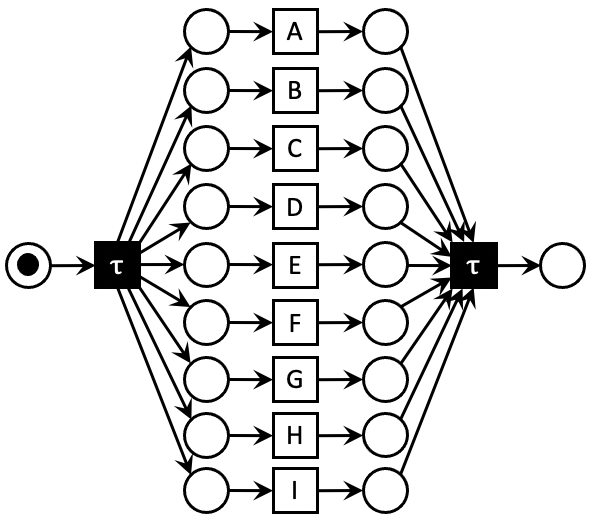}
  \caption{All-parallel Model} \label{fig:AA_AllParallelModel}
  \vspace{\baselineskip}
\end{subfigure}%
\begin{subfigure}{.5\textwidth}
  \centering
  \includegraphics[width=0.5\textwidth]{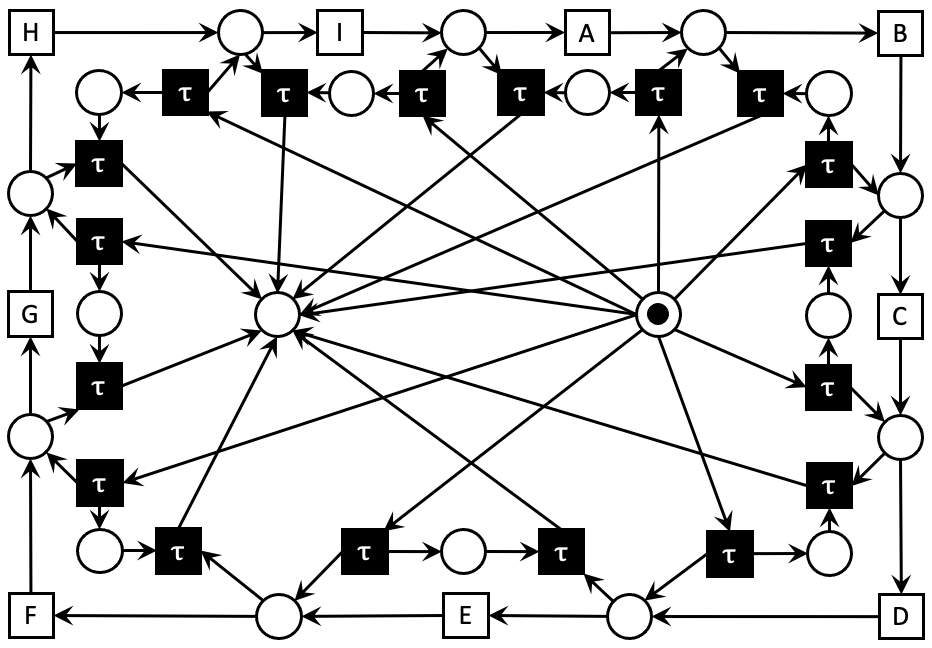}
  \caption{Round Robin Model} \label{fig:AA_RoundRobinModel}
\end{subfigure}%
\caption{Event log and process models in the AA dataset~\cite{AntiAlignments}}
\label{fig:DatasetAA}
\end{figure}

The AA dataset contains one event log and nine process models each representing this log with differing levels of generalization. Fig.~\ref{fig:DatasetAA} 
shows the event log and all process models. 

The event log (Fig.~\ref{fig:AA_Log_original}) contains five traces with differing trace counts that were generated by simulating the original process model shown in Fig.~\ref{fig:AA_GeneratingModel}. The two process models with a single trace (Fig.~\ref{fig:AA_SingleTraceModel}) or with four distinct traces have no generalizations for the event log, while the flower model (Fig.~\ref{fig:AA_FlowerModel}) has all possible generalizations. All other models have varying degrees of generalization by either using parallel structures or loops for a subset or all activities in the log. Process models from Fig.~\ref{fig:AA_GH_Parallel_Model} and Fig.~\ref{fig:AA_GH_SelfLoop_Model} generalize by modeling activities G and H as concurrent or repeatable, respectively. Fig.~\ref{fig:D_SelfLoop_Model} models activity D as a repeatable activity. The model from Fig.~\ref{fig:AA_AllParallelModel} uses a parallel structure for all activities. The round-robin model in Fig.~\ref{fig:AA_RoundRobinModel} has a set sequence for all activities that can be started at any activity and be completed once all other activities have been executed. This sequence can also be repeated any number of times.
Since the event log of the AA dataset \emph{AA-original} is rather simple and does not contain many patterns, we decided to extend this dataset with three additional logs shown in Fig.~\ref{fig:AA_additional_Logs} to demonstrate the behavior of the pattern-based generalization measure with an increasing amount of patterns:
\begin{enumerate}
    \item Event log \emph{AA-repetitive} (Fig.~\ref{fig:AA_Log_repetitive} in \ref{app:AA_additionalLogs}), where activities G, H and D are repeatable in different cycles;
    \item Event log \emph{AA-concurrent} (Fig.~\ref{fig:AA_Log_concurrent} in \ref{app:AA_additionalLogs}), where the activities  D, F, G and H are present in all different orders, i.e. they are concurrent;
    \item Event log \emph{AA-composite} (Fig.~\ref{fig:AA_Log_composite} in \ref{app:AA_additionalLogs}) that contains all the traces from the original AA log, log AA-repetitive and log AA-concurrent.
\end{enumerate}
We compare all four event logs, AA-original, AA-repetitive, AA-concurrent and AA-composite, against all nine process models of the AA dataset. 

The dataset in~\cite{NegativeEvents} contains one event log and five process models. Similar to the AA dataset, the NE dataset contains a generating model, a single trace model, a distinct traces model and a flower model. The last model is a fully connected model, where every activity can be executed after every other activity. This model from a behavioral viewpoint is equivalent to the flower model. Thus, we decided to discard it from our experiments. To replace this model, we extended the NE dataset by discovering two process models with the Split Miner and Inductive Miner. 
The total number of log-model pairs for the qualitative dataset is thus 40. 

\newpage
For the quantitative evaluation, we used the dataset in~\cite{TandemRepeats}, which contains a collection of 17 real-life event logs. They originate from the 4TU Center for Research Data\footnote{\url{https://data.4tu.nl/repository/collection:event\_logs\_real}}. This dataset was also used in a recent benchmark for process discovery \cite{PD-Discovery-BM}. 

It consists of logs from the Business Process Intelligence challenge (BPIC) series, i.e.\ BPIC12 \cite{BPIC12}, BPIC13\textsubscript{cp} \cite{BPIC13cp}, BPIC13\textsubscript{inc} \cite{BPIC13inc}, BPIC14 \cite{BPIC14}, BPIC15 \cite{BPIC15}, BPIC17 \cite{BPIC17}, BPIC18 \cite{BPIC18}, BPIC19 \cite{BPIC19}, the Road Traffic Fines Management process log (RTFMP) \cite{RTFMP} and the SEPSIS Cases log (SEPSIS) \cite{SEPSIS}. 
As in \cite{PD-Discovery-BM}, the logs for BPIC15, BPIC17, BPIC18 and BPIC19 were filtered by removing infrequent events with the filtering technique from~\cite{Noise-filtering} since it was not possible to discover process models from the corresponding unfiltered logs. 
The logs that were filtered are annotated with $``f''$. 
These public logs cover process executions from different domains such as finance, healthcare, government and IT service management. 

For each event log in the quantitative evaluation, we used two process models discovered automatically using SM and IM. This resulted in a total of 34 log-model pairs for the quantitative evaluation.

\begin{sidewaystable}[htbp!]                                   
{
\footnotesize{                                   
\setlength{\tabcolsep}{3pt}                                   
\centering{                                   
\begin{tabular}{|l|r r|r r|r r r r|r r r r|r r r r|}                                   
\hline                                   
\multirow{3}{*}{\bf{Log}} & \multicolumn{2}{c|}{\multirow{2}{*}{\bf{\#Events}}}   & \multicolumn{2}{c|}{\multirow{2}{*}{\bf{\#Traces}}}   & \multicolumn{4}{c|}{\multirow{2}{*}{\bf{Repetitive Patterns}}}       & \multicolumn{8}{c|}{\bf{Concurrent Patterns}}               \\  
 & \multicolumn{2}{c|}{}   & \multicolumn{2}{c|}{}   & \multicolumn{4}{c|}{}       & \multicolumn{4}{c|}{\bf{global}}       & \multicolumn{4}{c|}{\bf{local}}       \\  
 & Total & Unq & Total & Unq & \#Ext Tr & \#Patterns & Avg. labels & Weight & \#POs & \#Unq Tr & \#Patterns & Weight & \#POs & \#Unq Tr & \#Patterns & Weight \\ \hline 
\makecell{AA\\original} &  7,748  &  9  &  1,459  &  5  &  -    &  -    &  -    &  -    &  2  &  5  &  2  &  252  &  2  &  4  &  2  &  252  \\  
\makecell{AA\\repetitive} &  16,000  &  7  &  1,250  &  5  &  5  &  7  &  1.29  &  1,750  &  5  &  32  &  5  &  1,250  &  -    &  -    &  -    &  -    \\  
\makecell{AA\\concurrent} &  8,400  &  7  &  1,200  &  24  &  -    &  -    &  -    &  -    &  1  &  24  &  1  &  1,200  &  1  &  24  &  1  &  1,200  \\  
\makecell{AA\\composite} &  30,741  &  9  &  3,708  &  32  &  5  &  7  &  1.29  &  1,750  &  8  &  248  &  8  &  2,501  &  4  &  72  &  4  &  1,751  \\  
NE &  3,725  &  11  &  500  &  15  &  2  &  2  &  3.00  &  21  &  4  &  12  &  4  &  338  &  3  &  9  &  3  &  333  \\  \hline  \hline
BPIC12 &  262,200  &  24  &  13,087  &  4,366  &  2,279  &  8,071  &  1.24  &  19,851  &  4,006  &  1,794,135  &  13,439  &  15,874  &  \multicolumn{4}{c|}{t/out}        \\  
BPIC13\textsubscript{cp} &  6,660  &  4  &  1,487  &  183  &  104  &  172  &  1.84  &  581  &  -    &  -    &  -    &  -    &  102  &  666  &  105  &  170  \\  
BPIC13\textsubscript{inc} &  65,533  &  4  &  7,554  &  1,511  &  1,046  &  3,052  &  1.81  &  9,815  &  -    &  -    &  -    &  -    &  \multicolumn{4}{c|}{t/out}        \\  
BPIC14\textsubscript{f} &  369,485  &  9  &  41,353  &  14,948  &  10,002  &  18,038  &  1.54  &  20,181  &  -    &  -    &  -    &  -    &  \multicolumn{4}{c|}{t/out}        \\  
BPIC15\textsubscript{1f} &  21,656  &  70  &  902  &  295  &  -    &  -    &  -    &  -    &  -    &  -    &  -    &  -    &  -    &  -    &  -    &  -    \\  
BPIC15\textsubscript{2f} &  24,678  &  82  &  681  &  420  &  1  &  1  &  4.00  &  1  &  229  &  623  &  298  &  347  &  -    &  -    &  -    &  -    \\  
BPIC15\textsubscript{3f} &  43,786  &  62  &  1,369  &  826  &  -    &  -    &  -    &  -    &  659  &  9,095  &  1,689  &  2,982  &  -    &  -    &  -    &  -    \\  
BPIC15\textsubscript{4f} &  29,403  &  65  &  860  &  451  &  1  &  1  &  3.00  &  1  &  161  &  504  &  397  &  675  &  -    &  -    &  -    &  -    \\  
BPIC15\textsubscript{5f} &  30,030  &  74  &  975  &  446  &  -    &  -    &  -    &  -    &  -    &  -    &  -    &  -    &  -    &  -    &  -    &  -    \\  
BPIC17\textsubscript{f} &  714,198  &  18  &  21,861  &  8,767  &  2,648  &  14,228  &  1.39  &  109,289  &  8,767  &  638,955  &  20,947  &  48,181  &  \multicolumn{4}{c|}{t/out}        \\  
RTFMP &  561,470  &  11  &  150,370  &  231  &  12  &  12  &  1.00  &  3,895  &  133  &  1,008  &  175  &  24,318  &  74  &  907  &  135  &  4,584  \\  
SEPSIS &  15,214  &  16  &  1,050  &  846  &  444  &  840  &  1.61  &  848  &  469  &  938  &  469  &  563  &  \multicolumn{4}{c|}{t/out}        \\  
BPIC18 &  2,514,266  &  41  &  43,809  &  28,457  &  27,279  &  149,444  &  1.94  &  203,286  &  \multicolumn{4}{c|}{t/out}        &  \multicolumn{4}{c|}{t/out}        \\  
BPIC19\textsubscript{1} &  5,898  &  11  &  1,044  &  148  &  76  &  101  &  1.01  &  476  &  131  &  17,487  &  576  &  2,662  &  46  &  131  &  46  &  87  \\  
BPIC19\textsubscript{2} &  319,233  &  38  &  15,182  &  4,228  &  2,690  &  11,655  &  3.56  &  14,289  &  \multicolumn{4}{c|}{t/out}        &  \multicolumn{4}{c|}{t/out}        \\  
BPIC19\textsubscript{3} &  1,234,708  &  39  &  221,010  &  7,832  &  1,693  &  3,094  &  1.26  &  7,234  &  93  &  318  &  116  &  853  &  \multicolumn{4}{c|}{t/out}        \\  
BPIC19\textsubscript{4} &  36,084  &  15  &  14,498  &  281  &  67  &  82  &  1.16  &  1,040  &  12  &  31  &  12  &  46  &  18  &  113  &  22  &  38  \\  \hline  
\end{tabular}                                   
}                                   
\vspace{.5\baselineskip}                                   
\caption{Log and pattern statistics}\label{tb:LogStatistics}                                   
\vspace{.5\baselineskip}                                   
}}                                   
\end{sidewaystable}
                                                              

Table~\ref{tb:LogStatistics} shows the characteristics of all event logs used in our evaluation as well as the characteristics of all discovered patterns that were investigated in the pattern-based generalization measure. The logs feature a wide range of characteristics, including simple and complex logs. The size of the logs differs in terms of the total number of events (3.7K to 2.5M) and traces (500 to 221K). The number of unique ($``Unq''$) events varies from 4 to 82, while the number of unique traces varies from 5 to 28K. The number of unique traces relates to how many problems need to be solved by any generalization measure while the unique number of events? correlates with the difficulty of each problem. 

For the repetitive patterns, we report various measures related to the amount of repetitive behavior present in the event logs and the complexity to compute the generalization. The number of extended traces \#Ext (0 to 27K) represents the number of unique traces that contain tandem repeats, i.e.\ the number of traces for which alignments need to be computed to determine the generalization for repetitive behavior. The number of repetitive patterns \#Ps (0 to 150K) demonstrates the amount of repetitive behavior in the event log and relates to the number of extended tandem repeats. The average number of repeated labels $\varnothing$lbls (0 to 4) shows how many labels on average are repeated per pattern. The weight of the repetitive patterns (0 to 203K) is the sum of the number of traces related to the repetitive patterns. 

For the concurrent patterns, the table records various measures for both the global and the local concurrency oracle. For the qualitative evaluation, we applied no filters to either concurrency oracle. For the quantitative evaluation, we observed that infrequent activities would induce concurrencies that would lead to an exponential amount of patterns, sometimes leading to all activities being concurrent. Hence, we decided to apply filters to both concurrency oracles as described in Sec.~\ref{sec:Concurrency}: a filter of 5\% to the global oracle and an occurrence of 55\% and a balance of 10\% to the local oracle. For some datasets, these filters were not sufficient and our recording resulted in a timeout (t/out). For both oracles, we recorded the number of partial orders POs that we built from all unique traces (global 0 to 8K; local 0 to 102). It already becomes apparent that the local oracle discovers less concurrency, i.e.\ it is more precise at pinpointing concurrency. However, the local oracle scales less towards larger datasets as it had more timeouts than the global oracle (global 2; local 8). The number of unique traces \#Unq extracted from the partial orders indicates how many alignments need to be computed to determine the generalization for concurrent patterns (global 0 to 1.8M; local 0 to 907). BPIC12 highlights the problem of the exponential explosion of unique traces with 1.8 million traces. For this dataset, a higher filter should be considered. The number of concurrent patterns \#Ps indicates how many concurrent structures are being tested to determine the generalization for concurrency (global 0 to 21K; local 0 to 135). Finally, the weight of concurrent patterns is the sum of all trace counts related to the concurrent patterns (global 0 to 48K; local 0 to 4.5K).

\begin{table}[htbp!]              
{\footnotesize{              
\setlength{\tabcolsep}{3pt}              
\centering{              
\begin{tabular}{|l|l|r r r r r|}              
\hline              
\bf{Dataset} & \bf{Model} & \bf{Size} & \bf{Places} & \bf{Transitions} & \bf{Choices} & \bf{Parallel} \\ \hline
\multirow{9}{*}{AA} & Generating M. & 44 & 10 & 10 & 3 & 2 \\ 
 & Single trace & 21 & 6 & 5 & 0 & 0 \\ 
 & Flower M. & 36 & 3 & 11 & 1 & 0 \\ 
 & Distinct traces & 121 & 28 & 31 & 1 & 0 \\ 
 & G,H parallel & 50 & 11 & 11 & 4 & 2 \\ 
 & G,H self loop & 39 & 8 & 9 & 3 & 2 \\ 
 & D self loop & 43 & 9 & 10 & 3 & 2 \\ 
 & All parallel & 69 & 20 & 11 & 0 & 1 \\ 
 & Round robin & 119 & 20 & 27 & 10 & 9 \\ \hline
\multirow{6}{*}{NE} & Perfect M. & 46 & 11 & 11 & 2 & 1 \\ 
 & Single trace & 21 & 6 & 5 & 0 & 0 \\ 
 & Flower M. & 36 & 3 & 11 & 1 & 0 \\ 
 & Distinct traces & 102 & 24 & 26 & 1 & 0 \\ 
 & SM & 50 & 11 & 13 & 3 & 0 \\ 
 & IM & 78 & 17 & 19 & 5 & 2 \\ \hline
\multirow{17}{*}{IM} & BPIC12 & 177 & 32 & 45 & 16 & 2 \\ 
 & BPIC13\textsubscript{cp} & 31 & 7 & 8 & 2 & 0 \\ 
 & BPIC13\textsubscript{inc} & 56 & 13 & 13 & 3 & 1 \\ 
 & BPIC14\textsubscript{f} & 124 & 27 & 29 & 8 & 2 \\ 
 & BPIC15\textsubscript{1f} & 449 & 68 & 127 & 48 & 0 \\ 
 & BPIC15\textsubscript{2f} & 537 & 85 & 150 & 55 & 1 \\ 
 & BPIC15\textsubscript{3f} & 464 & 74 & 128 & 47 & 3 \\ 
 & BPIC15\textsubscript{4f} & 469 & 74 & 131 & 51 & 1 \\ 
 & BPIC15\textsubscript{5f} & 381 & 48 & 111 & 31 & 0 \\ 
 & BPIC17\textsubscript{f} & 121 & 22 & 33 & 8 & 0 \\ 
 & RTFMP & 111 & 23 & 26 & 9 & 2 \\ 
 & SEPSIS & 145 & 26 & 37 & 13 & 3 \\ 
 & BPIC18 & 235 & 48 & 57 & 18 & 6 \\ 
 & BPIC19\textsubscript{1} & 44 & 9 & 11 & 4 & 1 \\ 
 & BPIC19\textsubscript{2} & 186 & 31 & 47 & 13 & 4 \\ 
 & BPIC19\textsubscript{3} & 279 & 53 & 70 & 23 & 7 \\ 
 & BPIC19\textsubscript{4} & 85 & 14 & 23 & 8 & 1 \\ \hline
\multirow{17}{*}{SM} & BPIC12 & 315 & 58 & 85 & 29 & 1 \\ 
 & BPIC13\textsubscript{cp} & 49 & 10 & 13 & 4 & 0 \\ 
 & BPIC13\textsubscript{inc} & 56 & 11 & 15 & 5 & 0 \\ 
 & BPIC14\textsubscript{f} & 88 & 16 & 24 & 9 & 0 \\ 
 & BPIC15\textsubscript{1f} & 368 & 74 & 98 & 25 & 0 \\ 
 & BPIC15\textsubscript{2f} & 444 & 93 & 117 & 25 & 0 \\ 
 & BPIC15\textsubscript{3f} & 296 & 62 & 78 & 17 & 0 \\ 
 & BPIC15\textsubscript{4f} & 323 & 68 & 85 & 18 & 0 \\ 
 & BPIC15\textsubscript{5f} & 359 & 77 & 94 & 18 & 0 \\ 
 & BPIC17\textsubscript{f} & 149 & 29 & 40 & 12 & 0 \\ 
 & RTFMP & 102 & 18 & 28 & 11 & 0 \\ 
 & SEPSIS & 162 & 30 & 44 & 15 & 0 \\ 
 & BPIC18 & 251 & 35 & 72 & 16 & 0 \\ 
 & BPIC19\textsubscript{1} & 63 & 9 & 18 & 4 & 0 \\ 
 & BPIC19\textsubscript{2} & 232 & 28 & 68 & 14 & 0 \\ 
 & BPIC19\textsubscript{3} & 378 & 42 & 112 & 20 & 0 \\ 
 & BPIC19\textsubscript{4} & 106 & 13 & 31 & 8 & 0 \\ \hline
\end{tabular}              
}              
\vspace{.5\baselineskip}              
\caption{Model statistics}\label{tb:ModelStatistics}              
\vspace{.5\baselineskip}              
}}              
\end{table}              

Table~\ref{tb:ModelStatistics} shows the characteristics of the process models from the qualitative and quantitative evaluation. It reports their size, number of places and transitions as well as the number of choices, i.e.\ places with multiple outgoing arcs, and parallel structures, i.e.\ transitions with multiple outgoing arcs. The dataset contains process models of various sizes (21 to 537) to test the generalization measures in different settings. Of particular interest are the number of choices and parallel structures since they correspond to the generalizations that are being tested, i.e.\ choices may indicate cycles and parallel structures can represent concurrencies. One interesting finding is that Split Miner models rarely include parallel structures, but include some more choices in comparison to the Inductive Miner models.

\subsection{Qualitative evaluation}

Table~\ref{tb:qual_eval_generalization} shows the results of the qualitative evaluation. The results are presented by log groups, and the generalization results are compared among all models for each of the four logs: AA original, AA repetitive, AA concurrent, AA composite and NE. The table shows both generalization values and the ranking of models according to their generalization values for the baseline approaches and the pattern-based generalization measure with both partial matching and interleavings matching as well as both global and local concurrency oracle. The execution times of the qualitative evaluation can be found in~\ref{app:cost_comparison}.

\vspace{\baselineskip}
\begin{table}[htbp!]                            
{\footnotesize{                            
\setlength{\tabcolsep}{3pt}                            
\centering{                            
\begin{tabular}{|c|l|r r|r r|r r|c c|c c|c c|}                            
\cline{3-14}                            
\multicolumn{2}{c|}{}   & \multicolumn{6}{c|}{\bf{Generalization values}}           & \multicolumn{6}{c|}{\bf{Generalization ranking}}           \\ \cline{3-14}
\multicolumn{2}{c|}{}   & \multicolumn{2}{c|}{\multirow{2}{*}{\bf{Baselines}}}   & \multicolumn{4}{c|}{\bf{Pattern Generalization}}       & \multicolumn{2}{c|}{\multirow{2}{*}{\bf{Baselines}}}   & \multicolumn{4}{c|}{\bf{Pattern Generalization}}       \\ 
\multicolumn{2}{c|}{}   & \multicolumn{2}{c|}{}   & \multicolumn{2}{c|}{\bf{P. Matching}}   & \multicolumn{2}{c|}{\bf{I. Matching}}   & \multicolumn{2}{c|}{}   & \multicolumn{2}{c|}{\bf{P. Matching}}   & \multicolumn{2}{c|}{\bf{I. Matching}}   \\ \hline
\multicolumn{2}{|c|}{\bf{Dataset}}   & AA & NE & global & local & global & local & AA & NE & global & local & global & local \\ \hline
\multirow{9}{*}{\makecell{AA\\original}} & Generating M. & 0.30 & 0.57 & 1.00 & 1.00 & 1.00 & 1.00 & 7 & 6 & 1 & 1 & 1 & 1 \\ 
 & Single trace & 0.00 & 0.33 & 0.37 & 0.50 & 0.00 & 0.00 & 8 & 8 & 9 & 9 & 9 & 9 \\ 
 & Flower M. & 0.81 & 1.00 & 1.00 & 1.00 & 1.00 & 1.00 & 1 & 1 & 1 & 1 & 1 & 1 \\ 
 & Distinct traces & 0.00 & 0.36 & 0.91 & 1.00 & 0.73 & 1.00 & 8 & 7 & 7 & 1 & 7 & 1 \\ 
 & G,H parallel & 0.44 & 0.57 & 1.00 & 1.00 & 1.00 & 1.00 & 3 & 5 & 1 & 1 & 1 & 1 \\ 
 & G,H self loop & 0.36 & 0.58 & 1.00 & 1.00 & 1.00 & 1.00 & 6 & 4 & 1 & 1 & 1 & 1 \\ 
 & D self loop & 0.68 & t/out & 1.00 & 1.00 & 1.00 & 1.00 & 2 &  & 1 & 1 & 1 & 1 \\ 
 & All parallel & 0.43 & 0.84 & 1.00 & 1.00 & 1.00 & 1.00 & 4 & 3 & 1 & 1 & 1 & 1 \\ 
 & Round robin & 0.38 & 1.00 & 0.68 & 0.75 & 0.37 & 0.50 & 5 & 1 & 8 & 8 & 8 & 8 \\ \hline
\multirow{9}{*}{\makecell{AA\\repetitive}} & Generating M. & 0.44 & 0.70 & 0.24 & 0.00 & 0.00 & 0.00 & 3 & 2 & 4 & 4 & 4 & 4 \\ 
 & Single trace & 0.00 & 0.24 & 0.08 & 0.00 & 0.00 & 0.00 & 8 & 7 & 9 & 4 & 4 & 4 \\ 
 & Flower M. & 0.75 & 1.00 & 1.00 & 1.00 & 1.00 & 1.00 & 1 & 1 & 1 & 1 & 1 & 1 \\ 
 & Distinct traces & 0.00 & 0.69 & 0.23 & 0.00 & 0.00 & 0.00 & 8 & 3 & 7 & 4 & 4 & 4 \\ 
 & G,H parallel & 0.42 & 0.69 & 0.24 & 0.00 & 0.00 & 0.00 & 4 & 4 & 4 & 4 & 4 & 4 \\ 
 & G,H self loop & 0.27 & t/out & 0.79 & 0.71 & 0.67 & 0.71 & 6 & - & 2 & 2 & 2 & 2 \\ 
 & D self loop & 0.49 & t/out & 0.45 & 0.29 & 0.33 & 0.29 & 2 & - & 3 & 3 & 3 & 3 \\ 
 & All parallel & 0.25 & 0.36 & 0.24 & 0.00 & 0.00 & 0.00 & 7 & 6 & 4 & 4 & 4 & 4 \\ 
 & Round robin & 0.38 & 0.48 & 0.18 & 0.00 & 0.00 & 0.00 & 5 & 5 & 8 & 4 & 4 & 4 \\ \hline
\multirow{9}{*}{\makecell{AA\\concurrent}} & Generating M. & 0.44 & 0.74 & 0.71 & 0.71 & 0.13 & 0.13 & 3 & 3 & 5 & 5 & 5 & 5 \\ 
 & Single trace & 0.00 & 0.21 & 0.25 & 0.25 & 0.00 & 0.00 & 8 & 7 & 9 & 9 & 9 & 9 \\ 
 & Flower M. & 0.89 & 1.00 & 1.00 & 1.00 & 1.00 & 1.00 & 1 & 1 & 1 & 1 & 1 & 1 \\ 
 & Distinct traces & 0.00 & 0.70 & 0.68 & 0.68 & 0.08 & 0.08 & 8 & 5 & 7 & 7 & 7 & 7 \\ 
 & G,H parallel & 0.42 & 0.76 & 0.79 & 0.79 & 0.25 & 0.25 & 4 & 2 & 3 & 3 & 3 & 3 \\ 
 & G,H self loop & 0.15 & t/out & 0.79 & 0.79 & 0.25 & 0.25 & 7 & - & 3 & 3 & 3 & 3 \\ 
 & D self loop & 0.63 & t/out & 0.71 & 0.71 & 0.13 & 0.13 & 2 & - & 5 & 5 & 5 & 5 \\ 
 & All parallel & 0.33 & 0.71 & 1.00 & 1.00 & 1.00 & 1.00 & 6 & 4 & 1 & 1 & 1 & 1 \\ 
 & Round robin & 0.42 & 0.41 & 0.59 & 0.59 & 0.04 & 0.04 & 5 & 6 & 8 & 8 & 8 & 8 \\ \hline
\multirow{9}{*}{\makecell{AA\\composite}} & Generating M. & 0.26 & 0.61 & 0.37 & 0.34 & 0.04 & 0.05 & 7 & 3 & 6 & 6 & 6 & 6 \\ 
 & Single trace & 0.00 & 0.40 & 0.12 & 0.12 & 0.00 & 0.00 & 8 & 7 & 9 & 9 & 9 & 9 \\ 
 & Flower M. & 0.51 & 1.00 & 1.00 & 1.00 & 1.00 & 1.00 & 2 & 1 & 1 & 1 & 1 & 1 \\ 
 & Distinct traces & 0.00 & 0.46 & 0.35 & 0.32 & 0.03 & 0.03 & 8 & 6 & 7 & 7 & 7 & 7 \\ 
 & G,H parallel & 0.32 & 0.61 & 0.39 & 0.38 & 0.07 & 0.09 & 6 & 4 & 5 & 5 & 5 & 5 \\ 
 & G,H self loop & 0.35 & t/out & 0.76 & 0.76 & 0.39 & 0.52 & 5 & - & 2 & 2 & 2 & 2 \\ 
 & D self loop & 0.54 & t/out & 0.50 & 0.50 & 0.18 & 0.23 & 1 & - & 3 & 3 & 4 & 4 \\ 
 & All parallel & 0.39 & 0.60 & 0.48 & 0.45 & 0.29 & 0.36 & 4 & 5 & 4 & 4 & 3 & 3 \\ 
 & Round robin & 0.44 & 1.00 & 0.31 & 0.27 & 0.01 & 0.02 & 3 & 1 & 8 & 8 & 8 & 8 \\ \hline
\multirow{6}{*}{NE} & Perfect M. & 0.18 & 0.68 & 1.00 & 1.00 & 1.00 & 1.00 & 2 & 2 & 1 & 1 & 1 & 1 \\ 
 & Single trace & 0.00 & 0.32 & 0.00 & 0.00 & 0.00 & 0.00 & 3 & 5 & 6 & 6 & 6 & 6 \\ 
 & Flower M. & 0.84 & 1.00 & 1.00 & 1.00 & 1.00 & 1.00 & 1 & 1 & 1 & 1 & 1 & 1 \\ 
 & Distinct traces & 0.00 & 0.36 & 0.94 & 0.94 & 0.94 & 0.94 & 3 & 4 & 4 & 4 & 4 & 4 \\ 
 & SM & t/out & t/out & 0.78 & 0.79 & 0.37 & 0.37 & - & - & 5 & 5 & 5 & 5 \\ 
 & IM & t/out & 0.60 & 1.00 & 1.00 & 1.00 & 1.00 & - & 3 & 1 & 1 & 1 & 1 \\ \hline
\end{tabular}                            
}                            
\vspace{.5\baselineskip}                            
\caption{Generalization of qualitative experiments}\label{tb:qual_eval_generalization}                            
\vspace{.5\baselineskip}                            
}}                            
\end{table}                                                   

\newpage
\textbf{AA - original.}
The original event log of the AA dataset highlights the weakness of the proposed pattern-based generalization measure. The log only contains one concurrent pattern involving  activities D, G and H. It is fully fulfilled in the 
generating model, the flower model, the G, H parallel model, the G, H in a self loop model, the D self loop model and the all-parallel model and hence they all share a generalization ranking of 1.
The other models only partially fulfil the pattern and so the distinct traces model is ranked before the round robin model, in turn before the single trace model. 
The baselines AA and NE agree with ranking the flower model the highest and rank the Single trace, distinct traces model last.
For the other models, they achieve a more fine-grained ranking for this event log by either enhancing the event log with additional activities (NE) or by deriving the generalization from the model state space (AA).
The NE baseline ranks the round-robin model tied with the flower model with a perfect generalization, which is counterintuitive, because the round-robin model can only repeat a certain sequence of activities and cannot reproduce the observed concurrent pattern of activities G, H and D. The AA baseline disagrees with NE and ranks the round-robin model lower with rank five, which we still disagree with since it violates the concurrent pattern found.
The AA baseline ranks the D self loop model second highest-while the NE baseline timed out for this process model. We want to point out that activity D is never repeated in the event log and hence there is no support from the data that this generalization is beneficial for this event log. 
This argument also holds for the model with G, H in a self loop and all-parallel model as well, since neither G nor H are repeated in the log either.
The discussion about the original AA log therefore revolves around the question of whether generalizing structures of a process model should be rewarded a higher generalization without having any support in the event log. If this is the case, then NE and AA are better measures for event logs that contain little to no patterns. 
To further investigate the usefulness of the pattern generalization in different use cases, we extended the AA dataset with event logs that show clear patterns of generalization as described in Sec.~\ref{sec:datasets}, namely AA repetitive, AA concurrent and AA composite, in order to check how the ranking of the nine process models changes for all generalization measures.

\textbf{AA - repetitive.}
The repetitive log for AA contains several repeating activities, such as G, H and D as well as loops involving two activities, i.e.\ G, H and H, G. 
The repeating activities G and H occur more frequently than activity D such that the model with G, H in self loops should have a better generalization than the model with D in a self loop. 
The flower model should have the highest generalization since it includes any kind of repetitive behavior for all activities. The models without any cycles, such as the generating model, single trace, distinct traces, G, H parallel and the all-parallel model should all achieve a lower generalization. 
All measures agree on ranking the flower model the highest. The pattern-based generalization measures then all rank correctly the model with G, H in a loop over the model with D in a loop over all other process models. 
The pattern generalization measure with partial matching and the global oracle found one infrequent concurrent pattern to further distinguish between the remaining models.
The NE baseline timed out for the two models with D or G, H in loops. Besides that, we observed some counterintuitive rankings for the NE baseline, i.e.\ the generating and distinct trace models are ranked higher than G, H parallel or the all-parallel model event though the latter models contain more general structures than the former models.
The AA baseline correctly ranks high the model with D in a loop, but ranks the model with G, H in a loop very low (rank 6 out of 8) despite both activities G and H are  executed repeatedly with a high frequency. Some further counterintuitive results are to rank the all-parallel model lower than the generating model or the G, H parallel model since these models have a strictly lower ability to generalize concurrency.

\newpage
\textbf{AA - concurrent.}
The concurrent log for AA contains one concurrent pattern involving the four activities D, F, G and H, i.e.\ it contains traces with all possible orderings of the four activities.
The flower model and the all-parallel model should achieve a perfect generalization since these two models capture any kind of concurrency. Next, the models with G, H in parallel and G, H in loops should rank the second highest since they contain three activities G, H and D which are concurrent. After that the generating model and the model with D in a loop contain D concurrently with G and H, which are in a sequence. All other models do not contain a concurrent structure and hence should be ranked lower.
The pattern-based generalization measure in all settings finds the correct ranking as described above. Additionally, it distinguishes the ranking between the lowest ranked models of the distinct traces, single trace and round robin model by using the partial fulfilments of the concurrent pattern.
The AA baseline correctly ranks the flower model the highest, but ranks the all-parallel model very low (rank 6 out of 8), which contradicts the support for the concurrent structure given in the event log.
Further, the generating model is ranked higher than the G, H parallel or the G, H in a loop model, despite the latter two models contain a better generalizing structure for three of the four activities. The round robin model is also ranked rather high despite containing no concurrent structure.
The NE baseline again timed out for the models with D and G, H in loops. It ranks the flower model the highest and also ranks the all-parallel model rather poorly (rank 4 out of 7) behind the G, H parallel model and the generating model.

\textbf{AA - composite.}
The composite log for AA contains all traces from the original, repetitive and concurrent logs. All three event logs have the same trace count, so the composite log captures the characteristics of each of the three logs equally. Hence, models that can both generalize the loops of the repetitive log and the concurrent activities of the concurrent log should achieve a higher generalization. 
The pattern-based generalization measures agree on ranking the flower model the highest and the model with G, H in a loop the second highest since this latter model contains a mixture of repetitive and concurrent behavior. Next, the ranking differs whether concurrent patterns are evaluated with partial matching or with the interleavings matching. In the former case, concurrent behavior is easier to fulfil partially and hence the model with D in a loop is raked above the all-parallel model. The latter will reverse this order since concurrent patterns are evaluated more strictly letting the all-parallel model pull ahead.
The AA baseline actually does not rank the flower model first but second despite it having the most general structures a process model can possess. It ranks the round-robin model very high (rank 3 out of 8) despite having limited amount of repetitive behavior and no concurrent behavior. 
The NE baseline ranks the flower model the highest tied with the round-robin model. This is counterintuitive since the round-robin model has poorly generalizing repetitive structures, and no concurrent structures. Further, the generating model and the G, H parallel models are ranked higher than the all-parallel model despite the all-parallel model has more generalizing concurrent structures. 

\textbf{Conclusion of the AA experiments.}
With an increasing amount of patterns in the AA event logs, it becomes more and more apparent that both AA and NE baselines do not relate their ranking of process models to any support of patterns in an event log. 
Further, some mis-rankings show weaknesses in both the NE and the AA baselines. In contrast, our measure, in different settings, detects the fulfilment of the patterns quite well. With an increasing amount of patterns, the measure can discriminate more finely between process models. This also highlights a weakness of the approach, i.e.\ the measure will not be useful for very small event logs that do not contain any patterns. Rather, it works well with real-life event logs as these feature a large number of patterns as can be observed in the log statistics of Table~\ref{tb:LogStatistics}.

\textbf{NE dataset.}
The NE event log contains a small amount of concurrent and repetitive patterns. Hence, models with a good mixture of parallel and loop structures should achieve a good generalization.
The AA baseline times out for both models. It then ranks the flower model over the perfect model over the remaining models.
The NE baseline agrees with the ranking of the AA baseline, but includes the IM model after the perfect model and before the remaining models. It times out for the Split Miner model.
The pattern generalization measures rank the flower model, the IM model and the perfect model the highest. The log does not contain sufficient patterns to further distinguish the generalization of the three models. The SM model achieves a rather low generalization since it poorly fulfils the concurrent patterns.

\subsection{Quantitative evaluation}
The quantitative evaluation intends to assess the scalability and usefulness of all generalization measures for large real-life datasets. Table~\ref{tb:quant_eval_wBenchmarks} reports the execution times in milliseconds (ms) and the generalization values of all baselines and the pattern-based generalization measure with both global and local oracles as well as with partial and interleavings matching for concurrent patterns. 

\vspace{\baselineskip}
\begin{table}[htbp!]                             
{\footnotesize{                             
\setlength{\tabcolsep}{3pt}                             
\centering{                             
\begin{tabular}{|c|l|r r|r r|r r|r r|r r|r r|}                             
\cline{3-14}                             
\multicolumn{2}{c|}{}   & \multicolumn{6}{c|}{\bf{Execution times in ms}}           & \multicolumn{6}{c|}{\bf{Generalization values}}           \\ \cline{3-14} 
\multicolumn{2}{c|}{}   & \multicolumn{2}{c|}{\multirow{2}{*}{\bf{Baselines}}}   & \multicolumn{4}{c|}{\bf{Pattern Generalization}}       & \multicolumn{2}{c|}{\multirow{2}{*}{\bf{Baselines}}}   & \multicolumn{4}{c|}{\bf{Pattern Generalization}}       \\  
\multicolumn{2}{c|}{}   & \multicolumn{2}{c|}{}   & \multicolumn{2}{c|}{P.Matching}   & \multicolumn{2}{c|}{I.Matching}   & \multicolumn{2}{c|}{}   & \multicolumn{2}{c|}{P.Matching}   & \multicolumn{2}{c|}{I.Matching}   \\ \hline 
Miner & Dataset & AA & NE & global & local & global & local & AA & NE & global & local & global & local \\ \hline 
\multirow{17}{*}{IM} & BPIC12 &  t/out  &  \bf{6,145}  &  t/out  &  t/out  &  t/out  &  t/out  &  & 1.00 &  &  &  &  \\  
 & BPIC13\textsubscript{cp} &  t/out  &  453  &  527  &  2,292  &  \bf{422}  &  2,274  &  & 1.00 & 0.17 & 0.24 & 0.17 & 0.13 \\  
 & BPIC13\textsubscript{inc} &  t/out  &  \bf{2,631}  &  2,865  &  t/out  &  2,968  &  t/out  &  & 1.00 & 0.49 &  & 0.49 &  \\  
 & BPIC14\textsubscript{f} &  t/out  &  t/out  &  \bf{59,794}  &  t/out  &  62,750  &  t/out  &  &  & 0.58 &  & 0.58 &  \\  
 & BPIC15\textsubscript{1f} &  t/out  &  t/out  &  1,217  &  5,629  &  \bf{1,132}  &  5,492  &  &  & 1.00 & 1.00 & 1.00 & 1.00 \\  
 & BPIC15\textsubscript{2f} &  t/out  &  t/out  &  \bf{6,543}  &  18,446  &  6,756  &  18,035  &  &  & 0.69 & 1.00 & 0.61 & 1.00 \\  
 & BPIC15\textsubscript{3f} &  t/out  &  t/out  &  \bf{28,121}  &  45,201  &  28,280  &  45,905  &  &  & 0.92 & 1.00 & 0.83 & 1.00 \\  
 & BPIC15\textsubscript{4f} &  t/out  &  t/out  &  \bf{3,098}  &  13,436  &  3,314  &  13,833  &  &  & 1.00 & 1.00 & 0.99 & 1.00 \\  
 & BPIC15\textsubscript{5f} &  t/out  &  t/out  &  2,305  &  15,139  &  \bf{2,288}  &  16,119  &  &  & 1.00 & 1.00 & 1.00 & 1.00 \\  
 & BPIC17\textsubscript{f} &  t/out  &  t/out  &  \bf{155,805}  &  t/out  &  156,433  &  t/out  &  &  & 0.89 &  & 0.83 &  \\  
 & RTFMP &  t/out  &  t/out  &  \bf{1,604}  &  4,881  &  1,671  &  4,690  &  &  & 0.78 & 0.52 & 0.60 & 0.50 \\  
 & SEPSIS &  t/out  &  t/out  &  \bf{7,245}  &  t/out  &  7,277  &  t/out  &  &  & 1.00 &  & 1.00 &  \\  
 & BPIC18 &  t/out  &  t/out  &  t/out  &  t/out  &  t/out  &  t/out  &  &  &  &  &  &  \\  
 & BPIC19\textsubscript{1} &  t/out  &  6,749  &  1,685  &  2,963  &  \bf{1,499}  &  2,997  &  & 0.25 & 0.84 & 1.00 & 0.49 & 1.00 \\  
 & BPIC19\textsubscript{2} &  t/out  &  t/out  &  t/out  &  t/out  &  t/out  &  t/out  &  &  &  &  &  &  \\  
 & BPIC19\textsubscript{3} &  t/out  &  t/out  &  151,472  &  t/out  &  \bf{141,325}  &  t/out  &  &  & 0.09 &  & 0.07 &  \\  
 & BPIC19\textsubscript{4} &  t/out  &  6,013  &  746  &  1,479  &  \bf{745}  &  1,425  &  & 0.78 & 0.02 & 0.01 & 0.00 & 0.00 \\ \hline \hline
\multirow{17}{*}{SM} & BPIC12 &  t/out  &  \bf{6,772}  &  t/out  &  t/out  &  t/out  &  t/out  &  & 1.00 &  &  &  &  \\  
 & BPIC13\textsubscript{cp} &  t/out  &  438  &  541  &  2,382  &  \bf{417}  &  2,368  &  & 1.00 & 0.97 & 0.94 & 0.97 & 0.86 \\  
 & BPIC13\textsubscript{inc} &  t/out  &  2,674  &  2,269  &  t/out  &  \bf{2,234}  &  t/out  &  & 1.00 & 0.99 &  & 0.99 &  \\  
 & BPIC14\textsubscript{f} &  t/out  &  t/out  &  17,390  &  t/out  &  \bf{16,450}  &  t/out  &  &  & 0.86 &  & 0.86 &  \\  
 & BPIC15\textsubscript{1f} &  t/out  &  t/out  &  1,248  &  5,595  &  \bf{1,224}  &  5,801  &  &  & 1.00 & 1.00 & 1.00 & 1.00 \\  
 & BPIC15\textsubscript{2f} &  t/out  &  t/out  &  3,840  &  16,931  &  \bf{3,746}  &  16,722  &  &  & 0.51 & 0.00 & 0.07 & 0.00 \\  
 & BPIC15\textsubscript{3f} &  t/out  &  t/out  &  \bf{10,585}  &  43,772  &  10,894  &  43,911  &  &  & 0.59 & 1.00 & 0.23 & 1.00 \\  
 & BPIC15\textsubscript{4f} &  t/out  &  t/out  &  2,606  &  13,089  &  \bf{2,569}  &  12,431  &  &  & 0.70 & 0.00 & 0.39 & 0.00 \\  
 & BPIC15\textsubscript{5f} &  t/out  &  t/out  &  \bf{2,615}  &  15,662  &  2,673  &  15,508  &  &  & 1.00 & 1.00 & 1.00 & 1.00 \\  
 & BPIC17\textsubscript{f} &  t/out  &  t/out  &  \bf{161,424}  &  t/out  &  165,203  &  t/out  &  &  & 0.94 &  & 0.83 &  \\  
 & RTFMP &  3,214  &  8,700  &  1,352  &  4,976  &  \bf{1,341}  &  4,645  & 0.03 & 0.82 & 0.83 & 0.81 & 0.61 & 0.67 \\  
 & SEPSIS &  t/out  &  t/out  &  \bf{3,303}  &  t/out  &  3,505  &  t/out  &  &  & 0.85 &  & 0.76 &  \\  
 & BPIC18 &  t/out  &  t/out  &  t/out  &  t/out  &  t/out  &  t/out  &  &  &  &  &  &  \\  
 & BPIC19\textsubscript{1} &  t/out  &  t/out  &  2,424  &  3,058  &  \bf{2,408}  &  2,815  &  &  & 0.63 & 0.94 & 0.27 & 0.91 \\  
 & BPIC19\textsubscript{2} &  t/out  &  t/out  &  t/out  &  t/out  &  t/out  &  t/out  &  &  &  &  &  &  \\  
 & BPIC19\textsubscript{3} &  t/out  &  t/out  &  6,780  &  t/out  &  6,759  &  t/out  &  &  & 0.94 &  & 0.90 &  \\  
 & BPIC19\textsubscript{4} &  t/out  &  t/out  &  476  &  1,130  &  475  &  1,141  &  &  & 0.98 & 0.99 & 0.95 & 0.99 \\ \hline 
\multicolumn{2}{|c|}{\bf{\#t/outs}}   & 33 & 25 & \bf{6} & 16 & \bf{6} & 16 & \multicolumn{6}{c}{}           \\ \cline{1-8}
\multicolumn{2}{|c|}{\bf{\#Outperforming}} & 0 & 3 & 11 & 0 & \bf{16} & 0 & \multicolumn{6}{c}{}           \\ \cline{1-8}
\end{tabular}                             
}                             
\vspace{.5\baselineskip}                             
\caption{Results of  the quantitative experiments}\label{tb:quant_eval_wBenchmarks}                             
\vspace{.5\baselineskip}                             
}}                             
\end{table}                             

\textbf{Investigating scalability.}
From Table~\ref{tb:quant_eval_wBenchmarks} we can observe that both baselines performed poorly on the real-life datasets. AA timed out for 33 out of 34 datasets, while NE timed out in 25 out of 34 cases. Further, in six datasets where the NE baseline was successful in computing a value, it computed a perfect generalization of 1. When looking into these datasets, we found that the NE baseline found no generalizations and hence just reported a generalization of 1. Hence, in total the NE baseline provided useful results only for three out of 34 datasets. These results show that the baseline measures are not suitable to compute generalization in the context of (large) real-life datasets.
The pattern-based generalization measure, however, shows promising results. For the global oracle, the measure only timed out for 6 out of 34 datasets and for the local oracle it timed out for 16 out of 34 datasets. 

This shows that the local concurrency oracle is less scalable towards larger datasets trading computation time for more precise concurrency relations. 
For the three event logs BPIC12, BPIC18 and BPIC19\textsubscript{2}, the generalization measure with the global oracle timed out, because too many concurrent patterns were found, i.e.\ BPIC12 contains 13K concurrent patterns. For these logs, we recommend to apply a stricter filter for the concurrency oracle.

The best time performance for each dataset is highlighted in bold in Table~\ref{tb:quant_eval_wBenchmarks}. In total, the pattern generalization measure with a global oracle and interleavings matching outperformed the other measures in 16 out of 34 datasets. Considering all types of matching and only the global oracle, the pattern generalization measure outperforms the other measures in 27 out of 34 datasets improving to 30 out of 34 datasets considering the three erroneous datasets of the NE measure. 
The time performance of the global oracle is acceptable since in 27 out of 34 datasets it computed the result in 17 seconds and in one dataset computed within three minutes.
We recommend applying the global concurrency oracle for larger datasets due to its better scalability. 

\textbf{Investigating generalization pattern breakdown.}
Next, we analyse the usefulness of the pattern generalization by investigating the breakdown of the generalization values for the repetitive and concurrent patterns while comparing the models of the SM and IM algorithms. For this analysis, we focus on the pattern generalization with the global oracle. Table~\ref{tb:quant_eval_detail} shows the results of the breakdown using both the partial and the interleavings matching for the concurrent patterns.

\begin{table}[htbp!]                              
{\footnotesize{                              
\setlength{\tabcolsep}{3pt}                              
\centering{                              
\begin{tabular}{|l|r|r|r r r|r r r r r|r r r r|}                              
\hline                              
 \multirow{3}{*}{\bf{Dataset}} & \multirow{3}{*}{\bf{\makecell{Initial\\trace\\count}}} & \multirow{3}{*}{\bf{\makecell{Pattern\\trace\\count}}} & \multicolumn{3}{c|}{\bf{Repetitive Patterns}}     & \multicolumn{5}{c|}{\bf{Concurrent Patterns}}         & \multicolumn{4}{|c|}{\bf{Overall Generalization}}       \\ 
 &  &  & \multicolumn{3}{c|}{Both}     & Both & \multicolumn{2}{c}{P.Matching}   & \multicolumn{2}{c|}{I.Matching}   & \multicolumn{2}{|c}{P.Matching}   & \multicolumn{2}{c|}{I.Matching}   \\ 
 &  &  & Weight &  IM & SM & Weight &  IM & SM &  IM & SM &  IM & SM &  IM & SM \\ \hline
BPIC12 &  &  t/out  & 0\% &  &  & 0\% &  &  &  &  &  &  &  &  \\ 
BPIC13\textsubscript{cp} &  1,487  &  581  & 100\% & 0.17 & \bf{ 0.97 } & 0\% &  &  &  &  & 0.17 & \bf{ 0.97 } & 0.17 & \bf{ 0.97 } \\ 
BPIC13\textsubscript{inc} &  7,554  &  9,815  & 100\% & 0.49 & \bf{ 0.99 } & 0\% &  &  &  &  & 0.49 & \bf{ 0.99 } & 0.49 & \bf{ 0.99 } \\ 
BPIC14\textsubscript{f} &  41,353  &  20,181  & 100\% & 0.58 & \bf{ 0.86 } & 0\% &  &  &  &  & 0.58 & \bf{ 0.86 } & 0.58 & \bf{ 0.86 } \\ 
BPIC15\textsubscript{1f} &  902  &  -    & 0\% & 1.00 & 1.00 & 0\% & 1.00 & 1.00 & 1.00 & 1.00 & 1.00 & 1.00 & 1.00 & 1.00 \\ 
BPIC15\textsubscript{2f} &  681  &  348  & 0.3\% & \bf{ 1.00 } & 0.00 & 99.7\% & \bf{ 0.69 } & 0.51 & \bf{ 0.61 } & 0.07 & \bf{ 0.69 } & 0.51 & \bf{ 0.61 } & 0.07 \\ 
BPIC15\textsubscript{3f} &  1,369  &  2,982  & 0\% &  &  & 100\% & \bf{ 0.92 } & 0.59 & \bf{ 0.83 } & 0.23 & \bf{ 0.92 } & 0.59 & \bf{ 0.83 } & 0.23 \\ 
BPIC15\textsubscript{4f} &  860  &  676  & 0.2\% & \bf{ 1.00 } & 0.00 & 99.8\% & \bf{ 1.00 } & 0.70 & \bf{ 0.99 } & 0.39 & \bf{ 1.00 } & 0.70 & \bf{ 0.99 } & 0.39 \\ 
BPIC15\textsubscript{5f} &  975  &  -    & 0\% & 1.00 & 1.00 & 0\% & 1.00 & 1.00 & 1.00 & 1.00 & 1.00 & 1.00 & 1.00 & 1.00 \\ 
BPIC17\textsubscript{f} &  21,861  &  157,470  & 69.4\% & 0.87 & \bf{ 1.00 } & 30.6\% & \bf{ 0.92 } & 0.81 & \bf{ 0.75 } & 0.44 & 0.89 & \bf{ 0.94 } & \bf{ 0.83 } & 0.83 \\ 
RTFMP &  150,370  &  28,213  & 13.8\% & 0.00 & \bf{ 1.00 } & 86.2\% & \bf{ 0.90 } & 0.80 & \bf{ 0.70 } & 0.55 & 0.78 & \bf{ 0.83 } & 0.60 & \bf{ 0.61 } \\ 
SEPSIS &  1,050  &  1,411  & 60.1\% & \bf{ 1.00 } & 0.93 & 39.9\% & \bf{ 1.00 } & 0.72 & \bf{ 1.00 } & 0.50 & \bf{ 1.00 } & 0.85 & \bf{ 1.00 } & 0.76 \\ 
BPIC18 &  &  t/out  & 0\% &  &  & 0\% &  &  &  &  &  &  &  &  \\ 
BPIC19\textsubscript{1} &  1,044  &  3,138  & 15.2\% & \bf{ 1.00 } & 0.95 & 84.8\% & \bf{ 0.81 } & 0.57 & \bf{ 0.40 } & 0.15 & \bf{ 0.84 } & 0.63 & \bf{ 0.49 } & 0.27 \\ 
BPIC19\textsubscript{2} &  &  t/out  & 0\% &  &  & 0\% &  &  &  &  &  &  &  &  \\ 
BPIC19\textsubscript{3} &  221,010  &  8,087  & 89.5\% &  & \bf{ 0.97 } & 10.5\% &  & \bf{ 0.74 } &  & \bf{ 0.35 } &  & \bf{ 0.94 } &  & \bf{ 0.90 } \\ 
BPIC19\textsubscript{4} &  14,498  &  1,086  & 95.8\% & 0.00 & \bf{ 1.00 } & 4.2\% & 0.45 & \bf{ 0.55 } & \bf{ 0.00 } & \bf{ 0.00 } & 0.02 & \bf{ 0.98 } & 0.00 & \bf{ 0.95 } \\ \hline
\multicolumn{3}{|c|}{\#Outperforming:}     &  & 4 & \bf{ 7 } &  & \bf{ 7 } & 2 & \bf{ 7 } & 1 & 5 & \bf{ 7 } & \bf{ 6 } & \bf{ 6 }\\ \hline
\end{tabular}                              
}                              
\vspace{.5\baselineskip}                              
\caption{Breakdown of generalization values for patterns}\label{tb:quant_eval_detail}                              
\vspace{.5\baselineskip}                              
}}                              
\end{table}                              

Table~\ref{tb:quant_eval_detail} reports the initial trace count of each real-life log and the accumulated trace count corresponding to all patterns. Comparing the two columns, we can see how much generalized behavior the pattern generalization measure identifies in each log. The fraction of the pattern trace count to the initial trace count averages at 125\% and ranges from 4\% for BPIC19\textsubscript{3} to 720\% for BPIC17\textsubscript{f}. The weight columns indicate which fraction of the pattern traces belong to repetitive or concurrent patterns, respectively. In some datasets, there only appears one kind of pattern, e.g.\ BPIC13\textsubscript{cp} only contains repetitive patterns, while in others there is a balance between the two patterns, e.g.\ the SEPSIS log exhibits 60\% repetitive and 40\% concurrent patterns. 

Since a pattern's generalization is given by the weighted average of the pattern fulfilments, it can also be aggregated only for pattern types, i.e.\ we achieve two generalization values ranging from zero to one for repetitive and concurrent patterns, respectively, by only taking the weighted average of all pattern fulfilments of the corresponding type.
For the repetitive patterns, the generalization values of the SM models outperform the IM models in seven out of eleven cases. 

As for the concurrent patterns, we distinguish the comparison for using partial matching or interleavings matching. For partial matching, the IM models outperform the SM models in 7 out of 9 models and in 7 out of 8 for the interleavings matching. The only model for which the ranking of the concurrent generalization changes is BPIC19\textsubscript{4}, where in partial matching SM outperforms IM while for interleavings matching the two mining algorithms have a tie. These results are in line with the characteristics of the process models, i.e.\ the SM models only contain one AND-split in total for all models, and hence the generalization for concurrent patterns should be lower than for the IM models.
 
For the overall generalization, the SM models outperform the IM models in seven out of twelve cases and for the interleavings matching the two mining algorithms are tied. The difference between the two matching settings can be observed in BPIC17\textsubscript{f}, where the generalization of the concurrent patterns changed from 0.72 to 0.5 and hence the ranking of the overall generalization changed. 
When choosing the matching setting for concurrent patterns, interleavings matching will evaluate concurrency more strictly and will lower the generalization more quickly, when no parallel gateways are used in a model.

The breakdown of generalization values demonstrates the capabilities of the pattern generalization to find more fine-grained generalization problems for large datasets. In our experiments, we observed that SM models performed better for generalizing repetitive patterns while IM models performed better for concurrent patterns. The measure can then be further used to identify problematic patterns with the highest weight to discover generalization improvement ideas for a process model.
For example, for the RTFMP event log we identified the unfulfilled concurrent pattern with the highest trace count of 9K of 28K (32\%). It involves activities ``Add penalty'' and ``Payment'' which should be concurrent at trace positions 3 and 4 for the trace $\langle \text{Create Fine},\text{Send Fine},\text{Insert Fine Notification},\text{Add penalty},\text{Payment}\rangle$. Fixing a pattern, however, needs to be undertaken carefully since it might negatively affect other patterns. How to use the pattern fulfilments for automatically improving the generalization is a promising avenue for future work. 

\subsection{Threats to validity}
\noindent This evaluation exhibits threats to internal and external validity.

\textbf{Internal validity.} A possible threat to the internal validity of the experiments is provided by the limited selection of input parameters for the proposed pattern-based generalization measure. This measure can be configured by configuring a filter parameter for either the global or the local concurrency oracle as well as by specifying the type of matching for the concurrent patterns, i.e.\ partial or interleavings matching. While we tested both matching types for concurrent patterns, we only consider two parameter values for the filters for the concurrency oracles: no filter setting for the qualitative evaluation and a light filter of 5\% for the global oracle and 10\% for balance and 55\% for occurrence for the local concurrency oracle. We did not test higher filter settings for both concurrency oracles. However, higher filtering would only decrease the amount of concurrent patterns, hence increasing the scalability of the the measure. Therefore, limiting the filtering parameter to only low filtering values does not pose any concrete threat to internal validity.


\textbf{External validity.} The selection of datasets is an external threat to the validity of our experiments since the generalization measures might perform differently on a different set of event logs and process models. However, we used both artificial logs as well as large real-life event logs and process models discovered from two well-established discovery algorithms, namely Split Miner and Inductive miner. Hence, the results should be representative of the performance of the generalization measures. We also published all logs and process models used in both the qualitative and quantitative experiment to ensure the reproducibility of the evaluation results.

\newpage
\section{Conclusion}\label{sec:conclusion}
This article contributes a framework of generalization measures in the field of automated process discovery based on the idea that patterns in an event log should be generalized with corresponding structures in a process model, i.e.\ a repeating sequence should be captured with a loop in the process model. 
The patterns are defined with a set of representative traces that capture their behavior. The traces are then aligned with the process model and each pattern is assigned a partial fulfilment score by comparing the trace positions of the patterns against the alignments. The overall process model generalization w.r.t. the event log is computed as the average of all partial fulfilments weighted with the corresponding trace counts of the patterns. Hence, more frequent patterns will have a higher influence on the generalization value.

The article instantiates the framework by proposing a pattern generalization measure for repetitive and concurrent patterns.
The repetitive patterns are identified with the tandem repeats of the event log. Repeats are first reduced to collapse several traces and then extended such that a process model is forced to traverse a corresponding loop if possible when aligning the extended traces.
The partial fulfilment of repetitive patterns is the fraction of repeating labels that can be matched in every iteration of the extended tandem.
Concurrent patterns are identified with a concurrency oracle by constructing partial orders for the traces of the event log.
The representative traces and the concurrent trace positions are computed with a breadth first search over the partial orders.
The partial fulfilments of concurrent patterns can be measured via two methods: with partial matching as the fraction of matches at the trace positions of the alignments, or with interleavings matching as the fraction of alignments that align all concurrent labels with matches.

Our pattern generalization measure was tested extensively, using qualitative and quantitative evaluations focusing on the meaningfulness of the ranking of models and the scalability to large real-life datasets. The qualitative evaluation focused on testing a range of models with different generalizing characteristics against event logs with a different amount of patterns. We found that our pattern generalization measure always identifies the patterns in the event log and accompanying issues/fulfilment in the process model, while baseline techniques tend to ignore such patterns and merely evaluate generalization based on model characteristics.
The qualitative evaluation also highlights the limitation of our measure, i.e.\ the fact that it is less sensitive in ranking process models when the event log contains no or only a small amount of patterns. 

In the quantitative evaluation, the pattern generalization is shown to outperform existing baseline measures systematically in terms of execution time, with the baseline measures timing out in the great majority of datasets. 
Further, we highlight the strength of the pattern generalization to drill down the generalization by pattern types in order to determine the strengths and weaknesses of process models or to find the most frequent unfulfilled patterns that highlight  generalization improvement ideas for process models.

An avenue for future work is to initiate our pattern generalization framework with a larger set of patterns. Candidate patterns could be nested patterns such as repeating structures inside concurrent patterns and vice versa. With an increasing amount of patterns, we expect our measure to more finely distinguish the ability of process models to generalize. We also found that several patterns such as the workflow patterns~\cite{WorkflowPatterns} or change patterns~\cite{ChangePatterns} have not been linked to the recorded behavior of a process yet. It would be interesting to investigate how these patterns would manifest in an event log to better abstract the behavior of a process and extend the set of available patterns for the proposed framework.
Another avenue for future work is to automatically identify a set of unfulfilled patterns that could be used to improve a model's generalization ability. This is not a trivial endeavor since selecting to generalize a pattern by including the corresponding control structure in the process model might overlap with generalizing another pattern. 
Last, this article tackled the issue of identifying generalization issues by testing if patterns from an event log are fulfilled by control-flow structures in a model. Conversely, an avenue for future work could be to define a precision measure by identifying all control structures of a process model and check if they cover any corresponding pattern in the event log. If a control structure cannot identify a corresponding pattern in the log, the precision of a process model should be lowered since it includes behavior that is not contained in the log. This novel definition of a precision measure is promising since the behavior of a model is possibly infinite and hence comparing its behavior to a finite event log is difficult. However, the control-flow structures of a process model are finite and hence a comparison with the patterns of an event log is much more feasible.

\medskip
\noindent \emph{Acknowledgements.} This research is partly funded by the Australian Research Council (grant DP180102839).
\vspace{-1\baselineskip}



\bibliographystyle{elsarticle-num}
\bibliography{lit}
\appendix
\newpage
\section{Additional Event Logs for the Anti-Alignments dataset}\label{app:AA_additionalLogs}
\input{tex/figureAAadditionalLogs.tex}
\newpage
\section{Time performance of the qualitative experiments}\label{app:cost_comparison}
\begin{table}[htbp!]                
{\footnotesize{                
\setlength{\tabcolsep}{3pt}                
\centering{                
\begin{tabular}{|c|l|r r|r r|r r|}                
\cline{3-8}
    \multicolumn{2}{c|}{} & \multicolumn{2}{c|}{\multirow{2}{*}{\bf{Baselines}}} & \multicolumn{4}{c|}{\bf{Pattern Generalization}} \\          
  \multicolumn{2}{c|}{} & \multicolumn{2}{c|}{}   & \multicolumn{2}{c|}{\bf{P. Matching}}   & \multicolumn{2}{c|}{\bf{I. Matching}}   \\ \hline
\multicolumn{2}{|c|}{\bf{Dataset}} & AA & NE & global & local & global & local \\ \hline
 \multirow{9}{*}{\bf{\makecell{AA\\original}}} & Generating M. &  164  &  183  &  263  &  346  &  266  &  342  \\ 
 & Single trace &  141  &  148  &  222  &  298  &  224  &  300  \\ 
 & Flower M. &  150  &  151  &  230  &  317  &  229  &  309  \\ 
 & Distinct traces &  253  &  247  &  230  &  315  &  227  &  309  \\ 
 & G,H parallel &  183  &  189  &  275  &  359  &  278  &  357  \\ 
 & G,H self loop &  163  &  173  &  247  &  329  &  252  &  330  \\ 
 & D self loop &  165  &  t/out  &  255  &  336  &  252  &  338  \\ 
 & All parallel &  424  &  289  &  4,962  &  5,094  &  4,958  &  5,088  \\ 
 & Round robin &  258  &  252  &  305  &  393  &  312  &  391  \\ \hline
 \multirow{9}{*}{\bf{\makecell{AA\\repetitive}}} & Generating M. &  174  &  322  &  308  &  427  &  306  &  429  \\ 
 & Single trace &  154  &  199  &  264  &  393  &  266  &  390  \\ 
 & Flower M. &  160  &  217  &  269  &  388  &  269  &  405  \\ 
 & Distinct traces &  207  &  497  &  290  &  406  &  285  &  401  \\ 
 & G,H parallel &  182  &  420  &  320  &  472  &  315  &  458  \\ 
 & G,H self loop &  168  &  t/out  &  290  &  416  &  289  &  435  \\ 
 & D self loop &  178  &  t/out  &  301  &  441  &  306  &  438  \\ 
 & All parallel &  499  &  55,135  &  8,181  &  8,272  &  8,251  &  8,248  \\ 
 & Round robin &  294  &  7,784  &  431  &  485  &  433  &  498  \\ \hline
 \multirow{9}{*}{\bf{\makecell{AA\\concurrent}}} & Generating M. &  171  &  327  &  273  &  361  &  273  &  372  \\ 
 & Single trace &  139  &  200  &  235  &  331  &  230  &  331  \\ 
 & Flower M. &  218  &  221  &  242  &  333  &  245  &  334  \\ 
 & Distinct traces &  215  &  515  &  246  &  345  &  246  &  350  \\ 
 & G,H parallel &  181  &  384  &  288  &  389  &  293  &  397  \\ 
 & G,H self loop &  191  &  t/out  &  264  &  362  &  265  &  368  \\ 
 & D self loop &  171  &  t/out  &  274  &  370  &  280  &  370  \\ 
 & All parallel &  946  &  433  &  2,021  &  2,106  &  2,014  &  2,108  \\ 
 & Round robin &  303  &  999  &  348  &  435  &  335  &  433  \\ \hline
 \multirow{9}{*}{\bf{\makecell{AA\\composite}}} & Generating M. &  246  &  568  &  397  &  562  &  402  &  566  \\ 
 & Single trace &  194  &  360  &  347  &  523  &  351  &  526  \\ 
 & Flower M. &  303  &  373  &  361  &  532  &  374  &  534  \\ 
 & Distinct traces &  326  &  856  &  432  &  550  &  416  &  562  \\ 
 & G,H parallel &  260  &  644  &  429  &  584  &  433  &  589  \\ 
 & G,H self loop &  299  &  t/out  &  366  &  554  &  366  &  554  \\ 
 & D self loop &  254  &  t/out  &  389  &  566  &  388  &  570  \\ 
 & All parallel &  1,356  &  40,567  &  13,900  &  12,425  &  13,959  &  12,340  \\ 
 & Round robin &  380  &  751  &  712  &  706  &  704  &  729  \\ \hline
 \multirow{6}{*}{\bf{NE}} & Perfect M. &  334  &  348  &  278  &  392  &  277  &  386  \\ 
 & Single trace &  127  &  199  &  243  &  362  &  249  &  363  \\ 
 & Flower model &  217  &  230  &  237  &  360  &  243  &  361  \\ 
 & Distinct traces &  219  &  469  &  258  &  364  &  258  &  376  \\ 
 & SM &  t/out  &  t/out  &  251  &  362  &  251  &  375  \\ 
 & IM &  t/out  &  840  &  297  &  411  &  303  &  415  \\ \hline
\end{tabular}                
}                
\vspace{.5\baselineskip}                
\caption{Time performance in ms of the qualitative experiments}\label{tb:qual_eval_time}                
\vspace{.5\baselineskip}                
}}                
\end{table}                






\end{document}